%% file: main.tex
\pdfoutput=1
\documentclass{article}
\usepackage{iclr2027_conference,times}
\usepackage[T1]{fontenc}
\usepackage{amsmath,amssymb,booktabs,graphicx,xspace}
\usepackage{xcolor}
\usepackage{enumitem}
\setlist[itemize]{topsep=0pt,itemsep=0pt,parsep=0pt,partopsep=0pt,leftmargin=*}
\newcommand{\rev}[1]{#1}
\newcommand{\revb}[1]{#1}
\newcommand{\revc}[1]{#1}

\newcommand{\reve}[1]{#1}
\DeclareRobustCommand{\revf}[1]{#1}
\newcommand{\revr}[1]{{\color{blue}#1}}
\usepackage[normalem]{ulem}
\newcommand{\revcut}[1]{{\color{red}\sout{#1}}}
\newcommand{\revp}[1]{{\color{blue}#1}}
\newcommand{\revg}[1]{{\color{blue}#1}}
\newcommand{\revtable}{\color{blue}}
\usepackage{tikz}\usetikzlibrary{shapes.geometric,arrows.meta}
\definecolor{filtgreen}{HTML}{18AC78}\definecolor{filtred}{HTML}{E04848}\definecolor{filtblue}{HTML}{2878D4}
\DeclareRobustCommand{\symfirst}{\tikz[baseline=-0.55ex]\node[star,star points=5,star point ratio=2.4,draw=black,line width=0.35pt,fill=filtgreen,inner sep=0pt,minimum size=1.8ex]{};}
\DeclareRobustCommand{\symlast}{\tikz[baseline=-0.55ex]\node[rectangle,draw=black,line width=0.35pt,fill=filtred,inner sep=0pt,minimum size=1.2ex]{};}
\DeclareRobustCommand{\symcontact}{\tikz[baseline=-0.55ex]\node[star,star points=5,star point ratio=2.4,draw=black,line width=0.35pt,fill=white,inner sep=0pt,minimum size=1.8ex]{};}
\DeclareRobustCommand{\symspan}{\tikz[baseline=-0.55ex]{\draw[filtblue,line width=1.2pt] (0,0)--(1.5em,0); \foreach \x in {0.2em,0.75em,1.3em} \filldraw[fill=filtblue,draw=white,line width=0.3pt] (\x,0) circle (0.42ex);}}
\DeclareRobustCommand{\sympath}{\tikz[baseline=-0.55ex]\draw[black!22,line width=1.4pt] (0,0)--(1.5em,0);}
\definecolor{clickred}{HTML}{E8412C}
\DeclareRobustCommand{\symclick}{\tikz[baseline=-0.55ex]\node[diamond,fill=clickred,draw=clickred!75!black,line width=0.3pt,inner sep=0pt,minimum size=1.55ex]{};}
\definecolor{fsgreen}{HTML}{00FF55}\definecolor{fsyellow}{HTML}{FFEE00}
\DeclareRobustCommand{\symfstip}{\tikz[baseline=-0.55ex]\filldraw[fill=fsgreen,draw=black!55,line width=0.3pt] (0,0) circle (0.5ex);}
\DeclareRobustCommand{\symfsforce}{\tikz[baseline=-0.55ex]{\draw[-Stealth,black!55,line width=1.8pt] (0,0)--(1.45em,0); \draw[-Stealth,fsyellow,line width=1.1pt] (0.04em,0)--(1.4em,0);}}
\DeclareRobustCommand{\symfsgrip}{\tikz[baseline=-0.55ex]{\foreach \a/\b in {0em/0.64em,1.5em/0.86em}{\draw[-Stealth,black!55,line width=1.8pt] (\a,0)--(\b,0); \draw[-Stealth,fsgreen,line width=1.1pt] (\a,0)--(\b,0);}}}
\DeclareRobustCommand{\symwindow}{\tikz[baseline=-0.55ex]\draw[black!60,dash pattern=on 1.4pt off 1pt,line width=0.6pt] (0,-0.5ex) rectangle (1.3em,0.9ex);}
\usepackage{hyperref,url}
\renewcommand{\paragraph}[1]{\par\noindent\textbf{#1}\ \ignorespaces}
\usepackage{etoc}
\makeatletter
\newcommand{\compacttoc}{\begingroup\small\setlength{\parskip}{0pt}%
  \renewcommand{\color}[2][]{}%
  \renewcommand*\l@section[2]{\ifnum\c@tocdepth>\z@\addpenalty\@secpenalty\addvspace{0.35em \@plus\p@}%
    \setlength\@tempdima{1.5em}\begingroup\parindent\z@\rightskip\@pnumwidth\parfillskip-\@pnumwidth
    \leavevmode\bfseries\advance\leftskip\@tempdima\hskip-\leftskip ##1\nobreak\hfil\nobreak
    \hb@xt@\@pnumwidth{\hss ##2\kern-\p@\kern\p@}\par\endgroup\fi}%
  \tableofcontents\endgroup}
\makeatother
\makeatletter
\g@addto@macro\normalsize{\abovedisplayskip 4pt plus 1pt minus 1pt\relax\belowdisplayskip 4pt plus 1pt minus 1pt\relax
  \abovedisplayshortskip 4pt plus 1pt minus 1pt\relax\belowdisplayshortskip 4pt plus 1pt minus 1pt\relax}
\makeatother
\usepackage{multirow}
\usepackage{placeins}
\AddToHook{env/table/begin}{\setlength{\belowcaptionskip}{4pt}}
\AddToHook{env/figure/begin}{\setlength{\abovecaptionskip}{5pt}}
\hypersetup{hidelinks}
\newcommand{\method}{VidForce\xspace}
\newcommand{\sg}{\operatorname{sg}}
\newcommand{\EvalP}{Eval$'$\xspace}

\input{tables/evidence_macros}
\input{tables/experiment_macros}

\input{tables/results_macros}

\input{tables/rebuttal_guidance_macros}

\input{tables/sim_macros}

\input{tables/speed_macros}
\input{tables/speed_section_macros}

\input{tables/speed_eval_macros}
\input{tables/speed_drawer_macros}

\input{tables/speed_same_object_macros}

\input{tables/speed_action_thirds_macros}
\input{tables/speed_force_executions_macros}

\input{tables/rebuttal_category_input_macros}

\input{tables/rebuttal_force_range_macros}

\newcommand{\abltabscale}{0.857143}
\input{tables/prior_force_range_macros}
\newcounter{latetable}\newcounter{latefigure}
\providecommand{\lasttablenumber}{50}\providecommand{\lastfigurenumber}{18}
\makeatletter
\AtEndDocument{\immediate\write\@auxout{\string\gdef\string\lasttablenumber{\arabic{table}}}%
  \immediate\write\@auxout{\string\gdef\string\lastfigurenumber{\arabic{figure}}}}
\makeatother
\newcommand{\latetable}{\stepcounter{latetable}\addtocounter{table}{-1}%
  \renewcommand{\thetable}{\the\numexpr\lasttablenumber+\value{latetable}\relax}%
  \renewcommand{\theHtable}{late.\arabic{latetable}}}
\newcommand{\latefigure}{\stepcounter{latefigure}\addtocounter{figure}{-1}%
  \renewcommand{\thefigure}{\the\numexpr\lastfigurenumber+\value{latefigure}\relax}%
  \renewcommand{\theHfigure}{late.\arabic{latefigure}}}
\input{tables/profile_scaled_macros}
\iclrfinalcopy
\newcommand{\preprintheader}{Preprint.}
\renewcommand{\revr}[1]{#1}\renewcommand{\revp}[1]{#1}\renewcommand{\revg}[1]{#1}
\renewcommand{\revtable}{}\profscaledmarksfalse
\renewcommand{\revcut}[1]{\ignorespaces}
\renewcommand{\latetable}{}\renewcommand{\latefigure}{}
\title{VidForce: Predicting Force-Aware\\Articulation by Observing Humans}
\author{Rafay Mohiuddin$^{1,2,*}$\quad Jiaqi Chen$^{3}$\quad Marc Pollefeys$^{3,4}$\quad Hermann Blum$^{1,2}$\\
$^{1}$University of Bonn\quad $^{2}$Lamarr Institute\quad $^{3}$ETH Zurich\quad $^{4}$Microsoft\\
$^{*}$Corresponding author: \texttt{rmohiudd@uni-bonn.de}}
\begin{document}
\etocdepthtag.toc{mtbody}
\maketitle
\lhead{\preprintheader}
\begin{abstract}
\input{sections/abstract}
\end{abstract}
\section{Introduction}
\input{sections/introduction}
\section{Related Work}
\input{sections/related_work}
\section{Method}
\input{sections/method}
\section{Experiments}
\input{sections/experiments}
\input{sections/realworld}
\input{sections/limitations}
\section{Conclusion}
\input{sections/conclusion}
\input{sections/acknowledgments}
\input{sections/llm_statement}
\bibliography{references}
\bibliographystyle{iclr2027_conference}
\clearpage
\appendix
\renewcommand{\revcut}[1]{\ignorespaces}
\makeatletter
\let\obs@appsection\section
\renewcommand{\section}{\FloatBarrier\obs@appsection}
\renewenvironment{figure}[1][!htbp]{\@float{figure}[!htbp]}{\end@float}
\renewenvironment{table}[1][!htbp]{\@float{table}[!htbp]}{\end@float}
\makeatother
\etocdepthtag.toc{mtappendix}
\etocsettagdepth{mtbody}{none}
\etocsettagdepth{mtappendix}{subsection}
\renewcommand{\contentsname}{Appendix Contents}
\compacttoc
\section{Implementation Details}
\input{sections/appendix_method}

\section{Data and Processing}
\label{app:data}
\input{sections/appendix_data}
\section{Supplementary Evaluation}
\label{app:evidence}
\input{sections/appendix_experiments}
\input{sections/appendix_profile_scaling}
\section{Real-World Interactions and Robot Deployment}
\label{app:realworld}
\input{sections/appendix_realworld}
\end{document}

%% file: tables/experiment_macros.tex
\newcommand{\obsSelectedATE}{6.6}
\newcommand{\obsSelectedMAE}{4.81}
\newcommand{\obsSelectedRMSE}{5.92}

%% file: tables/results_macros.tex
\newcommand{\oursAte}{6.6}
\newcommand{\oursAteSd}{0.4}
\newcommand{\oursMae}{4.81}
\newcommand{\oursMaeSd}{0.38}
\newcommand{\oursRmse}{5.92}
\newcommand{\oursRmseSd}{0.42}
\newcommand{\oursPriorMae}{6.88}
\newcommand{\oursPriorMaeSd}{0.29}
\newcommand{\oursPriorRmse}{11.28}

\newcommand{\oursAdaptMae}{4.84}
\newcommand{\oursAdaptMaeSd}{0.31}
\newcommand{\oursAdaptRmse}{5.94}

\newcommand{\oursHandAte}{7.9}
\newcommand{\oursHandAteSd}{0.8}
\newcommand{\oursGripperAte}{5.6}
\newcommand{\oursGripperAteSd}{0.2}

\newcommand{\unseenAte}{6.7}
\newcommand{\unseenAteSd}{0.3}
\newcommand{\unseenMae}{4.77}
\newcommand{\unseenMaeSd}{0.54}

\newcommand{\seenAte}{6.6}
\newcommand{\seenAteSd}{0.4}
\newcommand{\seenMae}{4.84}
\newcommand{\seenMaeSd}{0.24}

\newcommand{\oursHardAte}{7.2}
\newcommand{\oursHardAteSd}{0.3}
\newcommand{\oursHardMae}{5.02}
\newcommand{\oursHardMaeSd}{0.32}
\newcommand{\oursHardRmse}{6.27}
\newcommand{\oursHardRmseSd}{0.49}
\newcommand{\oursHardHandAte}{8.2}
\newcommand{\oursHardHandAteSd}{0.5}

\newcommand{\selfClickAte}{6.9}
\newcommand{\selfClickAteSd}{0.9}
\newcommand{\selfClickMae}{4.49}
\newcommand{\selfClickMaeSd}{0.50}
\newcommand{\selfClickRmse}{5.53}
\newcommand{\selfClickRmseSd}{0.67}

\newcommand{\selfClickRawAte}{7.1}
\newcommand{\selfClickRawAteSd}{0.5}
\newcommand{\selfClickRawMae}{4.48}
\newcommand{\selfClickRawMaeSd}{0.48}
\newcommand{\selfClickRawRmse}{5.55}
\newcommand{\selfClickRawRmseSd}{0.59}
\newcommand{\selfClickRawLoc}{25}

\newcommand{\selfClickRawHandAte}{8.8}
\newcommand{\selfClickRawHandAteSd}{0.9}
\newcommand{\selfPromptAte}{6.8}
\newcommand{\selfPromptAteSd}{0.5}
\newcommand{\selfPromptMae}{5.85}
\newcommand{\selfPromptMaeSd}{0.10}
\newcommand{\selfPromptRmse}{7.58}
\newcommand{\selfPromptRmseSd}{0.26}

\newcommand{\selfPromptSelfClickAte}{7.8}
\newcommand{\selfPromptSelfClickAteSd}{0.4}
\newcommand{\selfPromptSelfClickMae}{5.66}
\newcommand{\selfPromptSelfClickMaeSd}{0.38}
\newcommand{\selfPromptSelfClickRmse}{7.87}
\newcommand{\selfPromptSelfClickRmseSd}{1.04}
\newcommand{\selfPromptSelfClickLoc}{26}

\newcommand{\profileRmse}{4.95}
\newcommand{\profileRmseSd}{0.20}

\newcommand{\locPxSelfClick}{18}

\newcommand{\nVal}{99}
\newcommand{\nValForce}{48}
\newcommand{\nValHand}{51}
\newcommand{\nValGripper}{48}
\newcommand{\nHard}{594}
\newcommand{\nHardForce}{288}

\newcommand{\nUnseen}{47}
\newcommand{\nUnseenForce}{23}
\newcommand{\nSeen}{52}
\newcommand{\nSeenForce}{25}

\newcommand{\nRecords}{630}
\newcommand{\nGripper}{343}
\newcommand{\nHuman}{287}
\newcommand{\nTrain}{531}
\newcommand{\nTrainGripper}{295}
\newcommand{\nTrainHuman}{236}
\newcommand{\nPaired}{201}
\newcommand{\nPairedSame}{169}
\newcommand{\nPairedCross}{32}

\newcommand{\nHardHand}{306}

\newcommand{\ateDrt}{11.8}
\newcommand{\ateDrtHuman}{11.1}
\newcommand{\ateDrtHard}{12.8}
\newcommand{\ateDrtSelf}{12.0}
\newcommand{\ateCt}{13.3}
\newcommand{\ateCtHuman}{14.2}
\newcommand{\ateCtHard}{12.6}

\newcommand{\ateCtSelf}{12.4}
\newcommand{\ateAA}{17.5}
\newcommand{\ateAAHuman}{17.3}

\newcommand{\aaFailed}{5}
\newcommand{\aaFailedHard}{40}
\newcommand{\ateEgoHuman}{15.1}

\newcommand{\ateCatMean}{16.5}
\newcommand{\ateStay}{17.5}
\newcommand{\ateStayHuman}{18.7}
\newcommand{\ateStayGripper}{13.2}

\newcommand{\ateCatMeanHuman}{19.1}
\newcommand{\ateCatMeanGripper}{13.0}

\newcommand{\ateDrtGripper}{12.1}

\newcommand{\ateCtGripper}{13.1}

\newcommand{\ateAAGripper}{18.7}

\newcommand{\ateDrtSelfHuman}{13.3}
\newcommand{\ateDrtSelfGripper}{11.6}
\newcommand{\ateCtSelfHuman}{16.5}
\newcommand{\ateCtSelfGripper}{11.4}
\newcommand{\selfClickRawGripperAte}{5.7}
\newcommand{\selfClickRawGripperAteSd}{0.0}

\newcommand{\maeConst}{8.98}
\newcommand{\rmseConst}{15.88}

\newcommand{\maeTable}{6.09}
\newcommand{\rmseTable}{8.57}
\newcommand{\maeGptText}{13.03}
\newcommand{\rmseGptText}{15.17}
\newcommand{\maeGptImg}{6.85}
\newcommand{\rmseGptImg}{11.88}
\newcommand{\maeGpt}{6.98}
\newcommand{\rmseGpt}{11.43}

\newcommand{\maeGptCrop}{7.75}
\newcommand{\rmseGptCrop}{12.35}
\newcommand{\maeGptCropText}{7.99}
\newcommand{\rmseGptCropText}{11.98}
\newcommand{\tqGptImg}{0.693}
\newcommand{\tqGptImgRmse}{0.842}

\newcommand{\tqGpt}{0.550}
\newcommand{\tqGptRmse}{0.725}

\newcommand{\oursHandRef}{10.64}
\newcommand{\oursHandRefSd}{0.50}
\newcommand{\oursHandRefRmse}{14.09}
\newcommand{\oursHandRefRmseSd}{0.72}
\newcommand{\oursHandBias}{2.1}
\newcommand{\oursHandBiasSd}{0.2}

\newcommand{\nHandRef}{40}
\newcommand{\nopairHandRef}{18.79}
\newcommand{\nopairHandRefSd}{4.14}
\newcommand{\nopairHandRefRmse}{21.21}
\newcommand{\nopairHandRefRmseSd}{3.63}
\newcommand{\nopairHandBias}{12.9}
\newcommand{\nopairHandBiasSd}{6.2}

\newcommand{\gptHandRef}{8.32}
\newcommand{\gptHandRefRmse}{13.09}
\newcommand{\gptHandBias}{-0.7}
\newcommand{\gptImgHandRef}{7.95}
\newcommand{\gptImgHandRefRmse}{13.72}
\newcommand{\gptImgHandBias}{-1.8}

\newcommand{\fsZsGripViewMae}{12.33}
\newcommand{\fsZsGripViewRmse}{19.53}

\newcommand{\fsLGripViewMae}{7.56}
\newcommand{\fsLGripViewMaeSd}{0.28}
\newcommand{\fsLGripViewRmse}{10.84}
\newcommand{\fsLGripViewRmseSd}{0.35}
\newcommand{\fsLGripViewProf}{4.71}
\newcommand{\fsLGripViewProfSd}{0.16}

\newcommand{\fsLCropMae}{10.30}
\newcommand{\fsLCropMaeSd}{0.75}
\newcommand{\fsLCropRmse}{16.60}
\newcommand{\fsLCropRmseSd}{0.92}
\newcommand{\fsLCropProf}{7.55}
\newcommand{\fsLCropProfSd}{0.27}

\newcommand{\fsLCropHandRef}{11.00}
\newcommand{\fsLCropHandRefSd}{0.46}
\newcommand{\fsLCropHandRefRmse}{17.64}
\newcommand{\fsLCropHandRefRmseSd}{0.83}
\newcommand{\fsLCropHandBias}{-10.4}
\newcommand{\fsLCropHandBiasSd}{0.7}
\newcommand{\fsLGripCropMae}{10.02}
\newcommand{\fsLGripCropMaeSd}{0.27}
\newcommand{\fsLGripCropRmse}{15.93}
\newcommand{\fsLGripCropRmseSd}{0.35}
\newcommand{\fsLGripCropProf}{7.39}
\newcommand{\fsLGripCropProfSd}{0.18}

\newcommand{\fsLGripCropHandRef}{12.08}
\newcommand{\fsLGripCropHandRefSd}{0.29}
\newcommand{\fsLGripCropHandRefRmse}{19.18}
\newcommand{\fsLGripCropHandRefRmseSd}{0.46}
\newcommand{\fsLGripCropHandBias}{-11.6}
\newcommand{\fsLGripCropHandBiasSd}{0.4}
\newcommand{\fsFtCropMae}{10.16}
\newcommand{\fsFtCropMaeSd}{0.26}
\newcommand{\fsFtCropRmse}{16.79}
\newcommand{\fsFtCropRmseSd}{0.49}

\newcommand{\fsFtMae}{7.26}
\newcommand{\fsFtMaeSd}{0.49}
\newcommand{\fsFtRmse}{10.29}
\newcommand{\fsFtRmseSd}{0.87}

\newcommand{\fsZsCropMae}{13.75}
\newcommand{\fsZsCropRmse}{20.96}

\newcommand{\fsZsCropHandRef}{16.11}
\newcommand{\fsZsCropHandRefRmse}{23.61}
\newcommand{\fsZsCropHandBias}{-16.1}

\newcommand{\mechMlpAte}{6.6}
\newcommand{\mechMlpAteSd}{0.4}
\newcommand{\mechMlpMae}{5.79}
\newcommand{\mechMlpMaeSd}{0.17}
\newcommand{\mechMlpRmse}{8.68}
\newcommand{\mechMlpRmseSd}{0.03}
\newcommand{\mechMlpProf}{5.72}
\newcommand{\mechMlpProfSd}{0.02}
\newcommand{\mechMlpPriorMae}{8.35}
\newcommand{\mechMlpPriorMaeSd}{0.04}

\newcommand{\mechMlpCorrMae}{5.78}
\newcommand{\mechMlpCorrMaeSd}{0.17}

\newcommand{\mechMlpHardAte}{7.4}
\newcommand{\mechMlpHardAteSd}{0.1}
\newcommand{\mechMlpHardMae}{6.17}
\newcommand{\mechMlpHardMaeSd}{0.30}
\newcommand{\mechMlpHardRmse}{9.18}
\newcommand{\mechMlpHardRmseSd}{0.27}
\newcommand{\mechMlpHardProf}{5.84}
\newcommand{\mechMlpHardProfSd}{0.11}
\newcommand{\mechGlobalAte}{6.7}
\newcommand{\mechGlobalAteSd}{0.1}
\newcommand{\mechGlobalMae}{5.61}
\newcommand{\mechGlobalMaeSd}{0.28}
\newcommand{\mechGlobalRmse}{7.64}
\newcommand{\mechGlobalRmseSd}{0.21}
\newcommand{\mechGlobalProf}{5.73}
\newcommand{\mechGlobalProfSd}{0.05}
\newcommand{\mechGlobalPriorMae}{8.99}
\newcommand{\mechGlobalPriorMaeSd}{0.00}

\newcommand{\mechGlobalCorrMae}{6.95}
\newcommand{\mechGlobalCorrMaeSd}{0.29}

\newcommand{\mechGlobalHardAte}{7.7}
\newcommand{\mechGlobalHardAteSd}{0.5}
\newcommand{\mechGlobalHardMae}{5.73}
\newcommand{\mechGlobalHardMaeSd}{0.22}
\newcommand{\mechGlobalHardRmse}{8.02}
\newcommand{\mechGlobalHardRmseSd}{0.17}
\newcommand{\mechGlobalHardProf}{5.85}
\newcommand{\mechGlobalHardProfSd}{0.15}
\newcommand{\mechNorouteAte}{6.9}
\newcommand{\mechNorouteAteSd}{0.4}
\newcommand{\mechNorouteMae}{5.64}
\newcommand{\mechNorouteMaeSd}{0.21}
\newcommand{\mechNorouteRmse}{7.26}
\newcommand{\mechNorouteRmseSd}{0.34}
\newcommand{\mechNorouteProf}{5.53}
\newcommand{\mechNorouteProfSd}{0.15}
\newcommand{\mechNoroutePriorMae}{7.91}
\newcommand{\mechNoroutePriorMaeSd}{0.13}

\newcommand{\mechNorouteCorrMae}{5.68}
\newcommand{\mechNorouteCorrMaeSd}{0.19}

\newcommand{\mechNorouteHardAte}{7.4}
\newcommand{\mechNorouteHardAteSd}{0.3}
\newcommand{\mechNorouteHardMae}{5.86}
\newcommand{\mechNorouteHardMaeSd}{0.40}
\newcommand{\mechNorouteHardRmse}{7.68}
\newcommand{\mechNorouteHardRmseSd}{0.60}
\newcommand{\mechNorouteHardProf}{5.69}
\newcommand{\mechNorouteHardProfSd}{0.09}

\newcommand{\mechNorouteEff}{2.9}
\newcommand{\mechNorouteEffSd}{0.6}
\newcommand{\mechFullEff}{7.0}
\newcommand{\mechFullEffSd}{0.1}
\newcommand{\mechNorouteHardEntries}{4.0}
\newcommand{\mechNorouteHardEntriesSd}{1.0}
\newcommand{\flowInitAte}{11.8}

\newcommand{\flowAHardProf}{5.11}
\newcommand{\flowAHardProfSd}{0.12}

\newcommand{\flowRegressAte}{10.3}
\newcommand{\flowRegressAteSd}{1.3}

\newcommand{\flowRegressMae}{6.73}
\newcommand{\flowRegressMaeSd}{0.38}
\newcommand{\flowRegressRmse}{10.20}
\newcommand{\flowRegressRmseSd}{0.55}
\newcommand{\flowRegressProf}{6.47}
\newcommand{\flowRegressProfSd}{0.19}
\newcommand{\flowRegressHardAte}{11.9}
\newcommand{\flowRegressHardAteSd}{0.8}
\newcommand{\flowRegressHardMae}{6.77}
\newcommand{\flowRegressHardMaeSd}{0.32}
\newcommand{\flowRegressHardRmse}{10.44}
\newcommand{\flowRegressHardRmseSd}{0.41}
\newcommand{\flowRegressHardProf}{6.51}
\newcommand{\flowRegressHardProfSd}{0.18}
\newcommand{\flowSplitAte}{6.9}
\newcommand{\flowSplitAteSd}{0.5}

\newcommand{\flowSplitMae}{5.33}
\newcommand{\flowSplitMaeSd}{0.46}
\newcommand{\flowSplitRmse}{6.57}
\newcommand{\flowSplitRmseSd}{0.70}
\newcommand{\flowSplitProf}{5.23}
\newcommand{\flowSplitProfSd}{0.13}
\newcommand{\flowSplitHardAte}{7.5}
\newcommand{\flowSplitHardAteSd}{0.1}
\newcommand{\flowSplitHardMae}{5.60}
\newcommand{\flowSplitHardMaeSd}{0.41}
\newcommand{\flowSplitHardRmse}{7.00}
\newcommand{\flowSplitHardRmseSd}{0.72}
\newcommand{\flowSplitHardProf}{5.39}
\newcommand{\flowSplitHardProfSd}{0.13}
\newcommand{\flowAprimeAte}{7.6}
\newcommand{\flowAprimeAteSd}{0.5}

\newcommand{\flowAprimeMae}{5.38}
\newcommand{\flowAprimeMaeSd}{0.23}
\newcommand{\flowAprimeRmse}{6.77}
\newcommand{\flowAprimeRmseSd}{0.42}
\newcommand{\flowAprimeProf}{5.38}
\newcommand{\flowAprimeProfSd}{0.20}
\newcommand{\flowAprimeHardAte}{8.4}
\newcommand{\flowAprimeHardAteSd}{0.1}
\newcommand{\flowAprimeHardMae}{5.61}
\newcommand{\flowAprimeHardMaeSd}{0.10}
\newcommand{\flowAprimeHardRmse}{7.26}
\newcommand{\flowAprimeHardRmseSd}{0.41}
\newcommand{\flowAprimeHardProf}{5.72}
\newcommand{\flowAprimeHardProfSd}{0.13}
\newcommand{\flowNofieldAte}{10.1}
\newcommand{\flowNofieldAteSd}{0.6}

\newcommand{\flowNofieldMae}{6.98}
\newcommand{\flowNofieldMaeSd}{0.34}
\newcommand{\flowNofieldRmse}{9.00}
\newcommand{\flowNofieldRmseSd}{1.13}
\newcommand{\flowNofieldProf}{6.37}
\newcommand{\flowNofieldProfSd}{0.14}
\newcommand{\flowNofieldHardAte}{12.7}
\newcommand{\flowNofieldHardAteSd}{0.6}
\newcommand{\flowNofieldHardMae}{6.97}
\newcommand{\flowNofieldHardMaeSd}{0.31}
\newcommand{\flowNofieldHardRmse}{9.21}
\newcommand{\flowNofieldHardRmseSd}{1.08}
\newcommand{\flowNofieldHardProf}{6.34}
\newcommand{\flowNofieldHardProfSd}{0.31}
\newcommand{\flowNolatAte}{7.8}
\newcommand{\flowNolatAteSd}{0.8}

\newcommand{\flowNolatMae}{5.00}
\newcommand{\flowNolatMaeSd}{0.91}
\newcommand{\flowNolatRmse}{6.61}
\newcommand{\flowNolatRmseSd}{1.33}
\newcommand{\flowNolatProf}{5.57}
\newcommand{\flowNolatProfSd}{0.09}
\newcommand{\flowNolatHardAte}{8.7}
\newcommand{\flowNolatHardAteSd}{0.5}
\newcommand{\flowNolatHardMae}{5.28}
\newcommand{\flowNolatHardMaeSd}{0.59}
\newcommand{\flowNolatHardRmse}{7.53}
\newcommand{\flowNolatHardRmseSd}{1.20}
\newcommand{\flowNolatHardProf}{5.85}
\newcommand{\flowNolatHardProfSd}{0.25}

\newcommand{\ssGlobalTwelveAte}{6.6}
\newcommand{\ssGlobalTwelveAteSd}{0.2}
\newcommand{\ssGlobalTwelveMae}{8.48}
\newcommand{\ssGlobalTwelveMaeSd}{0.25}
\newcommand{\ssGlobalTwelveRmse}{14.77}
\newcommand{\ssGlobalTwelveRmseSd}{0.17}
\newcommand{\ssGlobalTwelveProf}{7.52}
\newcommand{\ssGlobalTwelveProfSd}{0.07}
\newcommand{\ssGlobalTwelveHardMae}{8.40}
\newcommand{\ssGlobalTwelveHardMaeSd}{0.10}
\newcommand{\ssGlobalTwelveHardRmse}{14.78}
\newcommand{\ssGlobalTwelveHardRmseSd}{0.01}
\newcommand{\ssGlobalTwentyFiveAte}{6.6}
\newcommand{\ssGlobalTwentyFiveAteSd}{0.7}
\newcommand{\ssGlobalTwentyFiveMae}{7.20}
\newcommand{\ssGlobalTwentyFiveMaeSd}{0.75}
\newcommand{\ssGlobalTwentyFiveRmse}{10.31}
\newcommand{\ssGlobalTwentyFiveRmseSd}{0.42}
\newcommand{\ssGlobalTwentyFiveProf}{6.53}
\newcommand{\ssGlobalTwentyFiveProfSd}{0.22}
\newcommand{\ssGlobalTwentyFiveHardMae}{7.23}
\newcommand{\ssGlobalTwentyFiveHardMaeSd}{0.53}
\newcommand{\ssGlobalTwentyFiveHardRmse}{10.33}
\newcommand{\ssGlobalTwentyFiveHardRmseSd}{0.46}
\newcommand{\ssGlobalFiftyAte}{6.6}
\newcommand{\ssGlobalFiftyAteSd}{0.2}
\newcommand{\ssGlobalFiftyMae}{6.44}
\newcommand{\ssGlobalFiftyMaeSd}{0.25}
\newcommand{\ssGlobalFiftyRmse}{9.32}
\newcommand{\ssGlobalFiftyRmseSd}{0.33}
\newcommand{\ssGlobalFiftyProf}{5.97}
\newcommand{\ssGlobalFiftyProfSd}{0.07}
\newcommand{\ssGlobalFiftyHardMae}{6.70}
\newcommand{\ssGlobalFiftyHardMaeSd}{0.45}
\newcommand{\ssGlobalFiftyHardRmse}{9.77}
\newcommand{\ssGlobalFiftyHardRmseSd}{0.62}

\newcommand{\ssCodebookTwelveAte}{6.7}
\newcommand{\ssCodebookTwelveAteSd}{0.4}
\newcommand{\ssCodebookTwelveMae}{7.53}
\newcommand{\ssCodebookTwelveMaeSd}{0.37}
\newcommand{\ssCodebookTwelveRmse}{12.99}
\newcommand{\ssCodebookTwelveRmseSd}{0.38}
\newcommand{\ssCodebookTwelveProf}{7.15}
\newcommand{\ssCodebookTwelveProfSd}{0.05}
\newcommand{\ssCodebookTwelveHardMae}{7.57}
\newcommand{\ssCodebookTwelveHardMaeSd}{0.31}
\newcommand{\ssCodebookTwelveHardRmse}{12.96}
\newcommand{\ssCodebookTwelveHardRmseSd}{0.48}
\newcommand{\ssCodebookTwentyFiveAte}{6.8}
\newcommand{\ssCodebookTwentyFiveAteSd}{0.1}
\newcommand{\ssCodebookTwentyFiveMae}{6.27}
\newcommand{\ssCodebookTwentyFiveMaeSd}{0.46}
\newcommand{\ssCodebookTwentyFiveRmse}{8.23}
\newcommand{\ssCodebookTwentyFiveRmseSd}{1.37}
\newcommand{\ssCodebookTwentyFiveProf}{6.19}
\newcommand{\ssCodebookTwentyFiveProfSd}{0.14}
\newcommand{\ssCodebookTwentyFiveHardMae}{6.28}
\newcommand{\ssCodebookTwentyFiveHardMaeSd}{0.49}
\newcommand{\ssCodebookTwentyFiveHardRmse}{8.07}
\newcommand{\ssCodebookTwentyFiveHardRmseSd}{1.34}
\newcommand{\ssCodebookFiftyAte}{6.6}
\newcommand{\ssCodebookFiftyAteSd}{0.6}
\newcommand{\ssCodebookFiftyMae}{5.23}
\newcommand{\ssCodebookFiftyMaeSd}{0.28}
\newcommand{\ssCodebookFiftyRmse}{6.89}
\newcommand{\ssCodebookFiftyRmseSd}{0.70}
\newcommand{\ssCodebookFiftyProf}{5.80}
\newcommand{\ssCodebookFiftyProfSd}{0.23}
\newcommand{\ssCodebookFiftyHardMae}{5.25}
\newcommand{\ssCodebookFiftyHardMaeSd}{0.48}
\newcommand{\ssCodebookFiftyHardRmse}{6.83}
\newcommand{\ssCodebookFiftyHardRmseSd}{1.01}

\newcommand{\ssEpisodesTwelve}{24}
\newcommand{\ssRecordsTwelve}{39}
\newcommand{\ssEpisodesTwentyFive}{50}
\newcommand{\ssRecordsTwentyFive}{74}
\newcommand{\ssEpisodesFifty}{100}
\newcommand{\ssRecordsFifty}{142}
\newcommand{\ssEpisodesHundred}{200}
\newcommand{\ssRecordsHundred}{295}

\newcommand{\pairHardMae}{5.02}
\newcommand{\pairHardMaeSd}{0.32}

\newcommand{\nopairMae}{4.89}
\newcommand{\nopairMaeSd}{0.37}

\newcommand{\nopairProf}{5.32}
\newcommand{\nopairProfSd}{0.06}
\newcommand{\nopairHandAte}{7.5}
\newcommand{\nopairHandAteSd}{0.5}

\newcommand{\nopairHardMae}{4.96}
\newcommand{\nopairHardMaeSd}{0.53}

\newcommand{\orgAnnotAte}{7.1}
\newcommand{\orgAnnotAteSd}{0.4}
\newcommand{\orgAnnotMae}{4.94}
\newcommand{\orgAnnotMaeSd}{0.24}
\newcommand{\orgAnnotRmse}{5.91}
\newcommand{\orgAnnotRmseSd}{0.27}

\newcommand{\orgAnnotProf}{5.15}
\newcommand{\orgAnnotProfSd}{0.07}
\newcommand{\orgLlmAte}{7.1}
\newcommand{\orgLlmAteSd}{0.5}
\newcommand{\orgLlmMae}{4.86}
\newcommand{\orgLlmMaeSd}{0.25}
\newcommand{\orgLlmRmse}{5.85}
\newcommand{\orgLlmRmseSd}{0.53}

\newcommand{\orgLlmProf}{5.00}
\newcommand{\orgLlmProfSd}{0.04}
\newcommand{\orgKmeansAte}{6.9}
\newcommand{\orgKmeansAteSd}{0.3}
\newcommand{\orgKmeansMae}{5.32}
\newcommand{\orgKmeansMaeSd}{0.34}
\newcommand{\orgKmeansRmse}{6.56}
\newcommand{\orgKmeansRmseSd}{0.21}

\newcommand{\orgKmeansProf}{5.25}
\newcommand{\orgKmeansProfSd}{0.05}

\newcommand{\tqOurs}{0.301}
\newcommand{\tqOursSd}{0.021}
\newcommand{\tqOursRmse}{0.384}
\newcommand{\tqOursRmseSd}{0.023}
\newcommand{\tqOursHard}{0.298}
\newcommand{\tqOursHardSd}{0.021}

\newcommand{\tqAte}{7.2}
\newcommand{\tqAteSd}{0.4}

\newcommand{\tqKtwenty}{0.285}
\newcommand{\tqKtwentySd}{0.026}
\newcommand{\tqKtwentyRmse}{0.376}
\newcommand{\tqKtwentyRmseSd}{0.041}
\newcommand{\tqKtwentyProf}{0.314}
\newcommand{\tqKtwentyProfSd}{0.006}
\newcommand{\tqKtwentyAte}{7.2}
\newcommand{\tqKtwentyAteSd}{0.2}
\newcommand{\tqOursProf}{0.315}
\newcommand{\tqOursProfSd}{0.003}

\newcommand{\tqGlobalProf}{0.332}
\newcommand{\tqGlobalProfSd}{0.007}

\newcommand{\tqGlobal}{0.283}
\newcommand{\tqGlobalSd}{0.008}
\newcommand{\tqGlobalRmse}{0.353}
\newcommand{\tqGlobalRmseSd}{0.011}

\newcommand{\tqAnnot}{0.256}
\newcommand{\tqAnnotSd}{0.010}
\newcommand{\tqAnnotRmse}{0.331}
\newcommand{\tqAnnotRmseSd}{0.013}
\newcommand{\tqAnnotProf}{0.305}
\newcommand{\tqAnnotProfSd}{0.006}
\newcommand{\tqLlm}{0.281}
\newcommand{\tqLlmSd}{0.016}
\newcommand{\tqLlmRmse}{0.375}
\newcommand{\tqLlmRmseSd}{0.015}
\newcommand{\tqLlmProf}{0.327}
\newcommand{\tqLlmProfSd}{0.005}
\newcommand{\tqKmeans}{0.308}
\newcommand{\tqKmeansSd}{0.012}
\newcommand{\tqKmeansRmse}{0.398}
\newcommand{\tqKmeansRmseSd}{0.030}
\newcommand{\tqKmeansProf}{0.334}
\newcommand{\tqKmeansProfSd}{0.008}

\newcommand{\kFiveAte}{6.7}
\newcommand{\kFiveAteSd}{0.7}
\newcommand{\kFiveMae}{4.83}
\newcommand{\kFiveMaeSd}{0.47}
\newcommand{\kFiveRmse}{6.12}
\newcommand{\kFiveRmseSd}{0.72}
\newcommand{\kFiveUsed}{5.0}
\newcommand{\kFiveUsedSd}{0.0}
\newcommand{\kFiveEff}{4.3}
\newcommand{\kFiveEffSd}{0.3}
\newcommand{\kFiveUsedPct}{100}
\newcommand{\kFiveUsedPctSd}{0}
\newcommand{\kTenAte}{6.6}
\newcommand{\kTenAteSd}{0.4}
\newcommand{\kTenMae}{4.81}
\newcommand{\kTenMaeSd}{0.38}
\newcommand{\kTenRmse}{5.92}
\newcommand{\kTenRmseSd}{0.42}

\newcommand{\kTenUsed}{10.0}
\newcommand{\kTenUsedSd}{0.0}
\newcommand{\kTenEff}{6.5}
\newcommand{\kTenEffSd}{0.4}
\newcommand{\kTenUsedPct}{100}
\newcommand{\kTenUsedPctSd}{0}
\newcommand{\kFifteenAte}{6.7}
\newcommand{\kFifteenAteSd}{0.9}
\newcommand{\kFifteenMae}{5.32}
\newcommand{\kFifteenMaeSd}{0.46}
\newcommand{\kFifteenRmse}{6.66}
\newcommand{\kFifteenRmseSd}{0.41}
\newcommand{\kFifteenUsed}{14.0}
\newcommand{\kFifteenUsedSd}{0.0}
\newcommand{\kFifteenEff}{7.7}
\newcommand{\kFifteenEffSd}{0.2}
\newcommand{\kFifteenUsedPct}{93}
\newcommand{\kFifteenUsedPctSd}{0}
\newcommand{\kTwentyAte}{6.5}
\newcommand{\kTwentyAteSd}{0.4}
\newcommand{\kTwentyMae}{4.90}
\newcommand{\kTwentyMaeSd}{0.09}
\newcommand{\kTwentyRmse}{6.13}
\newcommand{\kTwentyRmseSd}{0.26}

\newcommand{\kTwentyUsed}{16.3}
\newcommand{\kTwentyUsedSd}{1.5}
\newcommand{\kTwentyEff}{9.1}
\newcommand{\kTwentyEffSd}{0.9}
\newcommand{\kTwentyUsedPct}{82}
\newcommand{\kTwentyUsedPctSd}{8}
\newcommand{\kThirtyAte}{7.0}
\newcommand{\kThirtyAteSd}{0.3}
\newcommand{\kThirtyMae}{4.87}
\newcommand{\kThirtyMaeSd}{0.57}
\newcommand{\kThirtyRmse}{6.19}
\newcommand{\kThirtyRmseSd}{0.93}

\newcommand{\kThirtyUsed}{25.0}
\newcommand{\kThirtyUsedSd}{2.0}
\newcommand{\kThirtyEff}{12.3}
\newcommand{\kThirtyEffSd}{0.2}
\newcommand{\kThirtyUsedPct}{83}
\newcommand{\kThirtyUsedPctSd}{7}

\newcommand{\amiForceTrain}{0.61}
\newcommand{\amiCatTrain}{0.16}
\newcommand{\amiForceVal}{0.16}
\newcommand{\amiCatVal}{0.21}
\newcommand{\spearmanLevel}{0.67}

\newcommand{\oursHandRefProf}{7.88}
\newcommand{\oursHandRefProfSd}{0.17}
\newcommand{\nopairHandRefProf}{9.56}
\newcommand{\nopairHandRefProfSd}{1.45}
\newcommand{\fsLCropHandRefProf}{8.29}
\newcommand{\fsLCropHandRefProfSd}{0.05}
\newcommand{\fsLGripCropHandRefProf}{8.61}
\newcommand{\fsLGripCropHandRefProfSd}{0.03}
\newcommand{\fsZsCropHandRefProf}{9.33}

%% file: tables/rebuttal_guidance_macros.tex
\newcommand{\selfClickRawProf}{5.04}
\newcommand{\selfClickRawProfSd}{0.12}
\newcommand{\selfPromptProf}{5.56}
\newcommand{\selfPromptProfSd}{0.23}
\newcommand{\selfPromptSelfClickProf}{5.54}
\newcommand{\selfPromptSelfClickProfSd}{0.38}
\newcommand{\adeStay}{27.7}
\newcommand{\adeCatMean}{25.9}
\newcommand{\adeDrt}{23.7}
\newcommand{\adeCt}{20.3}
\newcommand{\adeAA}{30.4}
\newcommand{\adeDrtSelf}{27.4}
\newcommand{\adeCtSelf}{23.2}
\newcommand{\oursAde}{11.4}
\newcommand{\oursAdeSd}{1.1}
\newcommand{\selfClickRawAde}{17.0}
\newcommand{\selfClickRawAdeSd}{0.5}
\newcommand{\unseenAde}{11.0}
\newcommand{\unseenAdeSd}{0.6}
\newcommand{\seenAde}{12.0}
\newcommand{\seenAdeSd}{1.8}
\newcommand{\mixArithMae}{4.83}
\newcommand{\mixArithMaeSd}{0.39}
\newcommand{\mixArithRmse}{5.95}
\newcommand{\mixArithRmseSd}{0.46}

%% file: tables/sim_macros.tex
\newcommand{\simTzProfProgReach}{96\%}

%% file: tables/speed_section_macros.tex
\newcommand{\spdFigEvalRhoErrDur}{+0.13}
\newcommand{\spdFigEvalRhoErrDurP}{=0.39}

%% file: tables/speed_drawer_macros.tex
\newcommand{\spdDxAllBothRhoTravel}{+0.63}
\newcommand{\spdDxAllBothRhoTravelP}{<0.01}
\newcommand{\spdDxAllBothRhoRun}{-0.20}
\newcommand{\spdDxAllBothRhoRunP}{=0.06}

%% file: tables/speed_same_object_macros.tex
\newcommand{\spdSoDrawerOpenRhoRef}{+0.83}
\newcommand{\spdSoDrawerOpenRhoRefP}{=0.04}

%% file: tables/speed_force_executions_macros.tex
\newcommand{\spdFxCupboardClosePartialPeakSpeedSize}{+0.41}
\newcommand{\spdFxCupboardClosePartialPeakSpeedSizeP}{<0.01}

\newcommand{\spdFxDrawerOpenPartialPeakSpeedSize}{+0.42}
\newcommand{\spdFxDrawerOpenPartialPeakSpeedSizeP}{<0.01}

\newcommand{\spdFxRunIqrRatioMin}{2.0}
\newcommand{\spdFxRunIqrRatioMax}{2.8}

%% file: tables/rebuttal_category_input_macros.tex
\newcommand{\catWithheldLlm}{9.95}
\newcommand{\catWithheldLlmSd}{0.65}

\newcommand{\catWithheldAnnot}{8.16}
\newcommand{\catWithheldAnnotSd}{1.66}

\newcommand{\catWithheldKmeans}{7.08}
\newcommand{\catWithheldKmeansSd}{1.11}

%% file: tables/rebuttal_force_range_macros.tex
\newcommand{\frngNLight}{39}
\newcommand{\frngNHeavy}{9}

\newcommand{\frngLightCatMean}{4.60}
\newcommand{\frngHeavyCatMean}{12.53}
\newcommand{\frngOvenCatMean}{44.11}
\newcommand{\frngMaxCatMean}{45.25}
\newcommand{\frngOtherCatMean}{12.07}
\newcommand{\frngAllCatMean}{6.09}

\newcommand{\frngLightAnnot}{4.43}
\newcommand{\frngLightAnnotSd}{0.12}
\newcommand{\frngHeavyAnnot}{7.14}
\newcommand{\frngHeavyAnnotSd}{1.43}
\newcommand{\frngOvenAnnot}{59.19}
\newcommand{\frngOvenAnnotSd}{2.32}
\newcommand{\frngMaxAnnot}{62.94}
\newcommand{\frngMaxAnnotSd}{4.26}
\newcommand{\frngOtherAnnot}{3.42}
\newcommand{\frngOtherAnnotSd}{0.34}

\newcommand{\frngLightLlm}{4.72}
\newcommand{\frngLightLlmSd}{0.25}
\newcommand{\frngHeavyLlm}{5.48}
\newcommand{\frngHeavyLlmSd}{0.34}
\newcommand{\frngOvenLlm}{65.19}
\newcommand{\frngOvenLlmSd}{1.56}
\newcommand{\frngMaxLlm}{67.97}
\newcommand{\frngMaxLlmSd}{1.35}
\newcommand{\frngOtherLlm}{6.89}
\newcommand{\frngOtherLlmSd}{0.84}

\newcommand{\frngLightKmeans}{4.80}
\newcommand{\frngLightKmeansSd}{0.50}
\newcommand{\frngHeavyKmeans}{7.56}
\newcommand{\frngHeavyKmeansSd}{0.80}
\newcommand{\frngOvenKmeans}{59.45}
\newcommand{\frngOvenKmeansSd}{0.92}
\newcommand{\frngMaxKmeans}{62.18}
\newcommand{\frngMaxKmeansSd}{1.98}
\newcommand{\frngOtherKmeans}{5.06}
\newcommand{\frngOtherKmeansSd}{3.00}

\newcommand{\frngLightGlobal}{4.18}
\newcommand{\frngLightGlobalSd}{0.33}
\newcommand{\frngHeavyGlobal}{11.82}
\newcommand{\frngHeavyGlobalSd}{0.07}
\newcommand{\frngOvenGlobal}{47.11}
\newcommand{\frngOvenGlobalSd}{0.11}
\newcommand{\frngMaxGlobal}{47.51}
\newcommand{\frngMaxGlobalSd}{0.00}
\newcommand{\frngOtherGlobal}{2.57}
\newcommand{\frngOtherGlobalSd}{0.32}

\newcommand{\frngLightOurs}{4.24}
\newcommand{\frngLightOursSd}{0.45}
\newcommand{\frngHeavyOurs}{7.26}
\newcommand{\frngHeavyOursSd}{0.08}
\newcommand{\frngOvenOurs}{58.09}
\newcommand{\frngOvenOursSd}{2.32}
\newcommand{\frngMaxOurs}{61.67}
\newcommand{\frngMaxOursSd}{3.58}
\newcommand{\frngOtherOurs}{1.77}
\newcommand{\frngOtherOursSd}{0.33}

%% file: tables/prior_force_range_macros.tex
\newcommand{\frngLightMlp}{3.91}
\newcommand{\frngLightMlpSd}{0.19}
\newcommand{\frngHeavyMlp}{13.94}
\newcommand{\frngHeavyMlpSd}{0.15}
\newcommand{\frngOtherMlp}{2.95}
\newcommand{\frngOtherMlpSd}{1.14}
\newcommand{\frngLightNoroute}{4.53}
\newcommand{\frngLightNorouteSd}{0.32}
\newcommand{\frngHeavyNoroute}{10.49}
\newcommand{\frngHeavyNorouteSd}{0.60}
\newcommand{\frngOtherNoroute}{1.80}
\newcommand{\frngOtherNorouteSd}{0.21}

\newcommand{\stagePriorKone}{8.99}
\newcommand{\stagePriorKoneSd}{0.00}
\newcommand{\stageAdaptKone}{6.95}
\newcommand{\stageAdaptKoneSd}{0.29}
\newcommand{\stagePriorKten}{6.88}
\newcommand{\stagePriorKtenSd}{0.29}
\newcommand{\stageAdaptKten}{4.84}
\newcommand{\stageAdaptKtenSd}{0.31}
\newcommand{\frngNOther}{3}

%% file: tables/profile_scaled_macros.tex
\newif\ifprofscaledmarks\profscaledmarkstrue
\providecommand{\profscaled}[1]{\ifprofscaledmarks{\color{blue}#1}\else#1\fi}

\providecommand{\profscaledtable}{\ifprofscaledmarks\color{blue}\fi}
\renewcommand{\profileRmse}{\profscaled{5.52}}
\renewcommand{\profileRmseSd}{\profscaled{0.25}}
\renewcommand{\flowAHardProf}{\profscaled{5.87}}
\renewcommand{\flowAHardProfSd}{\profscaled{0.06}}
\renewcommand{\mechGlobalProf}{\profscaled{5.82}}
\renewcommand{\mechGlobalProfSd}{\profscaled{0.17}}
\renewcommand{\mechGlobalHardProf}{\profscaled{6.17}}
\renewcommand{\mechGlobalHardProfSd}{\profscaled{0.23}}
\renewcommand{\mechMlpProf}{\profscaled{5.77}}
\renewcommand{\mechMlpProfSd}{\profscaled{0.16}}
\renewcommand{\mechMlpHardProf}{\profscaled{6.02}}
\renewcommand{\mechMlpHardProfSd}{\profscaled{0.33}}
\renewcommand{\mechNorouteProf}{\profscaled{5.90}}
\renewcommand{\mechNorouteProfSd}{\profscaled{0.19}}
\renewcommand{\mechNorouteHardProf}{\profscaled{6.22}}
\renewcommand{\mechNorouteHardProfSd}{\profscaled{0.16}}
\renewcommand{\orgKmeansProf}{\profscaled{5.87}}
\renewcommand{\orgKmeansProfSd}{\profscaled{0.07}}
\renewcommand{\orgLlmProf}{\profscaled{5.92}}
\renewcommand{\orgLlmProfSd}{\profscaled{0.17}}
\renewcommand{\orgAnnotProf}{\profscaled{5.88}}
\renewcommand{\orgAnnotProfSd}{\profscaled{0.09}}
\renewcommand{\flowRegressProf}{\profscaled{7.29}}
\renewcommand{\flowRegressProfSd}{\profscaled{0.48}}
\renewcommand{\flowRegressHardProf}{\profscaled{7.28}}
\renewcommand{\flowRegressHardProfSd}{\profscaled{0.46}}
\renewcommand{\flowSplitProf}{\profscaled{5.57}}
\renewcommand{\flowSplitProfSd}{\profscaled{0.14}}
\renewcommand{\flowSplitHardProf}{\profscaled{5.97}}
\renewcommand{\flowSplitHardProfSd}{\profscaled{0.17}}
\renewcommand{\flowNofieldProf}{\profscaled{7.01}}
\renewcommand{\flowNofieldProfSd}{\profscaled{0.48}}
\renewcommand{\flowNofieldHardProf}{\profscaled{7.36}}
\renewcommand{\flowNofieldHardProfSd}{\profscaled{0.51}}
\renewcommand{\flowAprimeProf}{\profscaled{5.58}}
\renewcommand{\flowAprimeProfSd}{\profscaled{0.13}}
\renewcommand{\flowAprimeHardProf}{\profscaled{6.06}}
\renewcommand{\flowAprimeHardProfSd}{\profscaled{0.22}}
\renewcommand{\flowNolatProf}{\profscaled{5.82}}
\renewcommand{\flowNolatProfSd}{\profscaled{0.08}}
\renewcommand{\flowNolatHardProf}{\profscaled{6.42}}
\renewcommand{\flowNolatHardProfSd}{\profscaled{0.01}}
\renewcommand{\nopairProf}{\profscaled{5.73}}
\renewcommand{\nopairProfSd}{\profscaled{0.41}}
\renewcommand{\ssCodebookTwelveProf}{\profscaled{6.94}}
\renewcommand{\ssCodebookTwelveProfSd}{\profscaled{0.10}}
\renewcommand{\ssCodebookTwentyFiveProf}{\profscaled{6.54}}
\renewcommand{\ssCodebookTwentyFiveProfSd}{\profscaled{0.22}}
\renewcommand{\ssCodebookFiftyProf}{\profscaled{6.08}}
\renewcommand{\ssCodebookFiftyProfSd}{\profscaled{0.37}}
\renewcommand{\ssGlobalTwelveProf}{\profscaled{7.50}}
\renewcommand{\ssGlobalTwelveProfSd}{\profscaled{0.25}}
\renewcommand{\ssGlobalTwentyFiveProf}{\profscaled{6.91}}
\renewcommand{\ssGlobalTwentyFiveProfSd}{\profscaled{0.51}}
\renewcommand{\ssGlobalFiftyProf}{\profscaled{6.03}}
\renewcommand{\ssGlobalFiftyProfSd}{\profscaled{0.14}}
\renewcommand{\tqOursProf}{\profscaled{0.351}}
\renewcommand{\tqOursProfSd}{\profscaled{0.014}}
\renewcommand{\tqKtwentyProf}{\profscaled{0.382}}
\renewcommand{\tqKtwentyProfSd}{\profscaled{0.027}}
\renewcommand{\tqGlobalProf}{\profscaled{0.396}}
\renewcommand{\tqGlobalProfSd}{\profscaled{0.006}}
\renewcommand{\tqAnnotProf}{\profscaled{0.356}}
\renewcommand{\tqAnnotProfSd}{\profscaled{0.011}}
\renewcommand{\tqLlmProf}{\profscaled{0.389}}
\renewcommand{\tqLlmProfSd}{\profscaled{0.019}}
\renewcommand{\tqKmeansProf}{\profscaled{0.413}}
\renewcommand{\tqKmeansProfSd}{\profscaled{0.009}}
\renewcommand{\selfClickRawProf}{\profscaled{5.77}}
\renewcommand{\selfClickRawProfSd}{\profscaled{0.11}}
\renewcommand{\selfPromptProf}{\profscaled{5.72}}
\renewcommand{\selfPromptProfSd}{\profscaled{0.20}}
\renewcommand{\selfPromptSelfClickProf}{\profscaled{5.79}}
\renewcommand{\selfPromptSelfClickProfSd}{\profscaled{0.25}}
\renewcommand{\oursHandRefProf}{\profscaled{9.83}}
\renewcommand{\oursHandRefProfSd}{\profscaled{0.24}}
\renewcommand{\nopairHandRefProf}{\profscaled{14.22}}
\renewcommand{\nopairHandRefProfSd}{\profscaled{2.01}}
\newcommand{\profMaxRatio}{0.71}

%% file: sections/abstract.tex
People can anticipate how an object moves and the effort an interaction may require by watching others. How can embodied agents acquire this understanding from video? Force measurements provide direct sensory supervision, but require specialized equipment and remain limited in quantity. We introduce \method, which estimates an interaction trajectory together with a force profile from a third-person RGB video; 2D contact clicks and textual object--action \revc{prompts} are optional. Our approach makes limited instrumented measurements useful for interpreting human interactions through learned physical prototypes as a prior. The prototypes share learned force levels and temporal profiles across interactions of comparable effort, with visual and semantic \revf{cues} determining their contribution to each observation. Geometric and vision--language features adapt this prior to the observed interaction, while a joint conditional flow refines an initial tracked trajectory together with the force profile. Paired human and instrumented-gripper recordings provide measured force references for supervising human video, and a separately trained torque model demonstrates that the formulation extends to other sensory quantities. We evaluate trajectory recovery and force estimation through baseline comparisons and controlled ablations on held-out interactions from the Hoi! dataset. On this evaluation subset, \method\ achieves \obsSelectedATE\,cm ATE, peak-force MAE and RMSE of \obsSelectedMAE\,N and \obsSelectedRMSE\,N, respectively, and force-profile RMSE of \profileRmse\,N. On real-world videos, \method\ infers motion and force reliably and consistently despite the variability in human execution. We further deploy \method\ on a Spot robot, where trajectories and force profiles estimated from human demonstrations drive real-world articulated-object manipulation, and report success rates across repeated trials.~\textbf{Project Page:}~\url{https://rafaymhddn.github.io/vidforce}

%% file: sections/introduction.tex
\label{sec:intro}
Watching another person interact with an object reveals more than its motion. We can anticipate how it articulates and form an expectation of the effort involved, using observed actions to guide our own~\citep{Reichelt2013ActionObservation}. Giving robots a comparable understanding would make human video a source of both motion references and sensory expectations for planning and interaction. We study this capability through third-person video: estimating the trajectory of an articulated interaction together with its force or torque profile. \revr{On human video, this profile is what a gripper, the end effector of most robots, would need to apply in the same interaction, not the force exerted by the person.}

Recent methods extract rich geometric and motion representations from interaction videos. CoTracker3~\citep{Karaev2025CoTracker3} follows image points, D4RT~\citep{Zhang2025D4RT} reconstructs and tracks dynamic scenes, and EgoInfinity~\citep{Wang2026EgoInfinity} recovers hand--object motion. VidBot~\citep{Chen2025VidBot} and AFUN~\citep{Wang2026AFUN} predict interaction trajectories from an RGB-D observation and a language \revc{prompt}. While these capabilities capture how an interaction unfolds, they do not estimate the force or torque required to execute it. Supervising such estimates\revf{, however,} comes with two critical challenges: (i) force and torque data \revf{require} specialized measurement equipment\revf{,} and such instrumented interaction data \revf{are therefore far scarcer} than ordinary videos. (ii) Even with the limited available data, how can the extracted knowledge be transferred \revf{so} that models can predict from uninstrumented human \revf{demonstrations}?

We address this gap by transferring sensory information across interactions that require comparable effort. \method captures this structure with a learned set of force prototypes, each encoding a characteristic force level and temporal profile. Vision--language features route each observation to relevant prototypes, while observed geometry and appearance adapt the resulting prior to the observed interaction. By organizing these prototypes around measured effort rather than predefined object categories, the model connects limited instrumented supervision to uninstrumented human video without supervising or fixing any object taxonomy. To adapt the prototype to the observed instance, a joint conditional flow couples trajectory and force prediction. Given an initial trajectory estimate from a motion tracker, the flow refines this proposal and the adapted force prior. 
During training, force measurements from instrumented-gripper recordings provide reference targets for paired human videos. At inference, the contact point and \revc{prompt} may be supplied or inferred from the video. The same formulation can be extended to other sensory quantities; we demonstrate this with a separately trained torque model.

\begin{itemize}
\item We formulate sensory-aware articulation understanding from third-person video: jointly recovering interaction trajectory and a force profile, with optional contact and language guidance.
\item We introduce \method, a model that predicts the effort of an observed interaction in the form of force profiles. The model organizes observations by measured effort into learned force prototypes and jointly refines the resulting prior and a tracked trajectory. This design makes effective use of the limited available data without relying on predefined object categories.
\item We evaluate trajectory and sensory prediction against matched baselines and isolate the contributions of key model components and input sources through controlled ablations. We further assess video-only inference on \revb{real-world} recordings outside the training dataset and conduct success-rate trials on a Spot robot using the predicted trajectories and force profiles for manipulation.
\end{itemize}

%% file: sections/related_work.tex
\label{sec:related}
\noindent\textbf{Interaction and articulation from human video.}
Human video can supply motion cues for robot learning without robot action labels: point tracks condition policies~\citep{Wen2024ATM,Jain2024Vid2Robot}, reconstructed object motion supports imitation~\citep{Kerr2024RSRD}, and object-centric 3D motion fields provide metric geometry~\citep{Yin2025MotionField}. CoTracker3~\citep{Karaev2025CoTracker3} tracks points through occlusion, D4RT~\citep{Zhang2025D4RT} combines tracking with dynamic 3D reconstruction, EgoInfinity~\citep{Wang2026EgoInfinity} recovers metric hand--object motion, and Articulate-Anything~\citep{Le2025ArticulateAnything} constructs articulated simulation assets. Affordance models such as VidBot~\citep{Chen2025VidBot} and AFUN~\citep{Wang2026AFUN} use visual observations and language \revc{prompts} to predict interaction trajectories. These methods capture where and how an interaction unfolds, but not the required force or torque; \method builds on D4RT by initializing from its trajectory estimate and conditioning joint trajectory--force prediction on the full motion field.

\noindent\textbf{Force inference from video.}
\citet{Ehsani_2020_CVPR} infer force by matching simulated object motion to video, while ForceSight~\citep{Collins2023ForceSight} predicts a single robot-centric force goal from RGB-D and language. \revr{\citet{Li2019MotionForces} recover person and tool motion with contact forces from monocular video by optimizing full-body dynamics, for rigid tools without articulation, and EgoPHI~\citep{Ilic2026EgoPHI} predicts per-vertex contact and normal force from a single egocentric image and the object mesh, trained on simulated forces.} \method{} jointly predicts a temporal force profile and interaction trajectory \revr{of articulated objects} from third-person human video, using paired human and gripper recordings to transfer measured force supervision. Recent methods use measured force to transfer human demonstrations to robots: FeelTheForce~\citep{Adeniji2025FeelTheForce} learns contact-driven policies from demonstrations with hand-force measurements, and ReForce~\citep{Wu2026ReForce} retargets hand motion while accounting for contact forces. \method addresses the complementary problem of estimating this missing force signal from video, where robots can use the predicted force profile for force-aware control (Sec.~\ref{sec:realworld}).

\noindent\textbf{Sensory supervision.}
Measured sensory signals can ground visual representations in physical interaction. Touch and Go~\citep{Yang2022TouchAndGo} learns visual--tactile correspondence, TVL~\citep{Fu2024TVL} aligns touch with vision--language representations, FEEL~\citep{Dessalene2026FEEL} pairs egocentric video with measured hand forces for pretraining, and TouchAnything~\citep{Zhou2026TouchAnything} predicts bimanual pressure maps. Hoi!~\citep{Engelbracht_2026_CVPR} provides paired bare-hand and instrumented-gripper recordings across viewpoints; \method uses the gripper measurements to supervise gripper videos and provide force references for paired human interactions, then shares this supervision through learned prototypes organized by effort.

\noindent\textbf{Learned prototype and mixture priors.}
Mixture-of-experts models~\citep{Jacobs1991Mixtures,Shazeer2017MoE} route inputs to jointly learned specialists, discrete codebooks~\citep{Oord2017VQVAE} learn finite latent sets and serve as human-motion priors~\citep{Lucas2022PoseGPT}, and prototype networks~\citep{Snell2017Proto,Chen2019ProtoPNet} represent classes through learned exemplars. \method combines these ideas but organizes its prototypes by measured effort: a routing loss associates each training interaction with nearby force levels, while the levels, temporal shapes, and semantic keys are learned jointly with the router. Their log-force levels and temporal shapes are mixed softly, yielding a force prior without class labels or a reconstruction objective.

%% file: sections/method.tex
\begin{figure}[!tbp]
\centering
\includegraphics[width=0.78\textwidth]{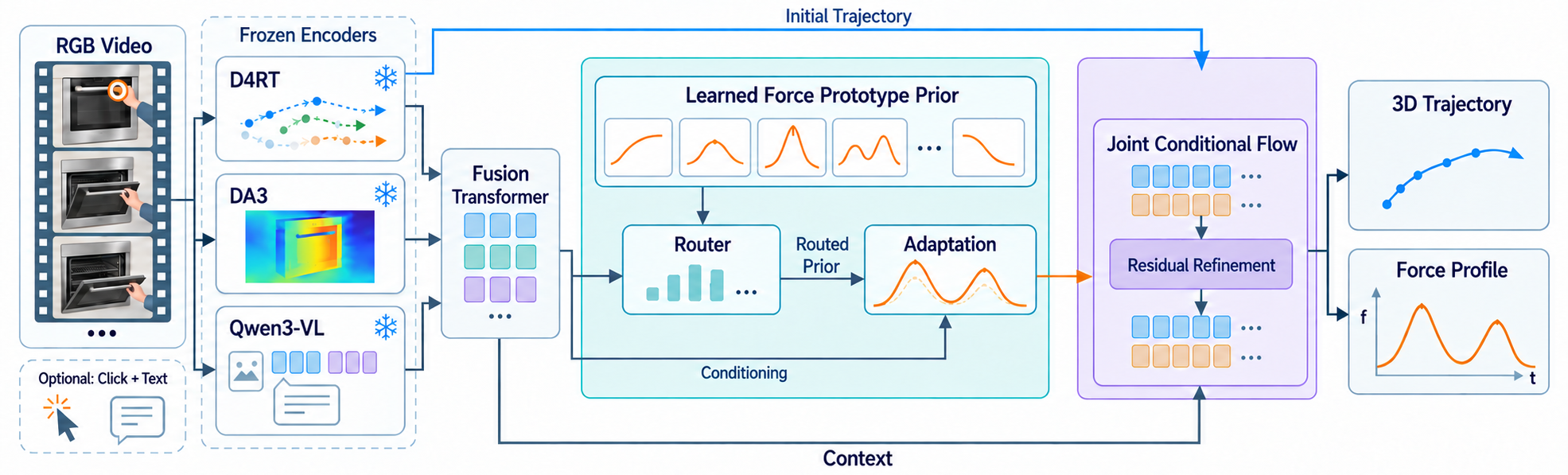}
\caption{\textbf{\method overview.} The model takes an RGB video, with an optional contact click and text describing the object and its motion. Frozen D4RT, DA3, and Qwen3-VL backbones extract motion, geometry, visual, and language features. A fusion transformer combines these features into a shared representation that routes and adapts learned force prototypes to form a video-specific force prior. A joint conditional flow refines this prior together with the initial tracked trajectory, producing the final 3D trajectory and force profile.}
\label{fig:overview}
\vspace{-3mm}
\end{figure}

\label{sec:method}
We present an overview of our method in Figure~\ref{fig:overview}. 
Given \revc{a video clip $V$, a sequence of 32 RGB frames showing one interaction from a static third-person camera and cropped to an object-centric window around the manipulated object}, \method predicts a camera-space contact trajectory $\mathbf P=(\mathbf p_i)_{i=1}^{N}$ at $N\rev{{=}16}$ waypoints, a force-magnitude profile $\mathbf f\in\mathbb R_{\geq0}^{L}$ over $L\rev{{=}32}$ samples of normalized interaction time, and a separate peak estimate $F$. A contact point and a text \revc{prompt} $q$ of the object and its motion are optional inputs. Visual and language features route the observation to learned force prototypes, which initialize the force prediction. A joint conditional flow then refines this prediction together with the tracked trajectory. The same formulation applies to other sensory channels; a separately trained model predicts sensor-frame torque.

\subsection{Representing the observed interaction}
\label{sec:representation}
Three frozen models extract complementary information (motion, depth, semantics) from the clip and atrainable transformer fuses these representations, while the tracks provide an initial trajectory proposal. D4RT tracks points throughout the clip to form a motion field. The median displacement of valid tracks near the contact defines the initial trajectory proposal $\mathbf B$. DA3~\citep{Lin2025DA3} estimates depth and camera intrinsics $\mathbf K$. Qwen3-VL~\citep{Bai2025Qwen3VL} provides visual features and the \revc{prompt} embedding $\mathbf e(q)$. \rev{Rather than a generic text embedding, $\mathbf e(q)$ contains the square-rooted probabilities assigned to 256 answer tokens and the two actions when Qwen3-VL is asked which part is manipulated and whether it opens or closes (Appendix~\ref{app:impl:repr}).} The fusion transformer reads these features together with track reliability, motion statistics\rev{, and an embodiment flag (human or gripper)}, and produces context tokens $\mathbf M$ and pooled features $\mathbf g$.

\textbf{Coordinates.}
Each waypoint $\mathbf p_i$ is a 3D point in the camera frame. It is encoded \rev{as $\mathbf x_i$} by its normalized image offset from the contact and its log depth relative to the contact depth $z_c$ (Eq.~\ref{eq:state}, Appendix~\ref{app:impl:repr}). This places the \rev{encoded} trajectory $\mathbf X\rev{{=}(\mathbf x_i)_{i=1}^{N}}\in\mathbb R^{N\times3}$, tracked proposal $\mathbf B$, and predicted residuals in the same scale-free coordinate frame. The encoding is invertible given $\mathbf K$, so metric motion can be recovered exactly\rev{, at the scale estimated by DA3}.

\subsection{A learned physical prototype prior}
\label{sec:prior}
The sensory prior consists of learned force prototypes that capture force levels and temporal profiles shared across interactions of comparable effort. A router weights these prototypes using visual and language features, and an adaptation head adjusts the resulting prior to the observed interaction. The prior can be of arbitrary size, and we choose $K\rev{{=}10}$ prototypes for all our experiments as the largest $K$ for which every prototype remains in use (Appendix~\ref{app:ksweep}). Each prototype comprises a log-force level $a_k$, a shape $\mathbf s_k\in\mathbb R^L$, and a learned semantic key $\mathbf e_k$. The level describes how much force is expected, the shape describes how it varies through the interaction, and the key is matched against language features during routing. We normalize the force profile by the separately predicted peak, $\boldsymbol\sigma=\mathbf f/F$, separating magnitude and temporal shape.

\textbf{Routing.}
For prototype $k$, the router combines a learned score from the pooled video features with the similarity of \revc{prompt} embedding $\mathbf e(q)$ and semantic key $\mathbf e_k$: $\ell_k=r_k(\mathbf g)+\beta\cos(\mathbf e(q),\mathbf e_k)$, where $\beta$ is learned. A softmax over the scores mixes the prototypes into the prior peak and shape:
\begin{equation}
\boldsymbol\pi=\operatorname{softmax}(\boldsymbol\ell),\quad
\log\bar F=\sum_k\pi_k a_k,\quad
\bar{\boldsymbol\sigma}=\bigl[\textstyle\sum_k\pi_k\mathbf s_k\bigr]_+ .
\label{eq:prior}
\end{equation}
Mixing peaks in the log domain makes partial weights interpolate force levels geometrically rather than additively. The operator $[\cdot]_+$ keeps the mixed shape nonnegative. We initialize the levels from quantiles of the training peaks, the shapes from the mean training profile, and the keys randomly. All prototype parameters are then learned jointly with the router as a mixture of experts~\citep{Jacobs1991Mixtures}. \rev{The \emph{global prior} is the median measured peak and mean normalized profile over the training \revc{sequences}. It randomly replaces the routed prior for 30\% of training \revc{sequences} as prior dropout and serves as the no-prototype control in Sec.~\ref{sec:exp}.}

\textbf{Routing loss.}
During training, prototypes whose levels are closer to the target peak receive larger target weights. Let $\delta_k=(\log\max(F,\epsilon)-\sg(a_k))/\tau_r$ be the scaled distance to prototype $k$. These distances define target weights $\rho_k(F)$ and a cross-entropy loss that trains the predicted weights $\pi_k$:
\begin{equation}
\boldsymbol\rho(F)=\operatorname{softmax}_k(-\delta_k^2/2),\qquad
\mathcal L_{\mathrm{route}}=-\sum_k\rho_k(F)\log\pi_k .
\label{eq:route-loss}
\end{equation}
Through stop-gradient, we hold prototype levels fixed when forming these targets. The width $\tau_r$ controls how broadly credit is shared among nearby levels. Prototype values learn through sensory prediction, while this loss teaches the router which prototypes to favor. \rev{Together, these losses let the prototypes self-organize around measured effort rather than object categories.} At inference, $\boldsymbol\pi$ is computed from the video features and, when supplied, the \revc{prompt}; no measurement enters.

\textbf{Instance adaptation.}
An adaptation head adjusts the shared prior to the observed interaction by predicting a scalar log-peak offset $d(\mathbf g)$ and a shape offset $\boldsymbol\eta(\mathbf g)$. The adapted estimates are $\bar F'=\bar F\exp(c\tanh(d(\mathbf g)/c))$ and $\bar{\boldsymbol\sigma}'=[\bar{\boldsymbol\sigma}+\boldsymbol\eta(\mathbf g)]_+$, where $c=\ln 2$ bounds the peak multiplier to a factor of two. A log-peak Huber loss and a shape squared-error loss jointly train the prototypes and adaptation head.

\subsection{Joint conditional flow for motion and effort}
\label{sec:flow}
A joint conditional flow predicts residual corrections to the trajectory proposal and adapted force prior, allowing motion and effort to exchange information during refinement. The resulting residual is decoded into the final trajectory, peak force, and force profile\profscaled{, scaled so that its maximum equals the peak}. The correction has three parts---waypoint offsets from the tracked trajectory, a log-scale change in peak force, and changes to the temporal shape---concatenated into one residual state:
\begin{equation}
\mathbf r=\left[\operatorname{vec}(\mathbf X-\mathbf B),\;
\log\frac{\max(F,\epsilon)}{\bar F'},\;
\boldsymbol\sigma-\bar{\boldsymbol\sigma}'\right]\in\mathbb R^{3N+1+L}.
\label{eq:joint-state}
\end{equation}
The operator $\operatorname{vec}$ stacks the waypoint offsets into a vector. \rev{The adapted prior enters the target with gradients detached, so the flow learns residuals around a fixed reference and the prior is shaped only by its own loss.} A DiT-style transformer velocity field~\citep{Peebles2023DiT} processes the residual tokens jointly. It attends to $\mathbf M$ and is conditioned on $\mathbf g$, the tracked proposal, and motion-derived features. \rev{Waypoint tokens receive tracked motion around the contact, while force tokens receive speed, acceleration, and displacement statistics.} This joint conditioning lets motion and effort inform one another without imposing a force--acceleration relation.

We train it with flow matching~\citep{Lipman2023FlowMatching} along the straight paths of rectified flow~\citep{Liu2023RectifiedFlow}: with $\widetilde{\mathbf r}$ the target \rev{standardized per coordinate}, $\boldsymbol\epsilon\sim\mathcal N(\mathbf0,\mathbf I)$, and flow time $t\sim\mathcal U[0,1]$ (distinct from interaction time),
\begin{equation}
\mathbf y_t=(1-t)\boldsymbol\epsilon+t\widetilde{\mathbf r},\qquad
\mathcal L_{\mathrm{flow}}=
\mathbb E\!\left[\left\|v_\theta(\mathbf y_t,t;\mathcal C)
 -(\widetilde{\mathbf r}-\boldsymbol\epsilon)\right\|^2_{\Omega}\right],
\label{eq:flow}
\end{equation}
where $\mathcal C$ is the video and motion context and $\Omega$ masks missing targets and balances the motion and sensory terms (Appendix~\ref{app:impl:flow}).

At inference, 20 Euler steps from $\mathbf y_0=\mathbf0$ produce the residual. We add this residual to the trajectory proposal and adapted prior, clipping the log-peak change to $\pm\rev{b}$\rev{, where $b=0.5$}. Inverse projection then recovers metric motion. \rev{Integration from the origin yields a single deterministic estimate; starting from sampled noise instead leaves the force errors unchanged within seed variation and worsens ATE by 0.2\,cm, and averaging five noise draws matches the zero start (Appendix~\ref{app:flowstart}).}

\subsection{Training and inference}
\label{sec:supervision}
Training transfers the few measured forces to human video through paired recordings, and inference needs only the clip, with the contact and the \revc{prompt} supplied by a human or a VLM.

\textbf{Training.}
Hoi!~\citep{Engelbracht_2026_CVPR} supplies \rev{tracked} trajectories for human and gripper recordings \rev{(Sec.~\ref{sec:exp:setup})} and force measurements for the gripper. A human recording inherits the peak and normalized profile of the paired gripper recording of the same episode, possibly from another camera, as a down-weighted reference \revr{for the corresponding gripper force a robot would need}. \revr{This assumes that the shared object and action mainly set the required force; differences in speed and contact location between gripper and human video pairs are not modeled.} The training loss is
\begin{equation}
\mathcal L=\mathcal L_{\mathrm{flow}}+\lambda_{\mathrm{peak}}\mathcal L_{\mathrm{peak}}+\lambda_{\mathrm{prior}}\mathcal L_{\mathrm{prior}}+\lambda_{\mathrm{route}}\mathcal L_{\mathrm{route}}+\lambda_{\mathrm{cons}}\mathcal L_{\mathrm{cons}}+\mathcal L_{\mathrm{aux}},
\label{eq:loss}
\end{equation}
Here, $\mathcal L_{\mathrm{flow}}$ is defined in Eq.~\ref{eq:flow}, $\mathcal L_{\mathrm{peak}}$ is a Huber loss in newtons on the peak decoded in one step from the current flow state, and $\mathcal L_{\mathrm{prior}}$ combines a log-peak Huber loss with shape squared error on the adapted prior. The routing loss $\mathcal L_{\mathrm{route}}$ is defined in Eq.~\ref{eq:route-loss}, $\mathcal L_{\mathrm{cons}}$ is a cross-entropy that aligns the language-key scores with the detached video-based routing, and $\mathcal L_{\mathrm{aux}}$ contains auxiliary scale, peak-count, and contact-localization terms. We set $\lambda_{\mathrm{prior}}=1$ and the other weights to $0.5$ (Appendix~\ref{app:impl:train}). Missing targets are masked. The few force targets are fit long before the trajectory converges. Once the force error has leveled off, a frozen exponential-moving-average copy of the model becomes the source of all force targets. This holds the force function fixed while trajectory training continues, allowing one checkpoint to serve both outputs (Appendix~\ref{app:impl:train}).

\textbf{Inference.}
Inference requires no force or torque measurements. \rev{The model takes a clip spanning one interaction; untrimmed recordings are delimited to the span in which the object moves, following the data pipeline of Sec.~\ref{sec:exp:setup} (Appendix~\ref{app:realworld}).} A \revc{manual click} centers the \revc{window} on the contact. Without a click, the vision--language model proposes the contact, optionally refined by a localization head. The motion field is then re-anchored on this proposal in a second pass (\emph{self-click}). The object--action \revc{prompt} can likewise be supplied or written by the same model from the frames (\emph{self-prompt}); the \revc{questions} are given in Appendix~\ref{app:impl:inference}. The torque model follows the same procedure with sensor-frame torque targets and its own prototypes.

%% file: sections/experiments.tex
\label{sec:exp}
We evaluate trajectory recovery, force and torque estimation, and the contribution of the learned prototype prior and the joint flow (Sec.~\ref{sec:exp:results}). The prior study also tests limited sensory supervision (Table~\ref{tab:sensory}); separate analyses examine paired transfer to human video and input guidance \reve{(Sec.~\ref{sec:exp:results})}. \revb{Real-world} videos and a robot deployment examine predictions beyond the recorded dataset (Sec.~\ref{sec:realworld}); the data, protocol, and baselines are described first (Sec.~\ref{sec:exp:setup}).

\subsection{Experiment Setup}
\label{sec:exp:setup}
\textbf{Data.}
We train and evaluate on Hoi!~\citep{Engelbracht_2026_CVPR}, which contains paired human and instrumented-gripper recordings of everyday articulated-object interactions viewed from two static cameras. Gripper recordings provide measured force and torque, while paired human recordings support transfer to bare-hand video (Sec.~\ref{sec:supervision}). 
Each sequence represents one interaction from one camera and one embodiment, sampled to 32 frames, \revc{cropped to the model's object window (Sec.~\ref{sec:method})}, and annotated with the first-frame contact and \revb{the dataset prompt}. Because occlusions can hide the contact or articulated motion, we verify contact visibility by a depth-ray test, and check whether at least half of the motion is in view, and the trajectory plausible.
Using these checks, we retain \nRecords{} sequences (\nGripper{} gripper and \nHuman{} human).
Appendix~\ref{app:data} details the pipeline and rejected examples.

\textbf{Training.}
We train on a train split of \nTrain{} sequences (\nTrainGripper{} gripper, \nTrainHuman{} human), increasing data diversity through shifted crops, rescaling around the object, and flipping (Appendix~\ref{app:aug}). Sequences with measured force targets have weight 1 and paired references weight 0.5 in the peak and shape terms. \reve{On 30\,\% of the training recordings, the contact is perturbed and the click-seeded tracks are dropped (click-free rows), so that the model also works from a proposed contact.} More details can be found in Appendix~\ref{app:aug} and \ref{app:impl:prior}.

\textbf{Evaluation.}
The evaluation set (\textbf{Eval}) contains \nVal{} sequences (\nValForce{} gripper, 51 human) with a \revc{\emph{manual click} and a \emph{manual prompt}}. It combines \nUnseen{} sequences (\nUnseenForce{} gripper) from an \emph{unseen room type}---the only living room among the 20 Hoi! scenes---and \nSeen{} sequences (\nSeenForce{} gripper) from interactions with \emph{held-out objects} in ten training scenes. \revb{\emph{Self-click} and \emph{self-prompt} (Sec.~\ref{sec:supervision}) replace the \revc{\emph{manual click} and \emph{manual prompt}} with Qwen3-VL proposals using the same checkpoints. \revr{Both manual-click and self-click runs use depth predicted by DA3. Manual-click runs sample this depth at the annotated contact pixel, whereas self-click runs sample it at the contact pixel predicted by Qwen.} Tracker baselines receive the same proposed point (Appendix~\ref{app:evalprotocol}).} \revb{Unless stated otherwise, every evaluation uses three training seeds and reports mean $\pm$ sample SD at the final checkpoint; deterministic trajectory baselines are run once.}

\begin{figure}[!t]
\centering
\includegraphics[width=0.8\textwidth]{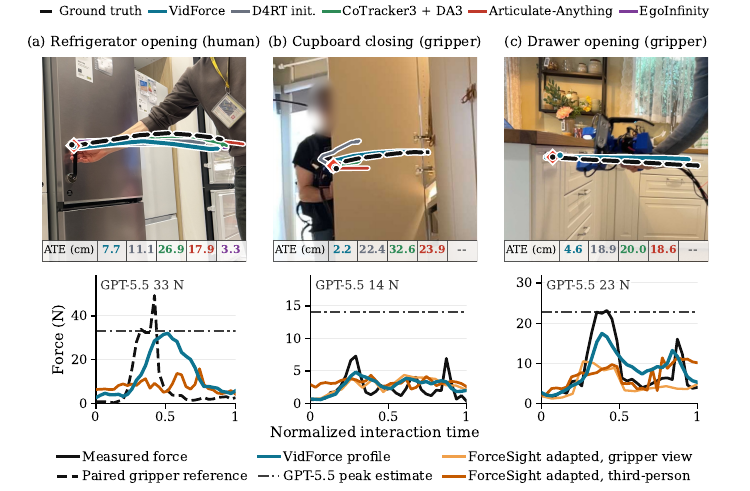}
\vspace{-3mm}
\caption{\textbf{Trajectory and force predictions.} Top: ground-truth and predicted contact trajectories in the object-centric window. All trajectory methods receive the manual click \symclick{}, and the lower strip reports each method's ATE in cm using its line color. Bottom: predicted force profiles and peak estimates against measured force; the human sequence uses its paired gripper reference.}
\label{fig:qual}
\vspace{-3mm}
\end{figure}

\textbf{Metrics.}
ATE (cm) is the mean waypoint error per sequence after translation-only alignment of the predicted and measured starts, reported as the median across sequences. \revp{ADE is the same error without the alignment.} Peak MAE and RMSE compare one predicted peak with one measured peak per instrumented sequence. Profile RMSE is \revb{the RMS error} over the 32 normalized-time samples \profscaled{of the profile scaled to the predicted peak}\revb{,} aggregated by RMS across sequences. Force errors are reported in N and torque errors in N\,m.

\textbf{Baselines.}
\revb{No prior method predicts trajectory and force \revr{profiles of articulated-object interactions from third-person video (Sec.~\ref{sec:related})}, so we use separate baselines.}
All trajectory baselines receive the same clip and contact and are evaluated at the same 16 waypoints. D4RT provides the tracked initialization without learned refinement. CoTracker3 tracks the contact pixel and lifts it to 3D using DA3 depth. Articulate-Anything~\citep{Le2025ArticulateAnything} reconstructs an articulated asset from the video; we register it to the contact and CoTracker3 track, then animate its predicted joint range at constant speed. EgoInfinity~\citep{Wang2026EgoInfinity} reconstructs the hand and uses the joint nearest the contact; we evaluate it only on human interactions.\\
For force, GPT-5.5~\citep{OpenAI2026GPT55} estimates the peak from eight frames, with or without object--action text. ForceSight~\citep{Collins2023ForceSight} predicts the force a robot should apply from one RGB-D frame and a \revc{prompt}; its released model was trained on a robot wrist camera. We evaluate it in the Hoi! gripper-mounted view---closest to its training setting%
---and in our third-person view. We test the released checkpoint zero-shot, a fine-tuned version on our data, and an adapted variant that replaces the force-goal output with a per-frame force prediction. 
Appendix~\ref{app:baselines} details all baselines.

\subsection{Results and Discussion}
\label{sec:exp:results}
\textbf{Trajectory estimation.}
\method{} achieves the lowest ATE among all baselines, reducing the error from \ateDrt\,cm for its D4RT initialization to $\oursAte\pm\oursAteSd$\,cm (Table~\ref{tab:trajectory}). Both video-free references and Articulate-Anything remain substantially worse, showing that neither category-level motion nor a registered articulation alone recovers the observed instance trajectory.

\begin{table}[!t]
\centering\setlength{\abovecaptionskip}{3pt}
\begin{minipage}[t]{0.485\textwidth}
\vspace{0pt}\centering\footnotesize
\begin{minipage}[t][37pt][t]{\linewidth}
\caption{\revb{\textbf{Trajectory estimation.} ATE \revp{and ADE} for manual and Qwen3-VL contacts. \method{} is best for both embodiments.}}
\label{tab:trajectory}
\end{minipage}\par\vspace{0pt}
\fontsize{6}{7}\selectfont\setlength{\tabcolsep}{1pt}\setlength{\medmuskip}{0mu}\renewcommand{\arraystretch}{0.9333}
\newcommand{\trajrow}{\leftskip=0.5em\baselineskip=\arraystretch\baselineskip\relax}
\begin{tabular*}{\linewidth}{@{}p{54pt}@{\extracolsep{\fill}}cccc@{}}
\toprule
ATE (cm) $\downarrow$ & All (\nVal) & Human (\nValHand) & Gripper (\nValGripper) & \revp{ADE All (\nVal)} \\
\midrule
\multicolumn{5}{@{}l}{\emph{Manual click}} \\
\trajrow Stay at contact (ref.) & \ateStay & \ateStayHuman & \ateStayGripper & \revp{\adeStay} \\
\trajrow Category mean (ref.) & \ateCatMean & \ateCatMeanHuman & \ateCatMeanGripper & \revp{\adeCatMean} \\
\trajrow D4RT init. & \underline{\ateDrt} & \underline{\ateDrtHuman} & \underline{\ateDrtGripper} & \revp{\adeDrt} \\
\trajrow CoTracker3 + DA3 & \ateCt & \ateCtHuman & \ateCtGripper & \revp{\underline{\adeCt}} \\
\trajrow Articulate-Anything\newline(adapted) & \ateAA & \ateAAHuman & \ateAAGripper & \revp{\adeAA} \\
\trajrow EgoInfinity (hands) & -- & \ateEgoHuman & -- & \revp{--} \\
\midrule
\trajrow \method & {\boldmath$\oursAte\pm\oursAteSd$} & {\boldmath$\oursHandAte\pm\oursHandAteSd$} & {\boldmath$\oursGripperAte\pm\oursGripperAteSd$} & \revp{\boldmath$\oursAde\pm\oursAdeSd$} \\
\midrule
\multicolumn{5}{@{}l}{\emph{Self-click (\revb{Qwen3-VL-proposed} contact)}} \\
\trajrow D4RT initialization & \underline{\ateDrtSelf} & \underline{\ateDrtSelfHuman} & \ateDrtSelfGripper & \revp{\adeDrtSelf} \\
\trajrow CoTracker3 + DA3\newline(per-frame depth) & \ateCtSelf & \ateCtSelfHuman & \underline{\ateCtSelfGripper} & \revp{\underline{\adeCtSelf}} \\
\midrule
\trajrow \method & {\boldmath$\revr{\selfClickRawAte\pm\selfClickRawAteSd}$} & {\boldmath$\revr{\selfClickRawHandAte\pm\selfClickRawHandAteSd}$} & {\boldmath$\revr{\selfClickRawGripperAte\pm\selfClickRawGripperAteSd}$} & {\boldmath$\revp{\selfClickRawAde\pm\selfClickRawAdeSd}$} \\
\bottomrule
\end{tabular*}
\end{minipage}\hfill
\begin{minipage}[t]{0.485\textwidth}
\vspace{0pt}\centering\footnotesize
\begin{minipage}[t][37pt][t]{\linewidth}
\caption{\textbf{Force and torque estimation.} GPT-5.5 uses eight frames. $^\dagger$Gripper-view ForceSight is scored once per episode.}
\label{tab:force}
\end{minipage}\par\vspace{0pt}
\fontsize{6}{7}\selectfont\setlength{\tabcolsep}{1pt}\setlength{\medmuskip}{0mu}\renewcommand{\arraystretch}{0.9333}
\begin{tabular*}{\linewidth}{@{}l@{\extracolsep{\fill}}ccc@{}}
\toprule
Error $\downarrow$ & Peak MAE & Peak RMSE & Profile RMSE \\
\midrule
\multicolumn{4}{@{}l}{\emph{Force (N)}} \\
\hspace{0.5em}GPT-5.5 & \underline{\maeGptImg} & \rmseGptImg & -- \\
\hspace{0.5em}GPT-5.5 + text & \maeGpt & \rmseGpt & -- \\
\multicolumn{4}{@{}l}{\hspace{0.5em}\emph{ForceSight: gripper view}\smash{$^\dagger$}} \\
\hspace{1em}Zero-shot & \fsZsGripViewMae & \fsZsGripViewRmse & -- \\
\hspace{1em}Adapted & $\fsLGripViewMae\pm\fsLGripViewMaeSd$ & $\underline{\fsLGripViewRmse\pm\fsLGripViewRmseSd}$ & {\boldmath$\fsLGripViewProf\pm\fsLGripViewProfSd$} \\
\multicolumn{4}{@{}l}{\hspace{0.5em}\emph{ForceSight: third-person view}} \\
\hspace{1em}Zero-shot & \fsZsCropMae & \fsZsCropRmse & -- \\
\hspace{1em}Adapted & $\fsLCropMae\pm\fsLCropMaeSd$ & $\fsLCropRmse\pm\fsLCropRmseSd$ & $\fsLCropProf\pm\fsLCropProfSd$ \\
\midrule
\hspace{0.5em}\method & {\boldmath$\oursMae\pm\oursMaeSd$} & {\boldmath$\oursRmse\pm\oursRmseSd$} & $\underline{\profileRmse\pm\profileRmseSd}$ \\
\midrule
\multicolumn{4}{@{}l}{\emph{Sensor-frame torque (N\,m)}} \\
\hspace{0.5em}GPT-5.5 & \tqGptImg & \tqGptImgRmse & -- \\
\hspace{0.5em}GPT-5.5 + text & \underline{\tqGpt} & \underline{\tqGptRmse} & -- \\
\midrule
\hspace{0.5em}\method & {\boldmath$\tqOurs\pm\tqOursSd$} & {\boldmath$\tqOursRmse\pm\tqOursRmseSd$} & {\boldmath$\tqOursProf\pm\tqOursProfSd$} \\
\bottomrule
\end{tabular*}
\end{minipage}
\vspace{-13.1pt}
\end{table}

The largest gains occur when contact-seeded tracking fails to follow the object, particularly under gripper occlusion. The same ordering holds for human and gripper recordings and for Qwen3-VL-proposed contacts (Table~\ref{tab:trajectory}).
The examples in Fig.~\ref{fig:qual} illustrate this pattern. \reve{EgoInfinity follows the visible hand during the refrigerator opening (a),} but the trackers and registered asset miss the cupboard motion (b), and gripper occlusion leaves the drawer baselines near their starting point (c). \method{} recovers the object motion in both cases.

\textbf{Force and torque estimation.}
\method{} achieves the lowest peak and profile errors among the third-person force methods (Table~\ref{tab:force}), reducing peak MAE from \maeGptImg\,N for GPT-5.5 to $\oursMae\pm\oursMaeSd$\,N. Errors remain concentrated in the heaviest interactions, which the model tends to underestimate.
ForceSight is strongly affected by camera viewpoint. Its released model predicts nearly constant low force goals. The adapted variant calibrates the force scale and adds a profile, but remains less accurate than \method{} in the third-person view (Table~\ref{tab:force}; Fig.~\ref{fig:qual}; Appendix~\ref{app:baselines}).
Table~\ref{tab:force} also evaluates sensor-frame torque, validating the applicability of our general approach to other modalities.

\label{sec:exp:ablations}

\textbf{\revf{How data-efficient is \method{}?}}
Force data is expensive to acquire\revp{, so \method{} has to learn from limited sensory data}. We test data-efficiency by retraining the learned prototypes and the global prior by partially withholding measurements from training episodes, stratified by object and action; withheld episodes keep their videos and motion labels but contribute no force targets, paired references, or prior statistics (Table~\ref{tab:sensory}). We observe that even with a quarter of the measurement data, \method{} still outperforms all baselines of Table~\ref{tab:force}. \revp{The learned prior provides a good initialization (Table~\ref{tab:sensory}, right), which lowers the burden on the flow for refinement. With half of the force labels, it beats the global prior with all of them, for five or ten prototypes alike (Table~\ref{tab:sensory-by-k}).}

\begin{table}[!t]
\centering
\setlength{\abovecaptionskip}{3pt}\setlength{\belowcaptionskip}{1pt}
\caption{\textbf{Limited force supervision.} Peak MAE (N) with force labels on a share of the training episodes (left), and after each stage at 100\,\% supervision (right). Full metrics: Table~\ref{tab:sensory-full}.}
\label{tab:sensory}
\scalebox{\abltabscale}{\begin{minipage}{1.166667\linewidth}\fontsize{7}{8.166667}\selectfont\setlength{\tabcolsep}{3pt}\setlength{\medmuskip}{0mu}
\begin{tabular*}{\linewidth}{@{}l@{\extracolsep{\fill}}cccccc@{}}
\toprule
 & \multicolumn{3}{c}{Force-label availability} & \multicolumn{3}{c}{\revp{Stages at 100\,\% supervision}} \\
\cmidrule(lr){2-4}\cmidrule(l){5-7}
Prior & 12\,\% & 25\,\% & 50\,\% & \revp{Prior} & \revp{+ Adaptation} & \revp{+ Flow} \\
\midrule
Global prior \revp{($K{=}1$)} & $\ssGlobalTwelveMae{\scriptscriptstyle\pm\ssGlobalTwelveMaeSd}$ & $\ssGlobalTwentyFiveMae{\scriptscriptstyle\pm\ssGlobalTwentyFiveMaeSd}$ & $\ssGlobalFiftyMae{\scriptscriptstyle\pm\ssGlobalFiftyMaeSd}$ & \revp{$\stagePriorKone{\scriptscriptstyle\pm\stagePriorKoneSd}$} & \revp{$\stageAdaptKone{\scriptscriptstyle\pm\stageAdaptKoneSd}$} & $\mechGlobalMae{\scriptscriptstyle\pm\mechGlobalMaeSd}$ \\
Prototypes (\method\revp{, $K{=}10$}) & {\boldmath$\ssCodebookTwelveMae{\scriptscriptstyle\pm\ssCodebookTwelveMaeSd}$} & {\boldmath$\ssCodebookTwentyFiveMae{\scriptscriptstyle\pm\ssCodebookTwentyFiveMaeSd}$} & {\boldmath$\ssCodebookFiftyMae{\scriptscriptstyle\pm\ssCodebookFiftyMaeSd}$} & \revp{{\boldmath$\stagePriorKten{\scriptscriptstyle\pm\stagePriorKtenSd}$}} & \revp{{\boldmath$\stageAdaptKten{\scriptscriptstyle\pm\stageAdaptKtenSd}$}} & {\boldmath$\oursMae{\scriptscriptstyle\pm\oursMaeSd}$} \\
\bottomrule
\end{tabular*}\end{minipage}}
\par\vspace{6pt}
\caption{\revp{\textbf{Prior types.} Peak MAE is split by measured peak and tail objects (``Others''). The word lists and the lookup use GT category information at test time. Peak RMSE: Tables~\ref{tab:mechanism} and~\ref{tab:organizations}.}}
\label{tab:prior-types}
\scalebox{\abltabscale}{\begin{minipage}{1.166667\linewidth}\fontsize{7}{8.166667}\selectfont\setlength{\tabcolsep}{2pt}\setlength{\medmuskip}{0mu}
\begin{tabular*}{\linewidth}{@{}l@{\extracolsep{\fill}}cccccc@{}}
\toprule
 & ATE & \multicolumn{4}{c}{Peak MAE (N) $\downarrow$} & Profile \\
\cmidrule(lr){3-6}
Variant & (cm) $\downarrow$ & \revp{Overall} & \revp{$<$20\,N ($n{=}\frngNLight$)} & \revp{$\geq$20\,N ($n{=}\frngNHeavy$)} & \revp{Tail ``Others'' ($n{=}\frngNOther$)} & RMSE (N) $\downarrow$ \\   %
\midrule
Direct MLP prior (matched size) & {\boldmath$\mechMlpAte{\scriptscriptstyle\pm\mechMlpAteSd}$} & $\mechMlpMae{\scriptscriptstyle\pm\mechMlpMaeSd}$ & \revp{{\boldmath$\frngLightMlp{\scriptscriptstyle\pm\frngLightMlpSd}$}} & \revp{$\frngHeavyMlp{\scriptscriptstyle\pm\frngHeavyMlpSd}$} & \revp{$\frngOtherMlp{\scriptscriptstyle\pm\frngOtherMlpSd}$} & $\underline{\mechMlpProf{\scriptscriptstyle\pm\mechMlpProfSd}}$ \\
Global prior + adaptation & $\underline{\mechGlobalAte{\scriptscriptstyle\pm\mechGlobalAteSd}}$ & $\mechGlobalMae{\scriptscriptstyle\pm\mechGlobalMaeSd}$ & \revp{$\underline{\frngLightGlobal{\scriptscriptstyle\pm\frngLightGlobalSd}}$} & \revp{$\frngHeavyGlobal{\scriptscriptstyle\pm\frngHeavyGlobalSd}$} & \revp{$\frngOtherGlobal{\scriptscriptstyle\pm\frngOtherGlobalSd}$} & $\mechGlobalProf{\scriptscriptstyle\pm\mechGlobalProfSd}$ \\
\revp{Prototype prior (no routing loss)} & $\mechNorouteAte{\scriptscriptstyle\pm\mechNorouteAteSd}$ & $\mechNorouteMae{\scriptscriptstyle\pm\mechNorouteMaeSd}$ & \revp{$\frngLightNoroute{\scriptscriptstyle\pm\frngLightNorouteSd}$} & \revp{$\frngHeavyNoroute{\scriptscriptstyle\pm\frngHeavyNorouteSd}$} & \revp{$\underline{\frngOtherNoroute{\scriptscriptstyle\pm\frngOtherNorouteSd}}$} & $\mechNorouteProf{\scriptscriptstyle\pm\mechNorouteProfSd}$ \\
Flat clusters (k-means, $K{=}10$) & $\orgKmeansAte{\scriptscriptstyle\pm\orgKmeansAteSd}$ & $\orgKmeansMae{\scriptscriptstyle\pm\orgKmeansMaeSd}$ & \revp{$\frngLightKmeans{\scriptscriptstyle\pm\frngLightKmeansSd}$} & \revp{$\frngHeavyKmeans{\scriptscriptstyle\pm\frngHeavyKmeansSd}$} & \revp{$\frngOtherKmeans{\scriptscriptstyle\pm\frngOtherKmeansSd}$} & $\orgKmeansProf{\scriptscriptstyle\pm\orgKmeansProfSd}$ \\
\revp{Word-list prior (LLM-assigned GT category)} & $\orgLlmAte{\scriptscriptstyle\pm\orgLlmAteSd}$ & $\underline{\orgLlmMae{\scriptscriptstyle\pm\orgLlmMaeSd}}$ & \revp{$\frngLightLlm{\scriptscriptstyle\pm\frngLightLlmSd}$} & \revp{{\boldmath$\frngHeavyLlm{\scriptscriptstyle\pm\frngHeavyLlmSd}$}} & \revp{$\frngOtherLlm{\scriptscriptstyle\pm\frngOtherLlmSd}$} & $\orgLlmProf{\scriptscriptstyle\pm\orgLlmProfSd}$ \\
\revp{Word-list prior (GT category)} & $\orgAnnotAte{\scriptscriptstyle\pm\orgAnnotAteSd}$ & $\orgAnnotMae{\scriptscriptstyle\pm\orgAnnotMaeSd}$ & \revp{$\frngLightAnnot{\scriptscriptstyle\pm\frngLightAnnotSd}$} & \revp{$\underline{\frngHeavyAnnot{\scriptscriptstyle\pm\frngHeavyAnnotSd}}$} & \revp{$\frngOtherAnnot{\scriptscriptstyle\pm\frngOtherAnnotSd}$} & $\orgAnnotProf{\scriptscriptstyle\pm\orgAnnotProfSd}$ \\
\revp{Category-mean lookup (GT object--action)} & \revp{--} & \revp{$\frngAllCatMean$} & \revp{$\frngLightCatMean$} & \revp{$\frngHeavyCatMean$} & \revp{$\frngOtherCatMean$} & \revp{--} \\
\midrule
\method{} (full model) & {\boldmath$\oursAte{\scriptscriptstyle\pm\oursAteSd}$} & {\boldmath$\oursMae{\scriptscriptstyle\pm\oursMaeSd}$} & \revp{$\frngLightOurs{\scriptscriptstyle\pm\frngLightOursSd}$} & \revp{$\frngHeavyOurs{\scriptscriptstyle\pm\frngHeavyOursSd}$} & \revp{{\boldmath$\frngOtherOurs{\scriptscriptstyle\pm\frngOtherOursSd}$}} & {\boldmath$\profileRmse{\scriptscriptstyle\pm\profileRmseSd}$} \\
\bottomrule
\end{tabular*}\end{minipage}}
\par\vspace{6pt}
\caption{\textbf{Ablations.} Each block varies one design choice\revp{; the full model (last row) is the reference, except for the motion cues and human video, with their own controls. Peak RMSE: Appendix~\ref{app:evidence}.}}
\label{tab:ablations}
\scalebox{\abltabscale}{\begin{minipage}{1.166667\linewidth}\fontsize{7}{8.166667}\selectfont\setlength{\tabcolsep}{1pt}\setlength{\medmuskip}{0mu}
\begin{tabular*}{\linewidth}{@{}p{95pt}@{\extracolsep{\fill}}ccc@{\hspace{1.25em}}p{95pt}ccc@{}}
\cmidrule[\heavyrulewidth](r){1-4}\cmidrule[\heavyrulewidth]{5-8}
Variant & ATE (cm) $\downarrow$ & \shortstack{Peak\\MAE (N) $\downarrow$} & \shortstack{Profile\\RMSE (N) $\downarrow$} & Variant & ATE (cm) $\downarrow$ & \shortstack{Peak\\MAE (N) $\downarrow$} & \shortstack{Profile\\RMSE (N) $\downarrow$} \\
\cmidrule[\lightrulewidth](r){1-4}\cmidrule[\lightrulewidth]{5-8}
\multicolumn{4}{@{}l}{\emph{Prediction head}} & \multicolumn{4}{l@{}}{\emph{Motion cues}} \\
\hspace{0.5em}One-shot regression & $\flowRegressAte{\scriptscriptstyle\pm\flowRegressAteSd}$ & $\flowRegressMae{\scriptscriptstyle\pm\flowRegressMaeSd}$ & $\flowRegressProf{\scriptscriptstyle\pm\flowRegressProfSd}$ & \hspace{0.5em}All motion cues (control) & $\underline{\flowAprimeAte{\scriptscriptstyle\pm\flowAprimeAteSd}}$ & $\underline{\flowAprimeMae{\scriptscriptstyle\pm\flowAprimeMaeSd}}$ & $\underline{\flowAprimeProf{\scriptscriptstyle\pm\flowAprimeProfSd}}$ \\
\hspace{0.5em}Split trajectory / force flows & $\flowSplitAte{\scriptscriptstyle\pm\flowSplitAteSd}$ & $\flowSplitMae{\scriptscriptstyle\pm\flowSplitMaeSd}$ & $\underline{\flowSplitProf{\scriptscriptstyle\pm\flowSplitProfSd}}$ & \hspace{0.5em}No dense motion field & $\flowNofieldAte{\scriptscriptstyle\pm\flowNofieldAteSd}$ & $\flowNofieldMae{\scriptscriptstyle\pm\flowNofieldMaeSd}$ & $\flowNofieldProf{\scriptscriptstyle\pm\flowNofieldProfSd}$ \\
\cmidrule(r){1-4}\cmidrule{5-8}
\multicolumn{4}{@{}l}{\emph{Guidance} (full model uses manual click + manual prompt)} & \multicolumn{4}{l@{}}{\emph{Human video}} \\
\hspace{0.5em}Self-click + manual prompt & $\revr{\selfClickRawAte{\scriptscriptstyle\pm\selfClickRawAteSd}}$ & {\boldmath$\revr{\selfClickRawMae{\scriptscriptstyle\pm\selfClickRawMaeSd}}$} & $\revr{\selfClickRawProf{\scriptscriptstyle\pm\selfClickRawProfSd}}$ & \hspace{0.5em}With paired gripper force ref. & $\oursHandAte{\scriptscriptstyle\pm\oursHandAteSd}$ & {\boldmath$\oursHandRef{\scriptscriptstyle\pm\oursHandRefSd}$} & {\boldmath$\oursHandRefProf{\scriptscriptstyle\pm\oursHandRefProfSd}$} \\
\hspace{0.5em}Manual click + self-prompt & $\underline{\selfPromptAte{\scriptscriptstyle\pm\selfPromptAteSd}}$ & $\selfPromptMae{\scriptscriptstyle\pm\selfPromptMaeSd}$ & $\revr{\selfPromptProf{\scriptscriptstyle\pm\selfPromptProfSd}}$ & \hspace{1em}(control) & & & \\
\hspace{0.5em}Self-click + self-prompt & $\revr{\selfPromptSelfClickAte{\scriptscriptstyle\pm\selfPromptSelfClickAteSd}}$ & $\revr{\selfPromptSelfClickMae{\scriptscriptstyle\pm\selfPromptSelfClickMaeSd}}$ & $\revr{\selfPromptSelfClickProf{\scriptscriptstyle\pm\selfPromptSelfClickProfSd}}$ & \hspace{0.5em}Without paired gripper force ref. & {\boldmath$\nopairHandAte{\scriptscriptstyle\pm\nopairHandAteSd}$} & $\nopairHandRef{\scriptscriptstyle\pm\nopairHandRefSd}$ & $\nopairHandRefProf{\scriptscriptstyle\pm\nopairHandRefProfSd}$ \\
\cmidrule[\lightrulewidth](r){1-4}\cmidrule[\lightrulewidth]{5-8}
\method{} (full model) & {\boldmath$\oursAte{\scriptscriptstyle\pm\oursAteSd}$} & $\underline{\oursMae{\scriptscriptstyle\pm\oursMaeSd}}$ & {\boldmath$\profileRmse{\scriptscriptstyle\pm\profileRmseSd}$} & \method{} (full model) & {\boldmath$\oursAte{\scriptscriptstyle\pm\oursAteSd}$} & {\boldmath$\oursMae{\scriptscriptstyle\pm\oursMaeSd}$} & {\boldmath$\profileRmse{\scriptscriptstyle\pm\profileRmseSd}$} \\
\cmidrule[\heavyrulewidth](r){1-4}\cmidrule[\heavyrulewidth]{5-8}
\end{tabular*}\end{minipage}}
\vspace{-16.5pt}
\end{table}

\textbf{\revp{What do learned prototypes and routing contribute?}}
With instance adaptation and joint flow fixed, learned prototypes with routing supervision outperform both a parameter-matched MLP prior\reve{, which maps the video and prompt features directly to a peak and profile,} and the shared global prior \reve{of Sec.~\ref{sec:prior}} in peak and profile error (Table~\ref{tab:prior-types})\revp{, and they model heavy and tail forces well}. Removing routing supervision raises peak MAE, indicating that prototype capacity alone does not account for the gain \revp{(Appendix~\ref{app:prior-contribution}). Accuracy is stable from 5 to 30 prototypes, and beyond 10 some stay unused (Appendix~\ref{app:ksweep}).}

\textbf{\revf{How do learned prototypes compare with other prior types?}}
\looseness=-1 \revp{The word lists group the training measurements by four object names and two actions. The category of a sequence is either its GT category or one that an LLM assigns from this list based on the prompt. Learned prototypes match both word lists in peak MAE (Table~\ref{tab:prior-types}) and peak RMSE (\oursRmse{} vs.\ \orgLlmRmse--\orgAnnotRmse\,N; Table~\ref{tab:organizations}), without category information at test time. Other priors are best on single subsets, the direct MLP below 20\,N and the LLM-assigned word list on peaks $\geq$20\,N, whereas \method{} is best overall, on tail objects and in profile RMSE, and holistically the most accurate prior.} Flat clusters, which assign each prompt to its nearest of ten $k$-means clusters, are less accurate than \revp{the prototypes and the word lists}.

\textbf{\revf{Does the joint flow improve the correction, and what do the motion cues contribute?}}
Flow matching improves prediction over one-shot regression\reve{, which predicts the same residual in a single step with the same network} (Table~\ref{tab:ablations}, \emph{Prediction head}).
\revf{The joint flow also improves on separate trajectory and force streams \reve{that share the conditioner but do not attend to each other}.} \revp{The flow complements the prior and the adaptation (Table~\ref{tab:sensory}, right): after the learned prior, it changes the peak little and refines the trajectory (Table~\ref{tab:trajectory}); if the prior is not good, as with one global prior ($K{=}1$), the flow does the heavy correction.} \revf{The dense motion field contributes to both outputs}: removing it significantly increases force prediction errors. Appendix~\ref{app:joint-prediction} reports the complete comparison.

\textbf{\revf{Does paired supervision improve force transfer to human video?}}
Paired supervision carries measured gripper force to human video (Table~\ref{tab:ablations}, \emph{Human video}; Appendix~\ref{app:transfer} for more): without it, the human-video peak disagreement with the paired gripper references rises from $\oursHandRef$ to $\nopairHandRef$\,N \revp{and the profile error rises too}, while the trajectory error stays similar.

\textbf{\revf{Can the click and prompt be automated?}}
\reve{Both generated inputs keep force prediction close to the manual setting (Table~\ref{tab:ablations}, \emph{Guidance}).} The automated self-click has errors within std of the full model, while the automated prompt slightly reduces force prediction accuracy.

%% file: sections/realworld.tex
\subsection{\texorpdfstring{Real-World Interactions, Robot Deployment, and \revr{Simulation}}{Real-World Interactions, Robot Deployment, and Simulation}}
\label{sec:realworld}

We test \method beyond Hoi! through separate video and robot studies (protocols in Appendix~\ref{app:realworld}). 

\textbf{Real-world videos.} We record seven phone videos of a person repeatedly opening and closing six kitchen objects---a refrigerator door, freezer, cabinet door, drawer, oven door, and dishwasher door---at varying speeds and extents. We divide them into 46 \revb{episodes of 32 frames}, each spanning contact to the end of motion. We evaluate \revr{the repeatability of the predictions,} action and motion direction, and agreement between predicted peaks and the scale readings \revr{(Appendix~\ref{app:rw:videos})}.
\definecolor{vfscale}{HTML}{222222}
\DeclareRobustCommand{\vfScaleDiamond}{\tikz[baseline=-0.5ex]\draw[vfscale, line width=0.55pt, fill=white] (0,0.6ex) -- (0.6ex,0) -- (0,-0.6ex) -- (-0.6ex,0) -- cycle;}
\definecolor{vflimit}{HTML}{D9D9D9}
\DeclareRobustCommand{\vfLimitBand}{\tikz[baseline=-0.5ex]\fill[vflimit] (0,-0.6ex) rectangle (2.2ex,0.6ex);}
\begin{figure}[t]
\centering
\includegraphics[width=0.836\textwidth]{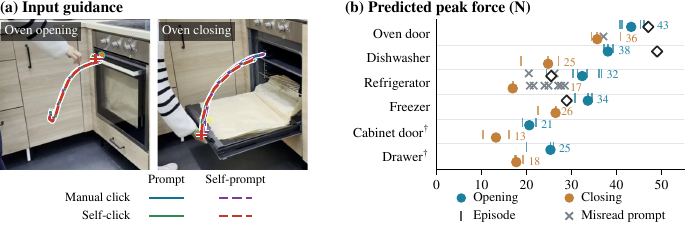}
\caption{\textbf{Real-world videos.} (a) Predicted trajectories under the four input combinations; plus signs mark the contact used. (b) Predicted peak forces with self-click and self-prompt; medians exclude misread \revc{prompts}. \vfScaleDiamond{}: opening force measured with a luggage scale; \revp{\dag: the scale was unable to capture a reading due to limited settling time.}}
\label{fig:wild}\label{fig:input-modes}
\vspace{-1.5mm}
\end{figure}
\makeatletter\newcommand{\vfFigureCaptionInTable}{\def\@captype{figure}}\makeatother
\begin{table}[!tbp]
\centering\setlength{\abovecaptionskip}{2.3pt}
\caption{\textbf{Spot trials.} Left: Spot opening the oven door and drawer. Right: successes out of five; failures: F force, T trajectory, P partial opening.}
\label{tab:spot}
\begin{minipage}[c]{0.33\textwidth}
\includegraphics[width=\linewidth]{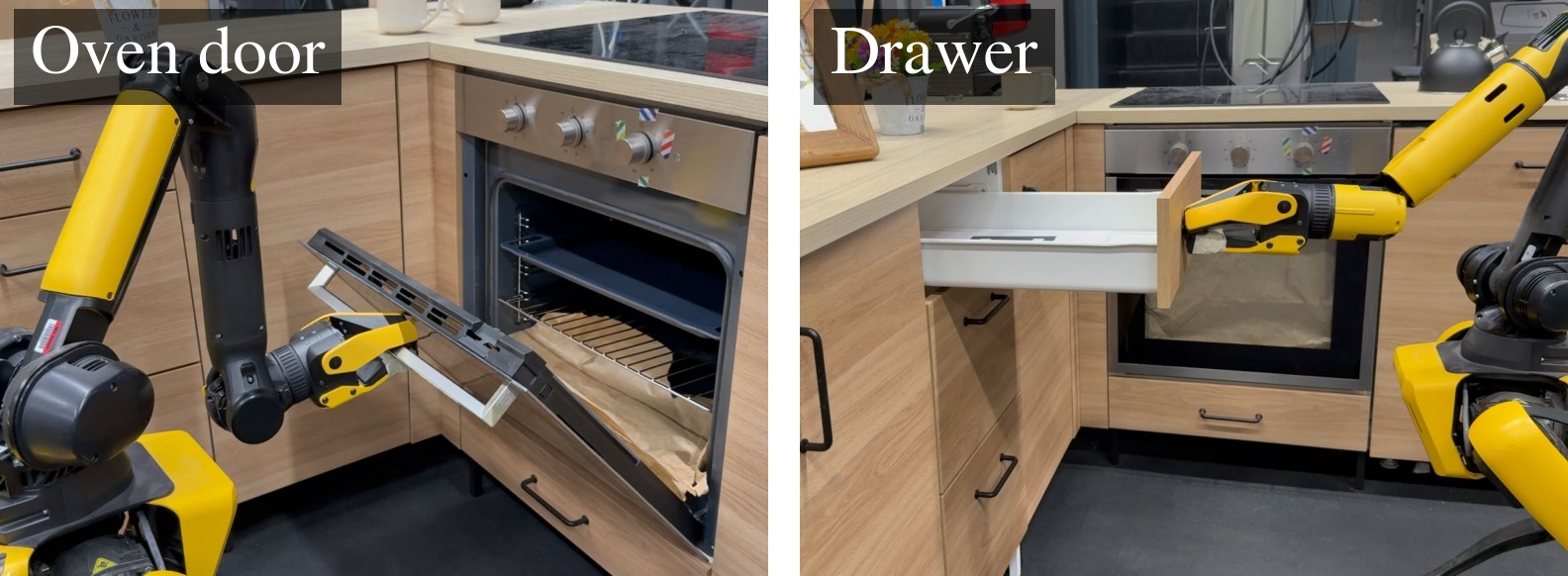}
\end{minipage}\hfill
\begin{minipage}[c]{0.65\textwidth}
\centering\fontsize{6}{7}\selectfont\setlength{\tabcolsep}{2.6pt}\renewcommand{\arraystretch}{0.85}
\newcommand{\spotfail}[1]{\,{\tiny(#1)}}
\begin{tabular*}{\linewidth}{@{\extracolsep{\fill}}llcccc@{}}
\toprule
 & & \multicolumn{4}{c}{\revp{Feed-forward}} \\
\cmidrule(lr){3-6}
Object & Viewpoint & \revp{Without force} & \multicolumn{2}{c}{\revp{Constant force}} & \revp{Force profile} \\
\midrule
Drawer & View 1 (empty) & \revp{5/5} & \multirow{2}{*}{\revp{8\,N}} & \revp{5/5} & 5/5 \\
(prismatic) & View 2 ($\sim$5\,kg) & \revp{3/5\spotfail{2T}} & & \revp{4/5\spotfail{1T}} & 4/5\spotfail{1T} \\
\addlinespace[2pt]
Oven door & View 1 & \revp{0/5\spotfail{4F, 1T}} & \multirow{2}{*}{\revp{20\,N}} & \revp{0/5\spotfail{4F, 1T}} & 3/5\spotfail{1P, 1T} \\
(revolute) & View 2 & \revp{0/5\spotfail{4F, 1T}} & & \revp{0/5\spotfail{4F, 1T}} & 4/5\spotfail{1T} \\
\bottomrule
\end{tabular*}
\end{minipage}
\par\vspace{4.7pt}
\centering
\includegraphics[width=0.9\linewidth]{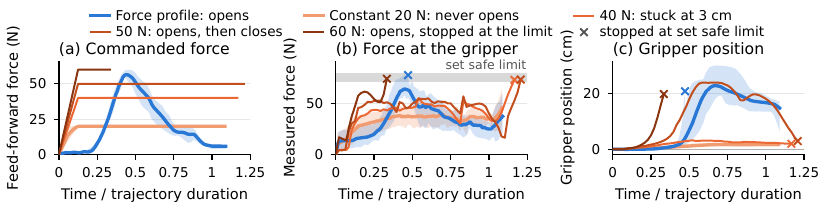}
{\vfFigureCaptionInTable\setlength{\abovecaptionskip}{4.2pt}
\caption{\revp{\textbf{Oven door: force profile and constant forces.} (a) Commanded and (b) measured force at the gripper, (c) gripper position along the opening. \vfLimitBand{}: set safe limit, 1.3 times the maximum planned force; $\times$: stopped at it.}}
\label{fig:spot-oven}}
\vspace{-17.2pt}
\end{table}

With self-click and self-prompt, \method{} produces distinct force estimates across actions and objects (Fig.~\ref{fig:wild}b). Every object receives a higher predicted peak for opening than for closing. The routed prior assigns 31\,N to 20 of the 21 correctly read openings, so instance adaptation and joint refinement produce most of the variation between objects. \reve{Prompt generation reads the action correctly in 36 of 46 episodes, with most misreads on refrigerator closings.} \revp{A drawer can be empty or hold a very heavy object. In simulation, a force profile like ours opens it either way (Table~\ref{tab:sim-impact}).}

Figure~\ref{fig:input-modes}a qualitatively compares the predicted trajectories for oven opening and closing. The four input combinations produce nearly identical trajectories and similar peak estimates. For closing, the predicted peak varies slightly more between 37 to 41\,N.

\input{tables/sim_impact}
\textbf{Robot deployment.} Spot records human demonstrations with its \revb{arm} camera, and \method{} \revp{predicts the contact trajectory and force profile of each} with self-click and self-\revc{prompt}. We execute the \revp{trajectories} on a drawer and an oven door from two viewpoints, \revp{without force, with a constant force, and} with the predicted force profile\revp{, under the same controller settings} (Table~\ref{tab:spot}). The predicted trajectories carry over across joint types and viewpoints\revp{: the drawer opens from the trajectory alone, even when loaded. The force profile lets Spot open the high-effort oven door, which stays shut without force and under the constant force. Without force feedback, small trajectory errors on revolute doors cause excessive force \citep{Wang2025ArticuBot}; to prevent damage from such a force build-up, we set a safe limit on the gripper force that stops a run. In a systematic constant-force sweep, analyzed in Appendix~\ref{app:rw:robot}, a constant 60\,N opened the oven door relatively quickly but hit our safe limit (Fig.~\ref{fig:spot-oven}). A constant force calibrated well for each object or category can also achieve similar success, as on the drawer. However, a constant force can overshoot at first contact \citep{Ji2026DreamingSound} or slam the panel (Table~\ref{tab:sim-impact}); a smoothly rising profile such as ours avoids both. With the force profile and trajectory predicted by \method{}, Spot can execute the full articulation safely (Fig.~\ref{fig:spot-oven}).}

\revp{\textbf{Simulation.} In MuJoCo, we open hand-built twins of a drawer (empty, with 3.4\,kg of small objects, or with a 35\,kg block) and of an oven door with a demonstrated force profile and constant forces (Table~\ref{tab:sim-impact}; Appendix~\ref{app:rw:sim}). The profile opens all three drawers with zero impact energy. A constant pull at the peak force keeps accelerating the drawer, resulting in high impact energy. A force profile leads to a low-impact, controlled articulation even when the full dynamics are unknown, while a constant force must be calibrated by hand for each object \citep{Jain2010doors,Jain2013forces}.}

%% file: tables/sim_impact.tex
\begin{table}[!tbp]\revtable
\centering\setlength{\abovecaptionskip}{3pt}
\caption{\revp{\textbf{Simulation.} Left: the simulated drawer, empty and with 3.4\,kg of free contents. Right: hand-built MuJoCo twins (Appendix~\ref{app:rw:sim}). End: travel completed; $E$: kinetic energy at the end stop.}}
\label{tab:sim-impact}
\begin{minipage}[c]{0.33\textwidth}
\includegraphics[width=\linewidth]{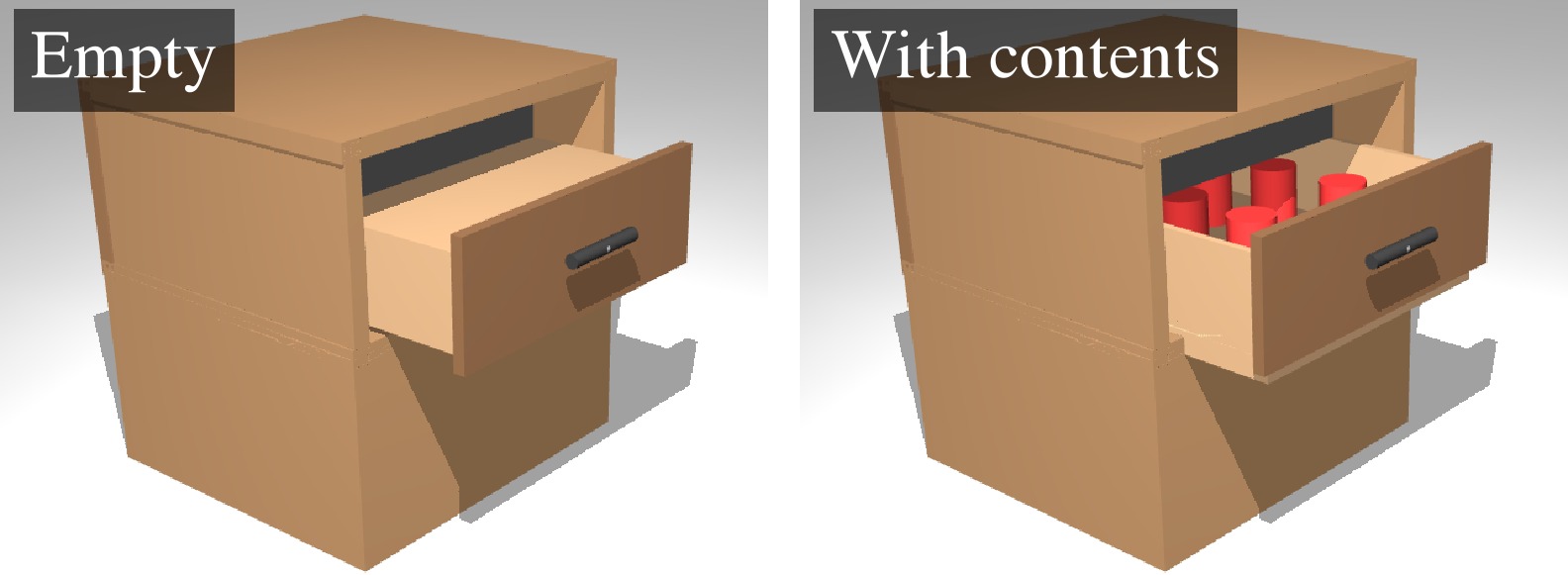}
\end{minipage}\hfill
\begin{minipage}[c]{0.65\textwidth}
\centering\fontsize{6}{7}\selectfont\setlength{\tabcolsep}{2.6pt}\renewcommand{\arraystretch}{0.85}
\newcommand{\simend}[1]{\edef\simreach{#1}\def\simyes{yes}\def\simno{no}\ifx\simreach\simyes\checkmark\else\ifx\simreach\simno$\times$\else #1\fi\fi}
\newcommand{\simconsider}[2]{\edef\simreach{\csname simC#1#2Reach\endcsname}\def\simno{no}\ifx\simreach\simno\else
  \ifdim\csname simC#1#2KE\endcsname pt<\dimen0 \dimen0=\csname simC#1#2KE\endcsname pt\relax\fi\fi}
\newcommand{\simconsidertwo}[2]{\edef\simreach{\csname simC#1#2Reach\endcsname}\def\simno{no}\ifx\simreach\simno\else
  \ifdim\csname simC#1#2KE\endcsname pt>\dimen0 \ifdim\csname simC#1#2KE\endcsname pt<\dimen2 \dimen2=\csname simC#1#2KE\endcsname pt\relax\fi\fi\fi}
\newcommand{\simsetmin}[1]{\dimen0=\maxdimen\simconsider{#1}{Prof}\simconsider{#1}{ConstPeak}\simconsider{#1}{CatMean}\simconsider{#1}{Shuf}%
  \dimen2=\maxdimen\simconsidertwo{#1}{Prof}\simconsidertwo{#1}{ConstPeak}\simconsidertwo{#1}{CatMean}\simconsidertwo{#1}{Shuf}%
  \expandafter\edef\csname simMin#1\endcsname{\the\dimen0}\expandafter\edef\csname simSec#1\endcsname{\the\dimen2}}
\simsetmin{Drawer}\simsetmin{Loaded}\simsetmin{Heavy}\simsetmin{Oven}
\newcommand{\simE}[4]{\edef\simreach{#1}\def\simno{no}\ifx\simreach\simno--\else\ifdim #2pt=#3\textbf{#2}\else\ifdim #2pt=#4\underline{#2}\else #2\fi\fi\fi}
\newcommand{\simrow}[2]{#1 & \simend{\csname simCDrawer#2Reach\endcsname} & \simE{\csname simCDrawer#2Reach\endcsname}{\csname simCDrawer#2KE\endcsname}{\simMinDrawer}{\simSecDrawer}
  & \simend{\csname simCLoaded#2Reach\endcsname} & \simE{\csname simCLoaded#2Reach\endcsname}{\csname simCLoaded#2KE\endcsname}{\simMinLoaded}{\simSecLoaded}
  & \simend{\csname simCHeavy#2Reach\endcsname} & \simE{\csname simCHeavy#2Reach\endcsname}{\csname simCHeavy#2KE\endcsname}{\simMinHeavy}{\simSecHeavy}
  & \simend{\csname simCOven#2Reach\endcsname} & \simE{\csname simCOven#2Reach\endcsname}{\csname simCOven#2KE\endcsname}{\simMinOven}{\simSecOven} \\}
\begin{tabular*}{\linewidth}{@{\extracolsep{\fill}}lcccccccc@{}}
\toprule
 & \multicolumn{2}{c}{Empty drawer} & \multicolumn{2}{c}{Drawer + 3.4\,kg} & \multicolumn{2}{c}{Drawer + 35\,kg} & \multicolumn{2}{c}{Oven door} \\
\cmidrule(lr){2-3}\cmidrule(lr){4-5}\cmidrule(lr){6-7}\cmidrule(lr){8-9}
Feed-forward & End & $E$ (J) & End & $E$ (J) & End & $E$ (J) & End & $E$ (J) \\
\midrule
\simrow{Force profile}{Prof}
\simrow{Constant (peak force)}{ConstPeak}
\simrow{Constant (profile mean)}{CatMean}
\simrow{Shuffled profile}{Shuf}
\bottomrule
\end{tabular*}
\end{minipage}
\vspace{-13.0pt}
\end{table}

%% file: sections/limitations.tex
\subsection{Limitations and Future Directions}
\label{sec:limitations}
\method{} estimates typical interaction effort. Direct human-force measurements would complement paired gripper supervision and support person-specific predictions, while broader force data would help distinguish visually similar objects with different resistance and refine estimates for high-effort interactions. Our evaluation focuses on visible interactions, with metric scale estimated from monocular depth. \emph{Future work} will expand object-centric data to interactions with distinct force signatures, such as stapler use and USB insertion at varying speeds. We plan to extend tracking to moving cameras and multiple viewpoints, and explore joint trajectory and force prediction from a single image.

%% file: sections/conclusion.tex
\label{sec:conclusion}
\reve{\method{} brings sensory estimation to articulated-object understanding by jointly predicting a contact trajectory and force profile from third-person video. Learned physical prototypes share limited instrumented measurements across interactions of comparable effort; geometric and vision--language features adapt this prior to the observation, and a conditional flow refines it together with a tracked trajectory. \method{} improves trajectory recovery over every baseline and achieves the lowest force errors among third-person methods on held-out Hoi!\ interactions. A separately trained torque model further shows that the formulation can extend to other sensory quantities. The model produces consistent motion and force predictions on real-world videos, and \revp{enables robots to articulate objects safely}. These results show how limited instrumented data can enrich ordinary interaction videos with estimates of how objects move and how much effort their manipulation may require.}

%% file: sections/acknowledgments.tex
\section*{Acknowledgments}
This research was partially supported by Google, by the Robotics Institute Germany, and by the Lamarr Institute for Machine Learning and Artificial Intelligence. The authors gratefully acknowledge the access to the Marvin cluster of the University of Bonn. The authors specifically thank Tim Engelbracht for support with robot experiments.

%% file: sections/llm_statement.tex
\section*{AI Use Statement}
LLM-based assistants were used while preparing this paper. They helped edit the text for clarity and consistency, draft and revise parts of sections under the authors' direction, write scripts that produce figures and tables from the logged results, and format references. Coding agents were used in implementing and debugging training, evaluation, and baseline code and in running experiments; the authors reviewed this code and checked the reported results against the experiment logs. Separately, Qwen3-VL and GPT-5.5 are components of \method{} and baselines in our experiments, as described in Sections~\ref{sec:method} and~\ref{sec:exp}. The authors take full responsibility for the content of this paper.

%% file: sections/appendix_method.tex
\label{app:impl}

\subsection{Inputs and representation}
\label{app:impl:repr}

\textbf{Object-centric video clip.}
The input is a square window of the recording around the manipulated object, sampled at $T=32$ frames over the interaction; the frozen models run on this window, and $\mathbf K$ denotes its intrinsics. For recorded evaluations with a \revc{manual click} the window is sized from the annotated articulation ($1.3\times$ its largest image-space reach from the contact, at least $0.4\times$ the frame's longer side); in the automatic setting it spans $0.55\times$ the longer side around the proposed contact. In training the window is shifted and rescaled so that the contact does not sit at a fixed position (Appendix~\ref{app:aug}).

\textbf{Frozen models and memory tokens.}
Three frozen models turn the clip into the tokens of Table~\ref{tab:tokens}. \textbf{\emph{D4RT}}, through its open reimplementation~\citep{OpenD4RT2026}, tracks a $32\times32$ lattice of points and a block of points seeded at the contact through the 32 frames. The lattice tracks become a motion field of 18 channels per cell and frame (displacements, contact offsets, visibility, confidence, normals, and depth consistency); a temporal convolution reduces it to 8 steps, and a $4\times4$ patchification turns it into 64 field tokens, with 8 timeline tokens summarizing the motion-weighted sequence. The contact-seeded tracks yield two \revf{contact-track} tokens and the tracked initialization $\mathbf B$, and D4RT's encoder features, pooled to $8\times4\times4$, enter as 128 tracker tokens. \textbf{\emph{DA3}}~\citep{Lin2025DA3} provides the depth map, the intrinsics $\mathbf K$, and the contact depth $z_c$; a scale token carries the metric-scale fit and per-frame scale statistics. \textbf{\emph{Qwen3-VL-8B}}~\citep{Bai2025Qwen3VL} provides 16 pooled vision tokens on each of 8 keyframes, a four-token embedding of the \revc{prompt}, and the contact proposal that serves as the contact cue. It also produces the 258-dimensional \revc{prompt} embedding $\mathbf e(q)$ read by the router: the square roots of the probabilities it assigns to a fixed vocabulary of 256 answer tokens and to the two actions when asked, given the \revc{prompt}, which object part is manipulated and whether it is opened or closed.

A fusion transformer reads all of these tokens and returns the context $\mathbf M$; the pooled features $\mathbf g$ are the mean over its output tokens. It has three pre-norm layers of width 512 with eight attention heads, a feed-forward width of 1024, and dropout 0.1.

\begin{table}[htbp]
\centering\scriptsize\setlength{\tabcolsep}{5pt}
\caption{\textbf{Memory tokens of the fusion transformer.} One row per token type of the object-centric clip: its count, the frozen model or signal that produces it, and what it encodes. The D4RT tracks yield the field, timeline, contact-track, and tracker tokens, DA3 the scale token, and Qwen3-VL the vision, prompt, and contact-cue tokens.}
\label{tab:tokens}
\begin{tabular}{lrp{0.27\linewidth}p{0.36\linewidth}}
\toprule
Token & Count & Source & Encodes \\\midrule
Field & 64 & D4RT lattice tracks & tracked motion field, 18 channels per cell and frame \\
Timeline & 8 & D4RT lattice tracks & motion-weighted temporal sequence \\
Contact tracks & 2 & contact-seeded D4RT tracks & mean track curve and its dispersion around the contact \\
Scale & 1 & DA3 depth and tracks & metric-scale fit and per-frame scale statistics \\
Tracker features & 128 & D4RT encoder, pooled to $8\times4\times4$ & learned geometry and motion representation \\
Vision & 128 & Qwen3-VL, 16 pooled tokens on 8 keyframes & appearance and semantics \\
\revc{Prompt} & 4 & Qwen3-VL text embedding & the object--action \revc{prompt} \\
Contact cue & 1 & Qwen3-VL proposal & proposed hand or gripper point; dropped with probability 0.5 in training \\
Embodiment and mode & 1 & metadata & human or gripper; click-free and prior-free flags \\
\bottomrule
\end{tabular}
\end{table}

\textbf{Coordinates and tracked initialization.}
A camera-space waypoint $\mathbf p_i=(X_i,Y_i,Z_i)^\top$ projecting to pixel $(u_i,v_i)$ of the window is encoded relative to the contact as
\begin{equation}
 \mathbf x_i=\bigl(u_i/(w-1)-c_u,\;v_i/(h-1)-c_v,\;\log(Z_i/z_c)\bigr)^\top ,
 \label{eq:state}
\end{equation}
with $(w,h)$ the window size in pixels, $(c_u,c_v)$ the normalized contact coordinates, and $z_c$ the DA3 depth at the contact; $\mathbf K$ and $z_c$ invert the encoding to metric coordinates. The tracked initialization takes, for each waypoint, the componentwise median over the contact-seeded tracks $\mathcal S_i$ visible at both the contact frame $t_c$ and the waypoint frame $t_i$ (at least eight):
\begin{equation}
 \mathbf b_i=\operatorname*{med}_{s\in\mathcal S_i}
 \left[\mathbf u_s(t_i)-\mathbf u_s(t_c),\;
 \log\frac{z_s(t_i)}{z_s(t_c)}\right].
 \label{eq:base}
\end{equation}
For human recordings the lattice cells within 0.06 of the contact replace the seeded tracks, which the hand occludes more often; waypoints \revf{without surviving tracks} copy the nearest \revf{tracked} waypoint or stay at the contact.

\subsection{Prototype prior}
\label{app:impl:prior}
The prior of Sec.~\ref{sec:prior} is trained with the settings in Table~\ref{tab:prior-settings}. All prototypes are initialized from the \emph{global prior}, the median measured peak and the mean normalized profile of the training \revc{sequences}, which also stands in for the routed prior on prior-dropout rows and serves as the no-prototype control in Sec.~\ref{sec:exp}; the levels are then spread to the quantiles of the measured training peaks, so the prototypes start as a force ladder. The number of prototypes, $K=10$, is the \rev{largest tested} value in the sweep of Table~\ref{tab:ksweep} at which every prototype stays in use.

\makeatletter\newenvironment{tablenofloatpage}{\@float{table}[!htb]}{\end@float}\makeatother
\begin{tablenofloatpage}
\centering\scriptsize\setlength{\tabcolsep}{5pt}
\caption{\textbf{Settings of the prototype prior.} Initialization and size of the prototypes, the router and its two losses, the instance-adaptation head with its bound and loss, and prior dropout, which substitutes the global prior on a share of the training sequences.}
\label{tab:prior-settings}
\begin{tabular}{p{0.24\linewidth}p{0.68\linewidth}}
\toprule
Setting & Value \\\midrule
Prototypes & $K=10$ (sweep in Table~\ref{tab:ksweep}); levels start at evenly spaced quantiles of the measured training peaks (5.0 to 49.3\,N), shapes at the mean normalized training profile, keys (258-dimensional) at random \\
Router & two-layer network of width 512 on $\mathbf g$ giving one score per prototype, plus the key cosine with a learned scale $\beta$ (initialized at 10); soft assignment in training and at inference \\
Routing loss & $\tau_r=0.3$; all recordings with a force target; weight 0.5 \\
Consistency loss & cross-entropy of the key scores toward the prototype selected by the video score (target detached); recordings with a \revc{prompt}; weight 0.5 \\
Adaptation head & two-layer network of width 512 on $\mathbf g$ giving the log-peak offset and the $L$ shape offsets; zero-initialized output; bound $c=\ln 2$, i.e.\ the peak changes by at most a factor of two \\
Adaptation loss & Huber on the log peak (threshold 0.5) plus squared error on the shape; weight 1 \\
Prior dropout & 30\,\% of training \revc{sequences} use the global prior instead of the routed prior, with the adaptation still applied \\
\bottomrule
\end{tabular}
\end{tablenofloatpage}

\subsection{Joint conditional flow}
\label{app:impl:flow}
The flow of Sec.~\ref{sec:flow} operates on a residual state of 49 tokens: 16 waypoint tokens (three coordinates each, relative to the tracked proposal), one peak token (the log ratio to the adapted prior peak), and 32 shape tokens (one per profile sample, relative to the adapted shape). Table~\ref{tab:flow-settings} lists its settings; the target uses the adapted prior with gradients detached, so the flow learns residuals around a fixed reference and the prior is shaped only by its own loss.

\begin{tablenofloatpage}
\centering\scriptsize\setlength{\tabcolsep}{5pt}
\caption{\textbf{Settings of the joint conditional flow.} The residual state and its standardization, the DiT-style velocity network, what the waypoint and force tokens read, the loss weights $\Omega$ and the one-step peak term, and the sampling and peak-bound settings used at inference.}
\label{tab:flow-settings}
\begin{tabular}{p{0.28\linewidth}p{0.64\linewidth}}
\toprule
Setting & Value \\\midrule
State & 81 coordinates in 49 tokens, standardized per coordinate over the training \revc{sequences}; unsupervised tokens are masked \\
Network & 4 transformer blocks, width 512, 8 heads: self-attention over the tokens, cross-attention to $\mathbf M$, modulation by $\mathbf g$ and the flow time; zero-initialized outputs \\
Token inputs & waypoint tokens read the tracked proposal and 13 channels of tracked motion around the contact; force tokens read speed, acceleration, and displacement statistics of the tracked proposal \\
Loss weights $\Omega$ & span mask on the waypoints; force weight 1 (measured), 0.5 (paired reference), 0 (none), scaled by $\sqrt{\text{peak}/\text{global peak}}$ within $[0.25,4]$; force term counted twice \\
One-step peak term & Huber loss (5\,N) on the peak decoded from the one-step estimate, weighted by $t$; weight 0.5 \\
Sampling & 20 Euler steps from zero; one deterministic sample \\
Peak bound $b$ & 0.5 in log space \\
\bottomrule
\end{tabular}
\end{tablenofloatpage}

\textbf{Decoding.}
After the standardization is reversed, the waypoint residual is added to the tracked proposal and mapped to metric coordinates with Eq.~\ref{eq:state}, and the force outputs are
\begin{equation}
 \widehat F=\bar F'\exp\!\left(\operatorname{clip}(\widehat\rho,-b,b)\right),\qquad
 \widehat{\mathbf f}=\widehat F\,[\bar{\boldsymbol\sigma}'+\widehat{\boldsymbol\delta}]_+,
\end{equation}
with $\widehat\rho$ the predicted log-peak residual and $\widehat{\boldsymbol\delta}$ the predicted shape residuals. The shape is not renormalized after decoding, so $\widehat F$ can differ from the maximum of $\widehat{\mathbf f}$\profscaled{. The output profile is therefore $\widehat{\mathbf f}$ scaled so that its maximum equals $\widehat F$; the figures, the robot and the profile errors use it.}

\subsection{Training}
\label{app:impl:train}
Every training \revc{sequence} supervises the trajectory. Force targets come from the instrumented gripper recordings and, for human recordings that have a gripper recording of the same episode, from that paired recording at half weight; human recordings without a partner supervise the trajectory only. The model has 42.9\,M trainable parameters on top of the frozen backbones. All training and evaluation were performed on four NVIDIA \revb{A100} GPUs. Tables~\ref{tab:training} and~\ref{tab:losses} give the schedule and the loss terms.

\begin{table}[htbp]
\centering\scriptsize\setlength{\tabcolsep}{5pt}
\caption{\textbf{Training schedule.} Optimizer and learning-rate schedule, the number of updates and batch size, weight averaging, the point at which the frozen teacher takes over the force targets, and the dropout and augmentation applied to the training sequences.}
\label{tab:training}
\begin{tabular}{p{0.24\linewidth}p{0.68\linewidth}}
\toprule
Setting & Value \\\midrule
Optimizer & AdamW, learning rate $5\times10^{-4}$, weight decay 0.01 (prototypes exempt); linear warm-up over the first 5\,\% of training, then cosine decay to zero \\
Length and batch & 12,000 updates with batches of 128 (32 per GPU); three seeds; the final checkpoint is used \\
Weight averaging & exponential moving average of the weights with decay 0.999; evaluation uses the averaged weights \\
Force teacher & at update 2000 the averaged weights are frozen as a teacher whose predictions \reve{become the force target of every training sequence, all with the same weight,} for the rest of training \\
Dropout & encoder 0.1; \revc{prompt} 0.2; prior 0.3 (the global prior is substituted); contact cue 0.5; click-free path on 30\,\% of recordings \\
Augmentation & shifted and rescaled windows with mirror images, contact and depth jitter, horizontal flips (Appendix~\ref{app:aug}) \\
\bottomrule
\end{tabular}
\end{table}

\begin{table}[htbp]
\centering\scriptsize\setlength{\tabcolsep}{5pt}
\caption{\textbf{Loss terms of Eq.~\ref{eq:loss}.} Each term with its definition and weight: the flow-matching loss with the one-step peak term, the prior losses (adapted prior, routing, consistency), and the auxiliary scale, peak-count, and contact-localization terms. Targets that a sequence lacks are masked.}
\label{tab:losses}
\begin{tabular}{p{0.24\linewidth}p{0.56\linewidth}r}
\toprule
Term & Definition & Weight \\\midrule
Flow & velocity error of Eq.~\ref{eq:flow} with the weights of Table~\ref{tab:flow-settings} & 1 \\
One-step peak & Huber loss in newtons (threshold 5\,N) on the peak decoded from the one-step estimate & 0.5 \\
Adapted prior & Huber loss on the log peak (threshold 0.5) plus squared error on the shape & 1 \\
Routing & cross-entropy of Eq.~\ref{eq:route-loss} & 0.5 \\
Consistency & cross-entropy of the key scores toward the video-selected prototype & 0.5 \\
Metric scale & smooth-$\ell_1$ error on the log scale of the tracked depth & 1 \\
Peak count & cross-entropy over the number of force peaks & 0.2 \\
Contact localization & cross-entropy over the 64 field tokens toward a Gaussian heatmap at the contact & 0.5 \\
\bottomrule
\end{tabular}
\end{table}

\textbf{Why a force teacher.}
Measured during development: without a teacher, the validation force error reached its minimum early in training and had risen by about 15\,\% by the time the trajectory error converged, so no single checkpoint served both outputs, and freezing the force weights alone did not help because the shared features keep moving. The teacher step is therefore fixed early, where the force error had leveled off in those runs. With the teacher, the force error keeps decreasing slowly after the switch while the trajectory error still falls by roughly half, and the final checkpoint is used for both outputs.

\subsection{Inference with generated inputs}
\label{app:impl:inference}
At inference the model needs the clip and, optionally, a contact click and a \revc{prompt}; either can be generated by the frozen Qwen3-VL-8B. The evaluation protocol for generated inputs is given in Appendix~\ref{app:inputs}.

\textbf{Self-click.}
The contact is the point Qwen3-VL proposes for where the hand or gripper touches the object. The object-centric window is re-centered on it (side $0.55\times$ the longer image side) and the tracks and trajectory proposal are recomputed there. A localization head can refine the proposal within 1.5 grid cells, after which the motion field is re-anchored in a second pass; the head is trained on the click-free training rows, where the contact is perturbed by 1.5 grid cells and the click-seeded \revf{track} tokens are dropped.

\textbf{Self-prompt.}
The \revc{prompt} is written by the same model. For the recorded evaluations it is composed from the model's answers about the manipulated part and its opening degree at the start and end of the interaction; for the \revb{real-world} and robot videos it is the model's answer to one free-form question over eight frames. A generated \revc{prompt} is encoded exactly like a supplied one.

\textbf{Questions.}
The questions below are used verbatim; nothing is fine-tuned.

\emph{Contact proposal (self-click).} Asked on the first frame of the interaction; $\langle$action$\rangle$ holds the supplied \revc{prompt} or, in the self-prompt setting, the model's own \revc{prompt}. The answer is a point in normalized image coordinates.
\begin{quote}\small
The image shows a person's hand or a robot gripper about to $\langle$action$\rangle$. Point to the exact pixel where the hand or gripper touches (or is about to touch) the object it manipulates---the contact point on the handle, knob, edge or surface. Answer with JSON only: \{``point\_2d'': [x, y]\}
\end{quote}

\emph{Self-written \revc{prompt}, recorded evaluations (self-prompt).} Two questions on windows around the proposed contact at the start and the end of the interaction. The action is the sign of the change in opening degree (at least 15 points; smaller changes fall back to a two-image comparison), and the \revc{prompt} is composed as ``$\langle$action$\rangle$ $\langle$attributes$\rangle$ $\langle$part$\rangle$'', for example ``open white toilet lid''.
\begin{quote}\small
(1) These two images are the first and the last frame of one short manipulation of an articulated object by a person's hand or by a robot gripper. Name the object part that is manipulated (the part that moves: a door, drawer, lid, \ldots). Answer with JSON only [fields \emph{object} and \emph{attributes}].

(2) Look at the $\langle$part$\rangle$ in this image. How open is it right now? Judge only the current geometry of the part: 0 means fully closed (flush with its frame or fully pushed in), 100 means fully open (swung or pulled out as far as it goes). Answer with JSON only: \{``open\_percent'': $\langle$integer from 0 to 100$\rangle$\}
\end{quote}

\emph{Self-written \revc{prompt}, \revb{real-world} videos and robot deployment.} One free-form question over eight frames spanning the episode; the answer is used as the \revc{prompt} without editing.
\begin{quote}\small
The frames are in temporal order and show a person acting on an object in a home. In one short imperative sentence, say what the person does and to which part of which object, as an \revc{prompt} someone could follow. Answer with the \revc{prompt} only.
\end{quote}

\subsection{\texorpdfstring{\revr{Inference cost}}{Inference cost}}
\label{app:impl:cost}
\revr{Table~\ref{tab:inference-cost} gives the time and GPU memory of every component per 32-frame episode on one RTX PRO 6000 Blackwell Max-Q (96\,GB), with all models resident. With a supplied prompt and contact, an episode takes 4.19\,s; fully automatic (self-prompt and self-click), 5.04\,s. Per-frame depth takes 67\,\% of the 4.19\,s, and the trained model (42.9\,M parameters) 2\,\%. The models occupy 27.6\,GiB (30.7\,GiB at peak), more than a 24-GB card holds. Baselines on the same per-frame depth take 2.95--3.64\,s, GPT-5.5 7.9--9.4\,s through the API, EgoInfinity 10.5\,s, and Articulate-Anything 557\,s.}

\input{tables/rebuttal_inference_cost}

%% file: tables/rebuttal_inference_cost.tex
\begin{table}[htbp]\revtable
\centering\scriptsize\setlength{\tabcolsep}{3.5pt}
\caption{\revr{\textbf{Inference cost per 32-frame episode.} Median time over the 99 Eval episodes on one RTX PRO 6000 Blackwell Max-Q (96\,GB), all models loaded. GPU memory in GiB: weights $+$ peak activation (totals: loaded / peak). $^a$Only without a supplied prompt or contact. $^b$Including data assembly; the network alone takes 53\,ms. $^c$From the evaluation logs: the full pipeline per clip, 93\,\% of it waiting for its 32 Gemini calls. $^d$Three API calls of eight frames each, network time included. $^e$Own models only.}}
\label{tab:inference-cost}
\begin{tabular}{@{}llrrrr@{}}
\toprule
Component / method & Model & Parameters & GPU memory (GiB) & Time (s) & API cost \\
\midrule
\multicolumn{6}{@{}l}{\emph{\method{}, deployed pipeline (one episode per call)}} \\
\hspace{0.5em}Self-prompt$^a$ & Qwen3-VL-8B & 8.77\,B & 16.3 + 0.8 & 0.48 &  \\
\hspace{0.5em}Contact proposal (self-click)$^a$ & Qwen3-VL-8B & (shared) & (shared) + 0.1 & 0.38 &  \\
\hspace{0.5em}Prompt encoding (word list, routing, embedding) & Qwen3-VL-8B & (shared) & (shared) + 0.1 & 0.30 &  \\
\hspace{0.5em}Depth, 32 single-frame passes & DA3-Nested-Giant-Large & 1.69\,B & 6.3 + 2.0 & 2.79 &  \\
\hspace{0.5em}Motion field (1{,}088 tracks) & OpenD4RT & 1.16\,B & 4.3 + 3.1 & 0.85 &  \\
\hspace{0.5em}Vision tokens, 8 frames & Qwen3-VL-8B vision & (shared) & (shared) + <0.1 & 0.10 &  \\
\hspace{0.5em}Feature assembly and cache I/O & -- & -- & -- & 0.07 &  \\
\hspace{0.5em}Trajectory and force, 20 Euler steps$^b$ & Velocity network (trained) & 42.9\,M & 0.2 + <0.1 & 0.07 &  \\
\textbf{Total, annotated prompt and contact} & & 11.66\,B & 27.6 / 30.7 & \textbf{4.19} & \\
\textbf{Total, fully automatic}$^a$ & & 11.66\,B & 27.6 / 30.7 & \textbf{5.04} & \\
\multicolumn{6}{@{}l}{\emph{Baselines (same GPU and episodes unless marked)}} \\
\hspace{0.5em}D4RT initialization & DA3 + OpenD4RT & 2.85\,B & 13.7$^e$ & 3.64 &  \\
\hspace{0.5em}CoTracker3 + DA3 & DA3 + CoTracker3 & 1.72\,B & 10.9$^e$ & 3.01 &  \\
\hspace{0.5em}ForceSight (released, adapted) & DA3 + RGB-D ViT-L + T5 & 2.15\,B & 10.7$^e$ & 2.95 &  \\
\hspace{0.5em}EgoInfinity (hands) & MoGe-2 + WiLoR + flow & 1.13\,B & 3.1 + 1.6 & 10.5 &  \\
\hspace{0.5em}Articulate-Anything$^c$ & Gemini 2.5 Flash + retrieval & -- & -- & 557 & \$0.24 \\
\hspace{0.5em}GPT-5.5 + text$^d$ & API, 3 calls & -- & 0 (API) & 7.93 & \$0.044 \\
\hspace{0.5em}GPT-5.5$^d$ & API, 3 calls & -- & 0 (API) & 9.40 & \$0.045 \\
\hspace{0.5em}Category table / constant & lookup & 0 & 0 & 0 &  \\
\bottomrule
\end{tabular}
\end{table}

%% file: sections/appendix_data.tex
\subsection{Source recordings}
\label{app:data:source}
Hoi!~\citep{Engelbracht_2026_CVPR} captures people and a hand-held, instrumented gripper operating the same articulated furniture and appliances, filmed by two static third-person cameras. We use the exocentric (third-person) recordings of the hand and Hoi!-gripper embodiments that were available on the Hoi! servers at the time of this work: 25 sessions in 20 indoor scenes, 18 of which have gripper recordings and therefore force measurements. A scene, session, and interaction \revc{span} define an \emph{episode}; our unit, a \emph{sequence}, is one camera view of an episode in one embodiment (Hoi! counts the two views as one sequence). Nearly every episode was performed by both embodiments, so it has up to four third-person recordings, 1{,}434 recordings of 395 episodes in total. Gripper recordings carry a tracked tool trajectory and force--torque measurements, human recordings a tracked hand trajectory and no force (below). For 191 recordings the release lacks an input the pipeline needs (gripper calibration or poses, articulation annotation, or hand landmarks); they are set aside before any quality assessment, a matter of availability rather than quality.

\textbf{Ground-truth trajectories.}
All Hoi! streams share one world frame (QR-tag time synchronization, visual localization against a laser scan of the scene). Gripper: the SLAM pose of the Aria device mounted on the gripper, mapped by the calibrated device-to-tool transform to the tool point; the force--torque sensor supplies the wrench in its own frame. Human: the thumb fingertip of the manipulating hand from the Aria glasses' hand tracking (landmark 0), mapped from the device to the world frame by the glasses' SLAM pose. Both are expressed in the iPhone camera frame through its aligned pose at the observation frame and resampled to 16 waypoints over the interaction \revc{span}.

\reve{\textbf{Force and torque targets.}
The gripper carries a six-axis Bota force--torque sensor; the release provides its compensated and filtered external wrench, and the gripper's contact trigger marks when the tool touches the object. The force target is the magnitude of the measured force, and the torque target is the magnitude of the torque in the sensor frame, as logged and without a shift to the contact point. Both are restricted to the contact span, from the first to the last contact sample. The peak is the maximum magnitude over the raw contact samples, and the 32-sample profile is the Savitzky--Golay-smoothed magnitude on a uniform grid over the span, rescaled so that its maximum equals the peak.}

\subsection{Quality filter}
\label{app:data:filter}
\begin{figure}[htbp]
\centering
\includegraphics[width=0.85\linewidth]{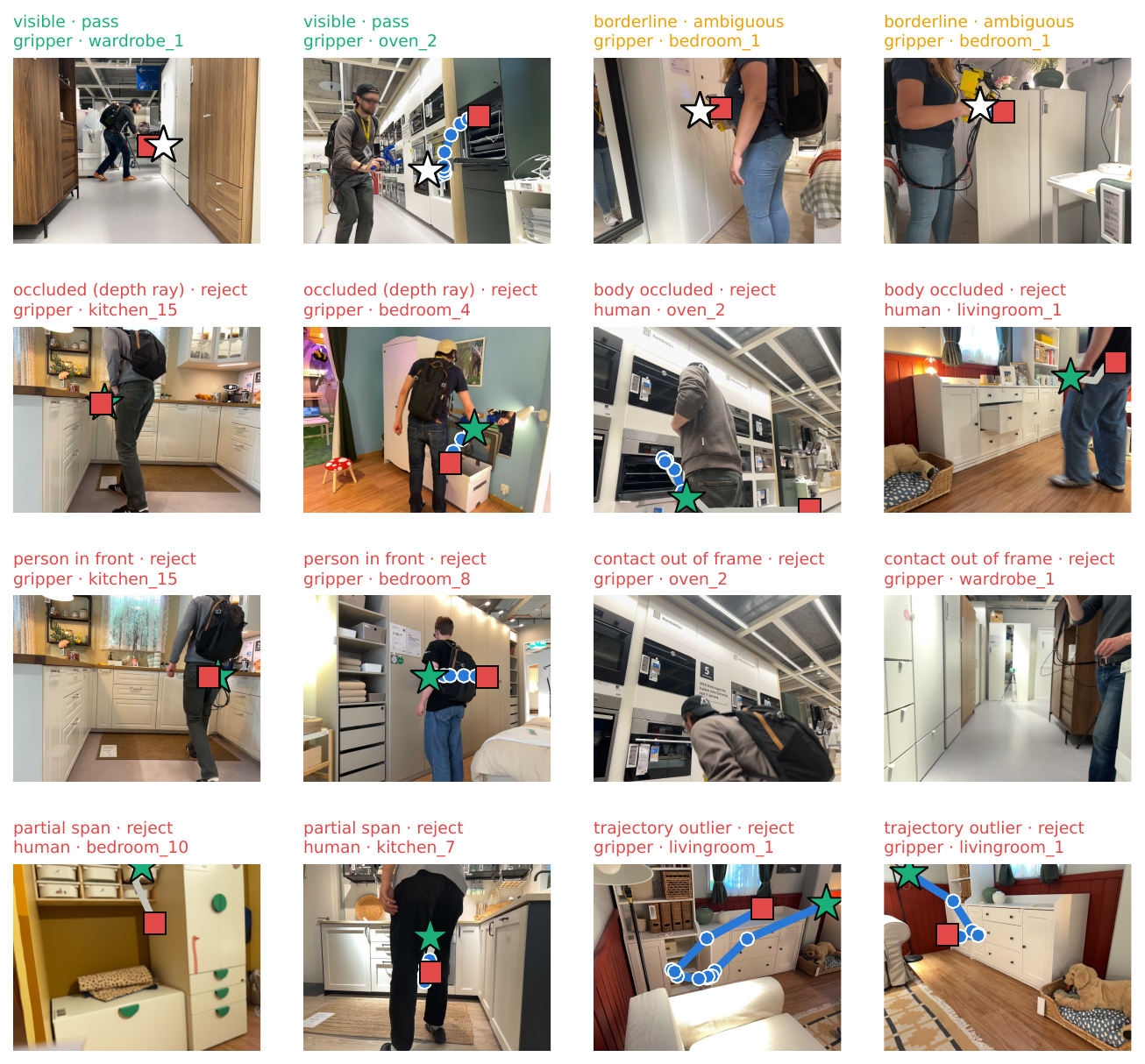}
\caption{\textbf{Filter examples.} Two random recordings per occlusion tag at their first frame, with the annotated path \sympath{}, its supervised span \symspan{}, the first \symfirst{} and last \symlast{} annotated point, and the annotated contact \symcontact{}.}
\label{fig:filter-examples}
\end{figure}

The filter keeps the recordings in which the interaction can actually be seen; in the third-person views the operator often walks the gripper rig between the camera and the object, and hands and contacts leave the frame. Every recording is tested at its annotated contact with a \emph{depth-ray gap}, the distance between the three-dimensional contact and the depth image at its projection: up to 0.20\,m the contact is visible, up to 0.45\,m borderline, beyond that occluded. Recordings occluded by the person, with the contact outside the image, with less than half of the articulation path in view, or with an implausibly long path (corrupted tracking) are rejected; borderline gaps are ambiguous; the rest pass. Human recordings are trimmed to the span in which the object moves, from grip onset to the first confident stop, release, or a per-scene cap; the median trimmed span covers 7 of the 16 waypoint frames, whereas gripper \revc{spans} are used as delimited by the release. Two thirds of the buildable recordings pass, 6\,\% are ambiguous and unused, and 28\,\% are rejected; Fig.~\ref{fig:filter-examples} shows two recordings per tag.

\subsection{Sequences, splits, and pairs}
\label{app:data:records}
A recording becomes a sequence when it passes the filter, can be calibrated, and shows the contact and the object in its first frame (Table~\ref{tab:funnel}); intrinsics come from the gripper recording of the same episode and camera, so human recordings without a buildable gripper counterpart remain unused. The sequences come from 280 episodes in 17 scenes. Splits were fixed per episode before filtering (Table~\ref{tab:splits}): the living room is held out as a whole, and a stratified set of episodes from ten training scenes forms the held-out interactions. Every human recording is paired with the gripper recording of its episode, from the same camera where one exists; the partner's peak and normalized profile serve as its reference targets and describe gripper force rather than the person's effort. \nPaired{} of the \nTrainHuman{} human training sequences have a partner (\nPairedSame{} same camera, \nPairedCross{} other camera), and \nHandRef{} evaluation human sequences have a measured evaluation partner (Appendix~\ref{app:transfer}). The release names no objects; for per-object reporting we name each episode by the part its gripper prompt mentions, with a visual check of the few evaluation episodes without a descriptive prompt. Measured peaks differ between objects and between opening and closing (Table~\ref{tab:peaks}); the spread within an object and action is what the instance adaptation and the flow have to explain.

\input{tables/tab_funnel}

\input{tables/tab_splits}

\input{tables/tab_peaks}

\subsection{Training windows and augmentation}
\label{app:aug}
Training windows are shifted and rescaled so that the model cannot read the contact off a fixed image position. Each training sequence contributes three windows with side $1.2$--$2.2\times$ the canonical window and the contact placed anywhere within the central 70\,\% of the window, each with its mirror image; trajectories, force profiles, intrinsics, and \revc{prompts} are transformed consistently, and the canonical window is sampled with probability 0.3. Further perturbations are one-cell contact jitter, log-depth jitter with scale 0.05, and horizontal flips with probability 0.5; the encoder, \revc{prompt}, prior, and contact-cue dropout rates are listed with the training schedule in Appendix~\ref{app:impl:train}. The same construction defines the re-framed evaluation set \EvalP (Appendix~\ref{app:evalprotocol}).

%% file: tables/tab_funnel.tex
\begin{table}[htbp]
\centering\scriptsize\setlength{\tabcolsep}{5pt}
\caption{\textbf{Data funnel.} Every third-person recording of the release, per embodiment and camera, counted at the first gate it fails: not buildable from the release, rejected or ambiguous in the quality filter, no gripper counterpart, contact or object not visible in the first frame, or corrupted motion capture; the last row gives the sequences that remain.}
\label{tab:funnel}
\begin{tabular}{lrrrrr}
\toprule
 & \multicolumn{2}{c}{Gripper} & \multicolumn{2}{c}{Human} & \\
\cmidrule(lr){2-3}\cmidrule(lr){4-5}
 & Camera 1 & Camera 2 & Camera 1 & Camera 2 & All \\
\midrule
Third-person recordings & 350 & 372 & 362 & 350 & 1434 \\
\midrule
Not buildable from the release & 86 & 86 & 7 & 12 & 191 \\
Rejected by the quality filter & 64 & 107 & 96 & 87 & 354 \\
Ambiguous filter verdict & 4 & 6 & 28 & 33 & 71 \\
No gripper counterpart & 0 & 0 & 80 & 46 & 126 \\
Contact or object not visible in the first frame & 9 & 14 & 13 & 17 & 53 \\
Corrupted motion capture & 2 & 1 & 3 & 3 & 9 \\
\midrule
Sequences & 185 & 158 & 135 & 152 & 630 \\
\bottomrule
\end{tabular}
\end{table}

%% file: tables/tab_splits.tex
\begin{table}[htbp]
\centering\scriptsize\setlength{\tabcolsep}{5pt}
\caption{\textbf{Sequences by split and embodiment.} Training, unseen-scene, and held-out-interaction sequences of the gripper and the person. Measured force exists for every gripper sequence and for none of the human sequences, whose force targets are the paired gripper references.}
\label{tab:splits}
\begin{tabular}{lrrr}
\toprule
Split & Gripper & Human & Total \\
\midrule
Training & 295 & 236 & 531 \\
Unseen scene & 23 & 24 & 47 \\
Held-out interactions & 25 & 27 & 52 \\
\midrule
All & 343 & 287 & 630 \\
\bottomrule
\end{tabular}
\end{table}

%% file: tables/tab_peaks.tex
\begin{table}[htbp]
\centering\scriptsize\setlength{\tabcolsep}{5pt}
\caption{\textbf{Measured peak force by object and action.} Count, median, and 10th and 90th percentiles of the measured peak over the 343 gripper sequences, for opening and closing each object; the spread within an object and action is what instance adaptation and the flow have to explain.}
\label{tab:peaks}
\begin{tabular}{llrrrr}
\toprule
Object & Action & $n$ & Median (N) & P10 (N) & P90 (N) \\
\midrule
Drawer & open & 58 & 14.6 & 5.9 & 26.9 \\
Drawer & close & 67 & 12.5 & 7.4 & 18.9 \\
Cupboard door & open & 64 & 13.2 & 7.5 & 21.9 \\
Cupboard door & close & 69 & 7.5 & 4.8 & 13.8 \\
Oven door & open & 33 & 48.6 & 31.0 & 68.8 \\
Oven door & close & 27 & 29.3 & 18.9 & 46.9 \\
Refrigerator door & open & 8 & 41.8 & 39.5 & 54.5 \\
Refrigerator door & close & 7 & 24.6 & 14.6 & 25.0 \\
Other & open & 7 & 11.2 & 8.3 & 27.9 \\
Other & close & 3 & 15.3 & 9.1 & 58.0 \\
\bottomrule
\end{tabular}
\end{table}

%% file: sections/appendix_experiments.tex
\renewcommand{\floatpagefraction}{0.8}
\subsection{Evaluation protocol}
\label{app:evalprotocol}
\textbf{Sets.}
Eval, the evaluation set of the main text, is the evaluation split of Appendix~\ref{app:data:records}: the \nVal{} sequences of the unseen living room (\nUnseen) and of the held-out interactions in ten training scenes (\nSeen), \nValForce{} of them instrumented. \EvalP applies the training-window construction of Appendix~\ref{app:aug} to every Eval sequence: three windows with side $1.2$--$2.2\times$ the canonical window that contain the whole articulation with the contact off-center, and their mirror images, six views per sequence (\nHard{} views, \nHardForce{} instrumented). Views of one sequence are correlated observations and count as views, not as additional interactions.

\textbf{Scoring.}
Every method receives the evaluation window and the same contact and is scored on the same 16 waypoints over the retained articulation span; on \EvalP every window is a row, mirrored windows against the mirrored reference. ATE aligns the starting point by translation only. Force metrics use the \nValForce{} measured gripper sequences (\nHardForce{} views on \EvalP); human references are never scored as measurements. The gripper-view ForceSight rows see the gripper-mounted camera and are scored once per episode (39 episodes of the same interactions). A dash marks a setting that was not evaluated. Nothing was tuned on Eval; the frozen export reproduces the model evaluator's D4RT-initialization ATE to within 0.01\,cm, and the baseline metric code was checked against it. Generated inputs (self-click, self-prompt) follow the protocol of Appendix~\ref{app:inputs}.

\subsection{Baseline implementation}
\label{app:baselines}
\begin{figure}[htbp]
\centering
\includegraphics[width=0.85\linewidth]{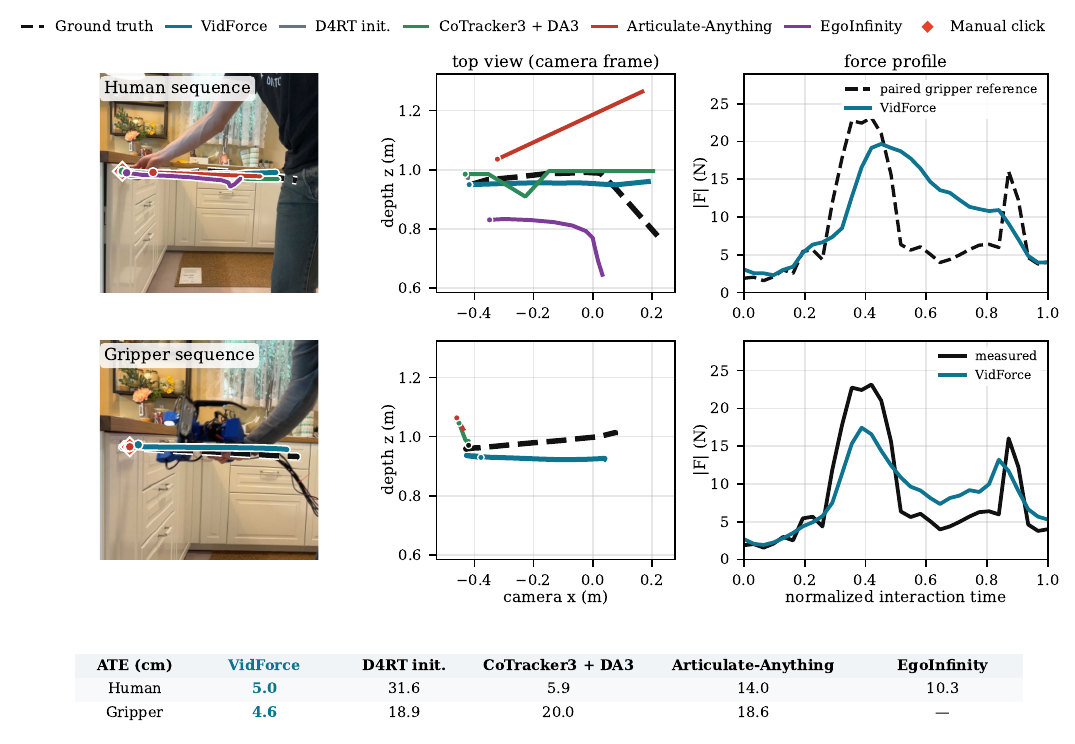}
\caption{\textbf{Trajectory baselines on one episode.} A kitchen drawer opened by the person and by the gripper, seen from the same camera; all methods use the manual click \symclick{}, and dots mark each start. \method{} is the seed-0 checkpoint.}
\label{fig:traj-baselines}
\end{figure}

Table~\ref{tab:baseline-overview} summarizes what each baseline does and which of the evaluated outputs it provides. Table~\ref{tab:extended-trajectory} lists ATE and ADE of every trajectory baseline and of our model on Eval and \EvalP, Fig.~\ref{fig:traj-baselines} shows them on one episode in both embodiments, and Table~\ref{tab:sensory-refs} the GPT-5.5 and ForceSight variants; all are scored as in Appendix~\ref{app:evalprotocol}.

\begin{table}[htbp]
\centering\scriptsize\setlength{\tabcolsep}{5pt}
\caption{\textbf{Baselines at a glance.} For each method: approach, input and camera view (T: third-person, G: gripper-mounted), whether it was trained on Hoi!, and which outputs it provides (trajectory, peak force, force profile). Only \method{} provides all three, and its contact point and prompt are optional.}
\label{tab:baseline-overview}
\begin{tabular}{lllcccc}
\toprule
Method & Approach & Input (view) & Trained & Traj. & Peak & Profile \\
\midrule
D4RT initialization & contact-seeded tracks & clip, contact (T) & -- & \checkmark & -- & -- \\
CoTracker3 + DA3 & tracked contact pixel & clip, contact (T) & -- & \checkmark & -- & -- \\
Articulate-Anything & \revb{Gemini-built} asset & 5 frames, contact (T) & -- & \checkmark & -- & -- \\
EgoInfinity & hand reconstruction & clip, contact (T, human) & -- & \checkmark & -- & -- \\
\midrule
GPT-5.5 & \revb{prompted} peak estimate & 8 frames, text (T) & -- & -- & \checkmark & -- \\
ForceSight, zero-shot & force goal & RGB-D frame, text (G, T) & -- & -- & \checkmark & -- \\
ForceSight, fine-tuned & goal fit to our force & RGB-D frame, text (G, T) & fine-tuned & -- & \checkmark & -- \\
ForceSight, adapted & per-frame force & RGB-D frame, text (G, T) & trained & -- & \checkmark & \checkmark \\
\midrule
\method{} & prior + joint flow & clip, contact, text (T) & trained & \checkmark & \checkmark & \checkmark \\
\bottomrule
\end{tabular}
\end{table}

\textbf{D4RT initialization.}
The proposal $\mathbf B$ of Eq.~\ref{eq:base} decoded to metric coordinates, i.e.\ the input our flow refines: contact-seeded D4RT tracks through the 32 frames, per-waypoint median image displacement and log-depth ratio over the tracks visible at both the contact and the waypoint frame, mapped to camera coordinates with the DA3 contact depth and $\mathbf K$. Waypoints without surviving tracks copy the nearest \revf{tracked} waypoint.

\textbf{CoTracker3 with monocular depth.}
The contact pixel is tracked through the 32 frames with the offline CoTracker3 model, backward from the contact frame, inside the evaluation window; each tracked pixel is lifted with the DA3 depth at that pixel and the frame intrinsics, occluded or invalid depth holds the last valid value, and the sequence is linearly resampled to the 16 waypoint times. The main variant smooths the lifted depth with a five-frame temporal median; the per-frame-depth variant is listed as well.

\textbf{EgoInfinity.}
EgoInfinity reconstructs MANO hands from a static-camera video and has no contact interface, so it runs on the \nValHand{} human sequences only. Each sequence is encoded as a \revc{video} of its 32 exported frames and processed with MoGe-2~\citep{Wang2025MoGe2} depth and focal length, WiLoR~\citep{Potamias2025WiLoR} hands with MANO~\citep{Romero2017MANO}, and the pipeline's motion stabilization; object reconstruction and tracking are disabled. The hand whose 2D joint is nearest the contact at the contact frame is followed and that joint is the contact proxy; sequences without a hand track fall back to the stationary prediction. The main row reads the trajectory in EgoInfinity's own reconstructed camera; a second row re-projects its joints with our camera intrinsics and rescales the trajectory so that its start depth equals the DA3 depth at the contact.

\textbf{Articulate-Anything.}
Gemini~2.5 Flash~\citep{Gemini25} with five frames per call, PartNet-Mobility\revb{~\citep{Xiang2020SAPIEN}} retrieval, and one actor--critic iteration (the original method used a different \revb{vision--language model}). The inferred asset is placed at the contact with yaw and scale fitted to the CoTracker3 track, and its trajectory follows the predicted joint range at constant speed. Uncompilable or unrenderable outputs (\aaFailed{} of \nVal{} Eval sequences, \aaFailedHard{} of \nHard{} re-framed windows) fall back to the stationary prediction and stay in the denominator, so the comparison includes reconstruction and registration errors.

\input{tables/extended_trajectory}

\textbf{Global prior with adaptation.}
This is our own model with the prototypes removed. Instead of $K$ routed prototypes, every sequence starts from one shared prior: the median measured training peak and the mean normalized training profile, the same statistics that initialize the prototypes, kept fixed during training. Everything else is unchanged and learned as in the main model: the instance adaptation corrects the shared prior from the video features, and the joint flow refines trajectory and force together with the same conditioner, losses, and schedule. The arm therefore isolates what the prototypes and their routing add over a shared starting point that the model still adapts to each video; read before adaptation, its error is that of predicting the training median for every sequence (Table~\ref{tab:prior-stages}). It is trained for force and for torque like the main model.

\textbf{GPT-5.5.}
Chat completions \revb{with the model snapshot \texttt{gpt-5.5-2026-04-23},} a strict JSON output schema\revb{,} and low reasoning effort; eight frames evenly spaced over the articulation span, either the full frame or the object-centric window, with or without the recorded \revc{prompt} and the object and action words, or the words and the room type alone (text only); the main text quotes the full-frame rows, Table~\ref{tab:sensory-refs} all of them. Three calls per input are averaged (sampling is the model's default, which cannot be fixed); identical full-frame inputs share calls across re-framed windows. The torque variant asks for the peak torque magnitude about the sensor frame in N\,m. These are adaptations for peak estimation, not models trained on the Hoi! targets.

\textbf{GPT-5.5 questions.}
The system message and the user templates are used verbatim; $\langle\cdot\rangle$ marks the fields filled per sequence.
\begin{quote}\small
\emph{System.} You are a careful physical-interaction analyst. You estimate the peak contact force (in newtons) needed at the hand or gripper contact point to perform an articulated-object interaction such as opening or closing a cupboard door, a drawer, an oven door or a fridge door. Think about the object type, its size and mass, hinge or slider friction, seals and dampers, and the motion visible in the frames. Answer with JSON only.

\emph{Frames + text.} You see 8 frames, evenly spaced over the interaction, from one fixed third-person camera (full frame). The interaction is: ``$\langle$\revc{prompt}$\rangle$''. The manipulated object is a $\langle$object$\rangle$ being $\langle$opened/closed$\rangle$. The contact point of the $\langle$hand/robot gripper$\rangle$ is marked by nothing; infer it from the frames. Estimate the PEAK force in newtons applied at the contact point over the whole interaction (the maximum of the force magnitude). Typical values for household articulated objects range from a few newtons to a few tens of newtons. Reply with JSON of the form \{``object'': ``...'', ``interaction'': ``...'', ``peak\_force\_newtons'': $\langle$number$\rangle$, ``confidence'': ``low|medium|high'', ``reasoning'': ``one or two sentences''\}.

\emph{Frames only.} You see 8 frames, evenly spaced over one manipulation, from one fixed third-person camera (full frame). First identify what articulated object is manipulated and what the interaction is (open or close, which part). Then estimate the PEAK force in newtons applied at the contact point over the whole interaction (the maximum of the force magnitude). Typical values \ldots [as above].

\emph{Text only.} A person or a robot gripper performs this interaction with an articulated household object: ``$\langle$\revc{prompt}$\rangle$''. The object is a $\langle$object$\rangle$ being $\langle$opened/closed$\rangle$, in a $\langle$room type$\rangle$. There is no image. Estimate the typical PEAK force in newtons applied at the contact point over the whole interaction (the maximum of the force magnitude). Reply with JSON \ldots [as above].
\end{quote}

\begin{figure}[htbp]
\centering
\includegraphics[width=0.85\linewidth]{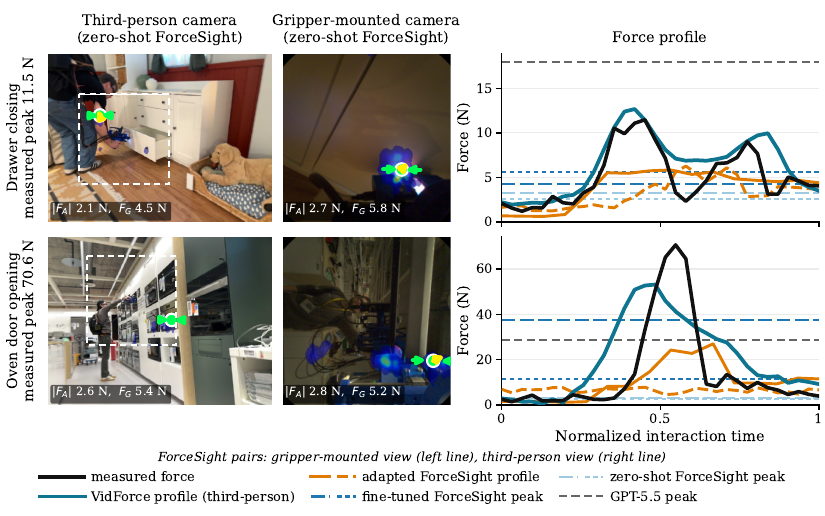}
\caption{\textbf{ForceSight on two episodes.} Zero-shot ForceSight at the frame of the measured peak: fingertip heat map, decoded fingertip \symfstip{}, applied-force goal \symfsforce{} and grip-force goals \symfsgrip{}, with their magnitudes $|F_A|$ and $F_G$; in the third-person view it sees only the object window \symwindow{}.}
\label{fig:forcesight}
\end{figure}

\textbf{ForceSight.}
ForceSight~\citep{Collins2023ForceSight} predicts, from one RGB-D frame and an \revc{prompt}, the force goal of a robot's next action; the released model is an RGB-D ViT-L with a frozen T5 text encoder, trained on a robot's wrist camera. Every variant sees frames squashed to $224^2$ with DA3 depth: in the third-person view the 32 frames of the object window our model sees, and in the gripper-mounted view, the view of the Hoi! zero-shot protocol~\citep{Engelbracht_2026_CVPR}, the whole image at 16 contact-anchored times per episode (39 Eval episodes); the peak is the maximum over frames. \emph{Zero-shot} rows run the released checkpoint with the authors' code and score the magnitude of its applied-force goal $|F_A|$; Fig.~\ref{fig:forcesight} draws its fingertip heat map and force goals. \emph{Fine-tuned} rows train all weights of the released checkpoint so that $|F_A|$ matches the measured force of each frame; the model keeps its force-goal output and is compared on the peak. Fine-tuning places no loss on the fingertip head, which then finds a fingertip in a quarter of the frames instead of most, so the heat maps in Fig.~\ref{fig:forcesight} are those of the zero-shot model. \emph{Adapted} ForceSight re-implements the architecture with the same ViT-L backbone and frozen T5 text path, replaces the goal outputs, most of which have no labels for human video, with one force output per frame, and trains it from ImageNet-21k initialization; its per-frame predictions form the profile. Fine-tuned and adapted models share data and schedule: in the third-person view, the 496 training sequences with a force target on the object window (295 measured gripper sequences and 201 paired human references at weight 0.5, as for our model); in the gripper-mounted view, the 199 measured training episodes of that view; 8{,}000 updates, batch 32, L2 loss, AdamW at $5\times10^{-5}$. In both views, adapted ForceSight matches the fine-tuned release within one seed standard deviation in peak error (Table~\ref{tab:sensory-refs}), so the adaptation does not weaken the baseline; trained on the object window with paired references it is the matched baseline of Table~\ref{tab:force}, and trained on the measured rows alone it performs similarly. None of the variants predicts a trajectory.

\begin{table}[htbp]
\centering\scriptsize\setlength{\tabcolsep}{2.8pt}
\caption{\textbf{Force baselines.} Errors on the instrumented evaluation sequences with \method{} for comparison. Per view, ForceSight is listed zero-shot, fine-tuned (all weights), and adapted; $^\ast$ marks the matched baseline of Table~\ref{tab:force}, and zero-shot and fine-tuned rows are scored on the peak only. Gripper-mounted rows cover the 39 episodes once each with 16-frame profiles; ``paired refs.'' adds the gripper measurements copied to the paired human sequences.}
\label{tab:sensory-refs}
\begin{tabular}{lllccc}
\toprule
& & & \multicolumn{3}{c}{Force error (N) $\downarrow$} \\
\cmidrule(lr){4-6}
Model & View & Training data & Peak MAE & Peak RMSE & Profile RMSE \\
\midrule
Training median (reference) & no video & measured & \maeConst & \rmseConst & -- \\
Object--action mean (reference) & no video & measured & \maeTable & \rmseTable & -- \\
\midrule
GPT-5.5 & 8 full frames & none & \maeGptImg & \rmseGptImg & -- \\
GPT-5.5 & 8 full frames + text & none & \maeGpt & \rmseGpt & -- \\
GPT-5.5 & 8 object-crop frames & none & \maeGptCrop & \rmseGptCrop & -- \\
GPT-5.5 & 8 object-crop frames + text & none & \maeGptCropText & \rmseGptCropText & -- \\
GPT-5.5 & text only & none & \maeGptText & \rmseGptText & -- \\
\midrule
ForceSight zero-shot & gripper-mounted & none & \fsZsGripViewMae & \fsZsGripViewRmse & -- \\
ForceSight fine-tuned & gripper-mounted & measured & $\fsFtMae\pm\fsFtMaeSd$ & $\fsFtRmse\pm\fsFtRmseSd$ & -- \\
ForceSight adapted & gripper-mounted & measured & $\fsLGripViewMae\pm\fsLGripViewMaeSd$ & $\fsLGripViewRmse\pm\fsLGripViewRmseSd$ & $\fsLGripViewProf\pm\fsLGripViewProfSd$ \\
\midrule
ForceSight zero-shot & third-person & none & \fsZsCropMae & \fsZsCropRmse & -- \\
ForceSight fine-tuned & third-person & measured + paired refs.\ & $\fsFtCropMae\pm\fsFtCropMaeSd$ & $\fsFtCropRmse\pm\fsFtCropRmseSd$ & -- \\
ForceSight adapted$^\ast$ & third-person & measured + paired refs.\ & $\fsLCropMae\pm\fsLCropMaeSd$ & $\fsLCropRmse\pm\fsLCropRmseSd$ & $\fsLCropProf\pm\fsLCropProfSd$ \\
ForceSight adapted & third-person & measured & $\fsLGripCropMae\pm\fsLGripCropMaeSd$ & $\fsLGripCropRmse\pm\fsLGripCropRmseSd$ & $\fsLGripCropProf\pm\fsLGripCropProfSd$ \\
\midrule
\method{} & third-person & measured + paired refs.\ & $\oursMae\pm\oursMaeSd$ & $\oursRmse\pm\oursRmseSd$ & $\profileRmse\pm\profileRmseSd$ \\
\bottomrule
\end{tabular}
\end{table}

\subsection{Dataset partitions and re-framing}
\label{app:partitions}
Performance is similar across the two Eval partitions. On the unseen living room and the held-out interactions in seen scenes, \method{} obtains ATEs of $\unseenAte\pm\unseenAteSd$ and $\seenAte\pm\seenAteSd$\,cm and peak MAEs of $\unseenMae\pm\unseenMaeSd$ and $\seenMae\pm\seenMaeSd$\,N, respectively (Table~\ref{tab:scene}). Table~\ref{tab:categories} provides the corresponding object-level breakdown.
\input{tables/tab_categories}

\input{tables/scene_results}

\EvalP tests sensitivity to framing rather than new interactions: it applies six shifted, rescaled, or mirrored views to each Eval sequence. \reve{Under re-framing, \method{} stays close to its canonical accuracy, with $\oursHardAte\pm\oursHardAteSd$ against $\oursAte\pm\oursAteSd$\,cm ATE and $\oursHardMae\pm\oursHardMaeSd$ against $\oursMae\pm\oursMaeSd$\,N peak MAE, and its trajectory advantage remains clear:} on \EvalP, D4RT and CoTracker3 score \ateDrtHard{} and \ateCtHard{}\,cm ATE, respectively (Table~\ref{tab:extended-trajectory}).

\begin{figure}[htbp]
\centering
\includegraphics[width=0.8\linewidth]{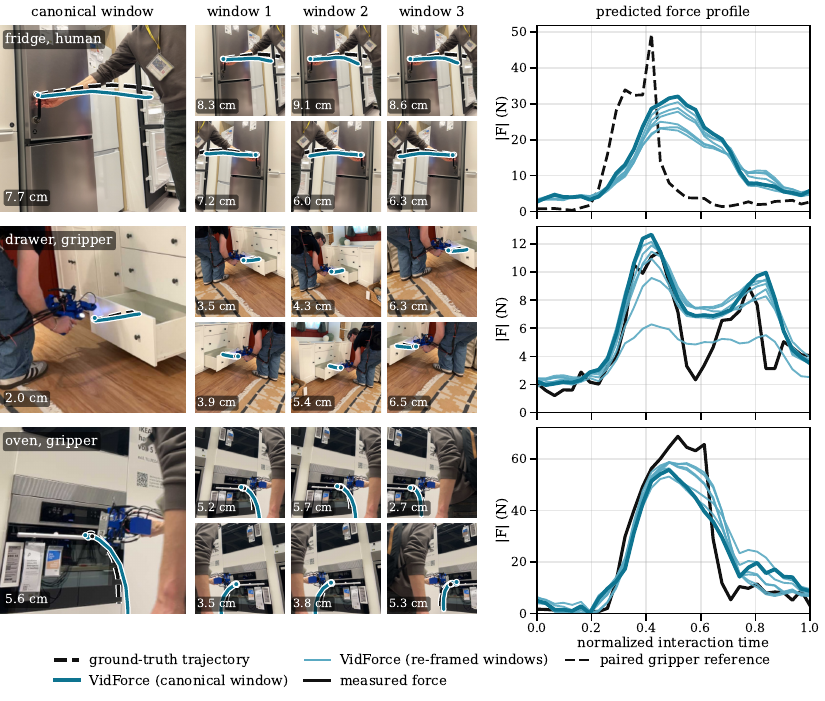}
\caption{\textbf{Predictions under re-framing.} Three interactions in the canonical view (left), three crops and their mirrors (center), and the corresponding force predictions (right). Trajectory overlays report ATE; force references are measured or, for the human sequence, paired.}
\label{fig:reframing}
\end{figure}

Fig.~\ref{fig:reframing} illustrates this behavior across three interactions. Applied to the real-world videos of Sec.~\ref{sec:realworld}, the same re-framings leave most peak-force estimates close to their canonical values.

\subsection{\texorpdfstring{\revr{Additional per-scene results}}{Additional per-scene results}}
\label{app:additional-scenes}
\revr{After development was complete, we obtained recordings of further Hoi! scenes that are not part of the public release. We report two of them, \texttt{livingroom\_2} and \texttt{kitchen\_6}, processed with the paper's pipeline and evaluated with the frozen checkpoints under the Eval protocol. The scenes \texttt{office\_4} and \texttt{kitchen\_2} were processed as well but are not reported here. Table~\ref{tab:additional-scenes} lists them per scene next to the unseen scene of Eval, \texttt{livingroom\_1} (Table~\ref{tab:scene}); the Eval results of the main text are unchanged.}
\revp{While the new scenes overall confirm robustness, we inspect the results for \texttt{livingroom\_2} further and find that the higher ATE comes from cabinet doors with glass inlets, which affect tracking and motion fields.}

\begin{table}[!ht]\revtable
\centering\scriptsize\setlength{\tabcolsep}{3pt}
\caption{\revr{\textbf{Results by unseen scene.} \method{} on the unseen scene of Eval, the two new unseen scenes, and all three combined.}}
\label{tab:additional-scenes}
\input{tables/additional_scene_results}
\end{table}

\subsection{Contribution of the learned prior}
\label{app:prior-contribution}
\textbf{Controlled comparison.}
Table~\ref{tab:mechanism} compares the learned prototypes with a parameter-matched MLP prior, a fixed global prior (Appendix~\ref{app:baselines}), and the same ten prototypes without $\mathcal L_{\mathrm{route}}$. \reve{The MLP prior replaces the prototype path with one MLP that maps the router's pooled video features and the \revc{prompt} embedding to a log-peak offset from the global prior, bounded to a factor of eight, and a 32-sample shape residual; it is zero-initialized, so it starts from the global prior, and its hidden width gives it the same number of parameters as the prototype path.} All variants were retrained with the same conditioner, instance adaptation, joint flow, data, and budget. Eval uses \revc{manual click and manual prompt}; force errors cover \nValForce{} instrumented sequences. The full model has the lowest mean peak and profile errors, while ATE varies little across priors (Table~\ref{tab:prior-stages}). The peak-MAE ordering also holds on \EvalP (Table~\ref{tab:hard-ablations}).

\begin{table}[htbp]
\centering\scriptsize\setlength{\tabcolsep}{5pt}
\caption{\textbf{Prior parameterization and routing.} With instance adaptation and joint flow fixed, learned prototypes with routing supervision achieve the lowest force errors.}
\label{tab:mechanism}
\begin{tabular}{lccc}
\toprule
Prior & Peak MAE (N) $\downarrow$ & Peak RMSE (N) $\downarrow$ & Profile RMSE (N) $\downarrow$ \\
\midrule
Direct MLP prior (matched size) & $\mechMlpMae\pm\mechMlpMaeSd$ & $\mechMlpRmse\pm\mechMlpRmseSd$ & $\mechMlpProf\pm\mechMlpProfSd$ \\
Global prior + adaptation & $\mechGlobalMae\pm\mechGlobalMaeSd$ & $\mechGlobalRmse\pm\mechGlobalRmseSd$ & $\mechGlobalProf\pm\mechGlobalProfSd$ \\
Prototypes, routing loss off & $\mechNorouteMae\pm\mechNorouteMaeSd$ & $\mechNorouteRmse\pm\mechNorouteRmseSd$ & $\mechNorouteProf\pm\mechNorouteProfSd$ \\
Prototypes + routing loss (\method) & $\oursMae\pm\oursMaeSd$ & $\oursRmse\pm\oursRmseSd$ & $\profileRmse\pm\profileRmseSd$ \\
\bottomrule
\end{tabular}
\end{table}

\begin{table}[htbp]
\centering\scriptsize\setlength{\tabcolsep}{5pt}
\caption{\textbf{Prior-stage readouts.} Peak MAE read from the same checkpoint before adaptation (prior), after adaptation (adapted), and after the joint flow (final); ATE is the final trajectory error. Final RMSEs are in Table~\ref{tab:mechanism}.}
\label{tab:prior-stages}
\begin{tabular}{lcccc}
\toprule
 & ATE (cm) $\downarrow$ & \multicolumn{3}{c}{Peak MAE (N) $\downarrow$} \\
\cmidrule(lr){3-5}
Prior & Final & Prior & Adapted & Final \\
\midrule
Direct MLP & $\mechMlpAte\pm\mechMlpAteSd$ & $\mechMlpPriorMae\pm\mechMlpPriorMaeSd$ & $\mechMlpCorrMae\pm\mechMlpCorrMaeSd$ & $\mechMlpMae\pm\mechMlpMaeSd$ \\
Global prior & $\mechGlobalAte\pm\mechGlobalAteSd$ & $\mechGlobalPriorMae\pm\mechGlobalPriorMaeSd$ & $\mechGlobalCorrMae\pm\mechGlobalCorrMaeSd$ & $\mechGlobalMae\pm\mechGlobalMaeSd$ \\
Prototypes, no routing loss & $\mechNorouteAte\pm\mechNorouteAteSd$ & $\mechNoroutePriorMae\pm\mechNoroutePriorMaeSd$ & $\mechNorouteCorrMae\pm\mechNorouteCorrMaeSd$ & $\mechNorouteMae\pm\mechNorouteMaeSd$ \\
Prototypes (\method) & $\oursAte\pm\oursAteSd$ & $\oursPriorMae\pm\oursPriorMaeSd$ & $\oursAdaptMae\pm\oursAdaptMaeSd$ & $\oursMae\pm\oursMaeSd$ \\
\bottomrule
\end{tabular}
\end{table}

\textbf{Prior and instance adaptation.}
The full model's prior has the lowest mean peak MAE at every stage (Table~\ref{tab:prior-stages}). Adaptation accounts for most of its peak-error reduction: peak RMSE falls from $\oursPriorRmse$ to $\oursAdaptRmse$\,N, close to the final $\oursRmse$\,N. The global-prior variant relies more on the flow, which lowers peak MAE from $\mechGlobalCorrMae$ to $\mechGlobalMae$\,N. These are intermediate readouts of trained models, not ablations retrained without adaptation or flow; Table~\ref{tab:flow-full} separately compares prediction heads.

\textbf{Routing and prototype organization.}
Without the routing loss, only $\mechNorouteHardEntries\pm\mechNorouteHardEntriesSd$ of ten prototypes receive the largest weight for any Eval sequence, compared with all ten in the full model. The mean effective number per sequence, $\exp(-\sum_k\pi_k\log\pi_k)$, falls from $\mechFullEff\pm\mechFullEffSd$ to $\mechNorouteEff\pm\mechNorouteEffSd$, indicating more concentrated assignments. The semantic diagnostic finds stronger alignment with measured-force terciles than with object--action categories on labelled training \revc{sequences} (adjusted mutual information $\amiForceTrain$ vs.\ $\amiCatTrain$). \reve{On Eval, prototype levels correlate with their members' median measured peaks (Spearman $\spearmanLevel$), with force and category alignment at $\amiForceVal$ and $\amiCatVal$. These diagnostics support an effort-based organization, most clearly on the training sequences.}

\textbf{\revr{Within-category accuracy.}}
\input{tables/rebuttal_within_category}
\revp{Table~\ref{tab:within-category} splits the peak error by object class. On drawers and cupboard doors the priors are close: the category mean is better on drawers and the prototypes are better on cupboard doors. The priors differ on the heavy interactions and on the objects outside the category list (``Other''). At or above 20\,N, the prototypes reach \frngHeavyOurs\,N, against \frngHeavyGlobal\,N for the global prior and \frngHeavyCatMean\,N for the category mean (Table~\ref{tab:force-range}).}

\input{tables/rebuttal_force_range}
\revr{Among the designs of Table~\ref{tab:force-range}, \method{} is the only one that covers low and high forces without a category input at test time. The word lists reach the high forces too, but only by selecting from a predefined list, either by annotation or by a language model that picks the entry from the prompt. The flat clusters reach them by assigning the prompt to a cluster. These clusters are fit on all prompts, including those of the evaluation sequences. Without the category or cluster, the word lists and the flat clusters have a higher peak error than the global prior (Table~\ref{tab:organizations}). On objects outside the list, both word lists have a higher error than \method{} (``Other'' in Table~\ref{tab:within-category}).}

\subsection{Prior contribution with limited force supervision}
\label{app:sensory}
\textbf{Measured subsets.}
The study reduces force supervision while retaining the video pool of \nTrain{} sequences and its motion labels. Nested subsets of 200 measured training episodes are stratified by object and action using one seeded permutation. An episode is retained or withheld as a whole, including both camera views and its paired human references. Withheld measurements enter neither the losses nor the paired targets, prior initialization, or normalization; their videos remain available without force labels. The three seeds measure training variation on these fixed subsets, not variation across independent subset draws.

\textbf{Training controls.}
Within each prior family, the architecture, prototype count where applicable, and schedule are fixed across fractions. The global batch size is 128 throughout. \reve{At 100\,\%, Tables~\ref{tab:sensory} and~\ref{tab:sensory-full} report the main \method{} model and the global prior of Table~\ref{tab:prior-types}, both trained with all \ssRecordsHundred{} measured sequences.}

\begin{table}[htbp]
\centering\scriptsize\setlength{\tabcolsep}{3pt}
\caption{\textbf{Limited force supervision, all metrics.} Measured episodes (sequences) per fraction: 12\,\% = \ssEpisodesTwelve{} (\ssRecordsTwelve), 25\,\% = \ssEpisodesTwentyFive{} (\ssRecordsTwentyFive), 50\,\% = \ssEpisodesFifty{} (\ssRecordsFifty), 100\,\% = \ssEpisodesHundred{} (\ssRecordsHundred).}
\label{tab:sensory-full}
\begin{tabular}{llcccccc}
\toprule
 & & \multicolumn{4}{c}{Eval, force in N $\downarrow$} & \multicolumn{2}{c}{\EvalP{}, force in N $\downarrow$} \\
\cmidrule(lr){3-6}\cmidrule(lr){7-8}
Prior & Fraction & ATE (cm) & Peak MAE & Peak RMSE & Profile RMSE & Peak MAE & Peak RMSE \\
\midrule
Global & 12\,\% & $\ssGlobalTwelveAte\pm\ssGlobalTwelveAteSd$ & $\ssGlobalTwelveMae\pm\ssGlobalTwelveMaeSd$ & $\ssGlobalTwelveRmse\pm\ssGlobalTwelveRmseSd$ & $\ssGlobalTwelveProf\pm\ssGlobalTwelveProfSd$ & $\ssGlobalTwelveHardMae\pm\ssGlobalTwelveHardMaeSd$ & $\ssGlobalTwelveHardRmse\pm\ssGlobalTwelveHardRmseSd$ \\
 & 25\,\% & $\ssGlobalTwentyFiveAte\pm\ssGlobalTwentyFiveAteSd$ & $\ssGlobalTwentyFiveMae\pm\ssGlobalTwentyFiveMaeSd$ & $\ssGlobalTwentyFiveRmse\pm\ssGlobalTwentyFiveRmseSd$ & $\ssGlobalTwentyFiveProf\pm\ssGlobalTwentyFiveProfSd$ & $\ssGlobalTwentyFiveHardMae\pm\ssGlobalTwentyFiveHardMaeSd$ & $\ssGlobalTwentyFiveHardRmse\pm\ssGlobalTwentyFiveHardRmseSd$ \\
 & 50\,\% & $\ssGlobalFiftyAte\pm\ssGlobalFiftyAteSd$ & $\ssGlobalFiftyMae\pm\ssGlobalFiftyMaeSd$ & $\ssGlobalFiftyRmse\pm\ssGlobalFiftyRmseSd$ & $\ssGlobalFiftyProf\pm\ssGlobalFiftyProfSd$ & $\ssGlobalFiftyHardMae\pm\ssGlobalFiftyHardMaeSd$ & $\ssGlobalFiftyHardRmse\pm\ssGlobalFiftyHardRmseSd$ \\
 & 100\,\% & $\mechGlobalAte\pm\mechGlobalAteSd$ & $\mechGlobalMae\pm\mechGlobalMaeSd$ & $\mechGlobalRmse\pm\mechGlobalRmseSd$ & $\mechGlobalProf\pm\mechGlobalProfSd$ & $\mechGlobalHardMae\pm\mechGlobalHardMaeSd$ & $\mechGlobalHardRmse\pm\mechGlobalHardRmseSd$ \\
\midrule
Prototypes & 12\,\% & $\ssCodebookTwelveAte\pm\ssCodebookTwelveAteSd$ & $\ssCodebookTwelveMae\pm\ssCodebookTwelveMaeSd$ & $\ssCodebookTwelveRmse\pm\ssCodebookTwelveRmseSd$ & $\ssCodebookTwelveProf\pm\ssCodebookTwelveProfSd$ & $\ssCodebookTwelveHardMae\pm\ssCodebookTwelveHardMaeSd$ & $\ssCodebookTwelveHardRmse\pm\ssCodebookTwelveHardRmseSd$ \\
 & 25\,\% & $\ssCodebookTwentyFiveAte\pm\ssCodebookTwentyFiveAteSd$ & $\ssCodebookTwentyFiveMae\pm\ssCodebookTwentyFiveMaeSd$ & $\ssCodebookTwentyFiveRmse\pm\ssCodebookTwentyFiveRmseSd$ & $\ssCodebookTwentyFiveProf\pm\ssCodebookTwentyFiveProfSd$ & $\ssCodebookTwentyFiveHardMae\pm\ssCodebookTwentyFiveHardMaeSd$ & $\ssCodebookTwentyFiveHardRmse\pm\ssCodebookTwentyFiveHardRmseSd$ \\
 & 50\,\% & $\ssCodebookFiftyAte\pm\ssCodebookFiftyAteSd$ & $\ssCodebookFiftyMae\pm\ssCodebookFiftyMaeSd$ & $\ssCodebookFiftyRmse\pm\ssCodebookFiftyRmseSd$ & $\ssCodebookFiftyProf\pm\ssCodebookFiftyProfSd$ & $\ssCodebookFiftyHardMae\pm\ssCodebookFiftyHardMaeSd$ & $\ssCodebookFiftyHardRmse\pm\ssCodebookFiftyHardRmseSd$ \\
 & 100\,\% & $\oursAte\pm\oursAteSd$ & $\oursMae\pm\oursMaeSd$ & $\oursRmse\pm\oursRmseSd$ & $\profileRmse\pm\profileRmseSd$ & $\oursHardMae\pm\oursHardMaeSd$ & $\oursHardRmse\pm\oursHardRmseSd$ \\
\bottomrule
\end{tabular}
\end{table}

The learned prototypes have lower mean peak MAE than the global prior at every measured fraction (Table~\ref{tab:sensory-full}). With half the measured episodes, their peak MAE is below that of the full-data global prior (\ssCodebookFiftyMae{} vs.\ \mechGlobalMae\,N)\reve{, and their profile RMSE stays within \profscaled{0.3}\,N of it (\ssCodebookFiftyProf{} vs.\ \mechGlobalProf\,N), so half the measurements suffice for peak prediction. \revr{Table~\ref{tab:sensory-by-k} repeats the study with fewer prototypes, and Table~\ref{tab:k-stage-range-labels} reads it by stage and force range for every share of training force labels.}}

\input{tables/rebuttal_limited_supervision_by_k}

\input{tables/rebuttal_k_stage_force_range_by_labels}

\subsection{Joint prediction and motion cues}
\label{app:joint-prediction}
\textbf{Prediction heads.}
The study retains the learned prototype prior and instance adaptation and changes the prediction head or its motion cues (Table~\ref{tab:flow-full}). The one-shot head directly regresses the standardized residual target with the same network architecture. The split-flow variant processes trajectory and force tokens in separate streams, retaining the shared conditioner but removing direct exchange between the two outputs\reve{; a block-diagonal self-attention mask restricts waypoint and force tokens to their own stream without changing the parameters}. All learned variants use the same data and training budget and are evaluated with \revc{manual click and manual prompt} over three seeds.

\begin{table}[htbp]
\centering\scriptsize\setlength{\tabcolsep}{5pt}
\caption{\textbf{Prediction heads and motion cues, full comparison.} The tracked initialization is decoded without refinement; the middle block replaces the joint flow with other prediction heads, and the lower block disables click-free training throughout.}
\label{tab:flow-full}
\begin{tabular}{lcccc}
\toprule
& & \multicolumn{3}{c}{Force error (N) $\downarrow$} \\
\cmidrule(lr){3-5}
Configuration & ATE (cm) $\downarrow$ & Peak MAE & Peak RMSE & Profile RMSE \\
\midrule
Tracked initialization & \flowInitAte & -- & -- & -- \\
Joint flow (\method) & $\oursAte\pm\oursAteSd$ & $\oursMae\pm\oursMaeSd$ & $\oursRmse\pm\oursRmseSd$ & $\profileRmse\pm\profileRmseSd$ \\
\midrule
One-shot regression & $\flowRegressAte\pm\flowRegressAteSd$ & $\flowRegressMae\pm\flowRegressMaeSd$ & $\flowRegressRmse\pm\flowRegressRmseSd$ & $\flowRegressProf\pm\flowRegressProfSd$ \\
Split trajectory / force flows & $\flowSplitAte\pm\flowSplitAteSd$ & $\flowSplitMae\pm\flowSplitMaeSd$ & $\flowSplitRmse\pm\flowSplitRmseSd$ & $\flowSplitProf\pm\flowSplitProfSd$ \\
\midrule
All motion cues (control) & $\flowAprimeAte\pm\flowAprimeAteSd$ & $\flowAprimeMae\pm\flowAprimeMaeSd$ & $\flowAprimeRmse\pm\flowAprimeRmseSd$ & $\flowAprimeProf\pm\flowAprimeProfSd$ \\
No lattice field; tracker tokens kept & $\flowNolatAte\pm\flowNolatAteSd$ & $\flowNolatMae\pm\flowNolatMaeSd$ & $\flowNolatRmse\pm\flowNolatRmseSd$ & $\flowNolatProf\pm\flowNolatProfSd$ \\
No dense motion field & $\flowNofieldAte\pm\flowNofieldAteSd$ & $\flowNofieldMae\pm\flowNofieldMaeSd$ & $\flowNofieldRmse\pm\flowNofieldRmseSd$ & $\flowNofieldProf\pm\flowNofieldProfSd$ \\
\bottomrule
\end{tabular}
\end{table}

\textbf{Prediction-head comparison.}
\reve{The joint flow of \method{} reaches $\oursAte\pm\oursAteSd$\,cm ATE and $\oursMae\pm\oursMaeSd$\,N peak MAE, improving on one-shot regression ($\flowRegressAte$\,cm, $\flowRegressMae$\,N) and on split trajectory and force flows ($\flowSplitAte$\,cm, $\flowSplitMae$\,N), with lower profile RMSE than both ($\profileRmse$ against $\flowRegressProf$ and $\flowSplitProf$\,N). These results support flow-based refinement that predicts both outputs together.}

\textbf{Motion cues.}
The dense-field ablation removes both lattice-field and tracker tokens. Because it cannot train the click-free path, its control also disables that path while retaining all motion cues. Removing the field increases both trajectory and force errors relative to this control. Retaining tracker tokens while removing only the lattice largely recovers trajectory accuracy ($\flowNolatAte$ vs.\ $\flowAprimeAte$\,cm)\reve{, so the tracker tokens and the lattice carry overlapping \revf{motion cues}. The study therefore supports the motion cues as a whole.}

\textbf{Re-framed evaluation.}
Table~\ref{tab:hard-ablations} reports the same variants on \EvalP. Removing \revf{the dense motion field} increases ATE from $\flowAprimeHardAte$ to $\flowNofieldHardAte$\,cm relative to its control, consistent with the Eval result.

\subsection{Additional ablations and prototype count}
\label{app:ksweep}
Table~\ref{tab:hard-ablations} extends the prior and flow comparisons of Tables~\ref{tab:mechanism} and~\ref{tab:flow-full} to \EvalP. Table~\ref{tab:ksweep} varies the number of prototypes at fixed recipe. Across $K\in\{5,10,15,20,30\}$, \reve{mean peak MAE stays between \kTenMae{} and \kFifteenMae\,N, so accuracy is stable across prototype counts.} Every prototype is used at $K{=}5$ and $K{=}10$, while larger settings leave some unused. We retain $K{=}10$, the largest tested setting with full usage. \revr{Table~\ref{tab:ksweep-torque} repeats the comparison for sensor-frame torque with $K{=}10$ and $K{=}20$.} \revr{Table~\ref{tab:k-stage-range} reads the peak error for \revp{$K{=}1$, $2$, $5$ and $10$} by stage and force range.} \revp{The prototype levels are mixed in the log domain (Sec.~\ref{sec:prior}). Mixing them arithmetically at test time, with all trained weights kept, gives nearly the same error: peak MAE $\mixArithMae\pm\mixArithMaeSd$ against $\oursMae\pm\oursMaeSd$\,N, and peak RMSE $\mixArithRmse\pm\mixArithRmseSd$ against $\oursRmse\pm\oursRmseSd$\,N.}

\begin{table}[htbp]
\centering\scriptsize\setlength{\tabcolsep}{5pt}
\caption{\textbf{Mechanism and flow arms on Eval$'$.} 594 views, 288 of them instrumented.}
\label{tab:hard-ablations}
\begin{tabular}{lcccc}
\toprule
& & \multicolumn{3}{c}{Force error (N) $\downarrow$} \\
\cmidrule(lr){3-5}
Configuration & ATE (cm) $\downarrow$ & Peak MAE & Peak RMSE & Profile RMSE \\
\midrule
Direct MLP prior & $\mechMlpHardAte\pm\mechMlpHardAteSd$ & $\mechMlpHardMae\pm\mechMlpHardMaeSd$ & $\mechMlpHardRmse\pm\mechMlpHardRmseSd$ & $\mechMlpHardProf\pm\mechMlpHardProfSd$ \\
Global prior + adaptation & $\mechGlobalHardAte\pm\mechGlobalHardAteSd$ & $\mechGlobalHardMae\pm\mechGlobalHardMaeSd$ & $\mechGlobalHardRmse\pm\mechGlobalHardRmseSd$ & $\mechGlobalHardProf\pm\mechGlobalHardProfSd$ \\
Prototypes, routing loss off & $\mechNorouteHardAte\pm\mechNorouteHardAteSd$ & $\mechNorouteHardMae\pm\mechNorouteHardMaeSd$ & $\mechNorouteHardRmse\pm\mechNorouteHardRmseSd$ & $\mechNorouteHardProf\pm\mechNorouteHardProfSd$ \\
One-shot regression head & $\flowRegressHardAte\pm\flowRegressHardAteSd$ & $\flowRegressHardMae\pm\flowRegressHardMaeSd$ & $\flowRegressHardRmse\pm\flowRegressHardRmseSd$ & $\flowRegressHardProf\pm\flowRegressHardProfSd$ \\
Split trajectory / force streams & $\flowSplitHardAte\pm\flowSplitHardAteSd$ & $\flowSplitHardMae\pm\flowSplitHardMaeSd$ & $\flowSplitHardRmse\pm\flowSplitHardRmseSd$ & $\flowSplitHardProf\pm\flowSplitHardProfSd$ \\
Control: no click-free path & $\flowAprimeHardAte\pm\flowAprimeHardAteSd$ & $\flowAprimeHardMae\pm\flowAprimeHardMaeSd$ & $\flowAprimeHardRmse\pm\flowAprimeHardRmseSd$ & $\flowAprimeHardProf\pm\flowAprimeHardProfSd$ \\
No lattice field, tracker tokens kept & $\flowNolatHardAte\pm\flowNolatHardAteSd$ & $\flowNolatHardMae\pm\flowNolatHardMaeSd$ & $\flowNolatHardRmse\pm\flowNolatHardRmseSd$ & $\flowNolatHardProf\pm\flowNolatHardProfSd$ \\
No dense motion field & $\flowNofieldHardAte\pm\flowNofieldHardAteSd$ & $\flowNofieldHardMae\pm\flowNofieldHardMaeSd$ & $\flowNofieldHardRmse\pm\flowNofieldHardRmseSd$ & $\flowNofieldHardProf\pm\flowNofieldHardProfSd$ \\
\method & $\oursHardAte\pm\oursHardAteSd$ & $\oursHardMae\pm\oursHardMaeSd$ & $\oursHardRmse\pm\oursHardRmseSd$ & $\flowAHardProf\pm\flowAHardProfSd$ \\
\bottomrule
\end{tabular}
\end{table}

\setcounter{topnumber}{3}\renewcommand{\floatpagefraction}{0.9}
\input{tables/tab_ksweep}

\input{tables/tab_ksweep_torque}

\input{tables/rebuttal_k_stage_force_range}

\subsection{Inference start point}
\label{app:flowstart}
\rev{The flow is integrated from $\mathbf y_0=\mathbf 0$ (Sec.~\ref{sec:flow}), whereas training draws $\mathbf y_0\sim\mathcal N(\mathbf 0,\mathbf I)$. Table~\ref{tab:flowstart} repeats the Eval protocol with the same checkpoints and 20 Euler steps but starts from sampled noise in the standardized state space, with five fixed draws per sequence: \emph{single draw} reports each seed's mean over the five draws, and \emph{five-draw average} decodes the mean of the five residual states (clipping each draw before averaging changes no number by more than 0.002). Force errors are equal within noise: the differences are below the draw-to-draw SD of 0.07\,N and the seed SD of 0.38\,N. A single draw is 0.16\,cm worse in ATE on every seed (draw-to-draw SD 0.18\,cm), and the five-draw average matches the zero start, so the origin acts as the ensemble mean at one fifth of the cost. The sampled starts have norm $8.1\pm1.1$ over the dimensions the network reads, against 0 for the zero start.}

\begin{table}[htbp]
\centering\scriptsize\setlength{\tabcolsep}{5pt}
\caption{\textbf{Start point of the flow at inference.} Zero is the paper's setting; the noise rows start from $\mathbf y_0\sim\mathcal N(\mathbf 0,\mathbf I)$ with five fixed draws per sequence.}
\label{tab:flowstart}
\begin{tabular}{lcccc}
\toprule
& & \multicolumn{3}{c}{Force error (N) $\downarrow$} \\
\cmidrule(lr){3-5}
Start & ATE (cm) $\downarrow$ & Peak MAE & Peak RMSE & Profile RMSE \\
\midrule
Zero (ref.) & $6.6\pm0.4$ & $4.81\pm0.38$ & $5.92\pm0.42$ & \profscaled{$5.52\pm0.25$} \\
Noise, single draw & $6.8\pm0.5$ & $4.78\pm0.36$ & $5.90\pm0.41$ & \profscaled{$5.50\pm0.24$} \\
Noise, five-draw average & $6.6\pm0.5$ & $4.77\pm0.36$ & $5.89\pm0.41$ & \profscaled{$5.47\pm0.24$} \\
\bottomrule
\end{tabular}
\end{table}

\subsection{Transfer to human videos}
\label{app:transfer}
\textbf{Reference agreement.}
The \nHandRef{} matched Eval human recordings each have a same-episode gripper partner with measured force in Eval. Table~\ref{tab:transfer} compares the human-video prediction with that partner's peak and 32-sample profile, aligned by normalized interaction time. \reve{The comparison thus measures how well the measured gripper force transfers to the human video of the same interaction.} GPT-5.5 uses the \revc{questions} of Appendix~\ref{app:baselines}; the frames-and-text variant names the hand as the contact.

\begin{table}[htbp]
\centering\scriptsize\setlength{\tabcolsep}{1.5pt}
\caption{\textbf{Human video against the paired gripper reference.} \nHandRef{} matched evaluation recordings; peak errors and bias (mean signed difference) compare the predicted peak with the partner's measured peak, profile RMSE compares the 32-sample profiles by normalized time, and GPT-5.5 estimates the peak only. Gripper MAE is each model's peak MAE on the \nValForce{} measured gripper sequences.}
\label{tab:transfer}
\begin{tabular}{lccccc}
\toprule
 & & \multicolumn{4}{c}{Human video vs.\ paired reference (N) $\downarrow$} \\
\cmidrule(lr){3-6}
Model & Gripper MAE (N) $\downarrow$ & Peak MAE & Peak RMSE & Profile RMSE & Bias \\
\midrule
\method{} (paired references) & $\oursMae\pm\oursMaeSd$ & $\oursHandRef\pm\oursHandRefSd$ & $\oursHandRefRmse\pm\oursHandRefRmseSd$ & $\oursHandRefProf\pm\oursHandRefProfSd$ & $\oursHandBias\pm\oursHandBiasSd$ \\
Same recipe, no paired references & $\nopairMae\pm\nopairMaeSd$ & $\nopairHandRef\pm\nopairHandRefSd$ & $\nopairHandRefRmse\pm\nopairHandRefRmseSd$ & $\nopairHandRefProf\pm\nopairHandRefProfSd$ & $\nopairHandBias\pm\nopairHandBiasSd$ \\
ForceSight adapted (third-person), paired references & $\fsLCropMae\pm\fsLCropMaeSd$ & $\fsLCropHandRef\pm\fsLCropHandRefSd$ & $\fsLCropHandRefRmse\pm\fsLCropHandRefRmseSd$ & $\fsLCropHandRefProf\pm\fsLCropHandRefProfSd$ & $\fsLCropHandBias\pm\fsLCropHandBiasSd$ \\
ForceSight adapted (third-person), measured rows only & $\fsLGripCropMae\pm\fsLGripCropMaeSd$ & $\fsLGripCropHandRef\pm\fsLGripCropHandRefSd$ & $\fsLGripCropHandRefRmse\pm\fsLGripCropHandRefRmseSd$ & $\fsLGripCropHandRefProf\pm\fsLGripCropHandRefProfSd$ & $\fsLGripCropHandBias\pm\fsLGripCropHandBiasSd$ \\
ForceSight released checkpoint (one run) & \fsZsCropMae & \fsZsCropHandRef & \fsZsCropHandRefRmse & \fsZsCropHandRefProf & $\fsZsCropHandBias$ \\
GPT-5.5 (8 frames + text) & \maeGpt & \gptHandRef & \gptHandRefRmse & -- & $\gptHandBias$ \\
GPT-5.5 (8 frames) & \maeGptImg & \gptImgHandRef & \gptImgHandRefRmse & -- & $\gptImgHandBias$ \\
\bottomrule
\end{tabular}
\end{table}

\textbf{\reve{Ablation setup.}}
\reve{The ablated model is trained on the same \nTrain{} sequences as \method{}, with the same videos, trajectory labels, schedule, and settings; only the paired force targets of the \nPaired{} human training sequences are removed. Without them, these sequences still supervise the trajectory and motion but receive no measured force. The self-distillation teacher of the training schedule (Appendix~\ref{app:impl:train}) is kept: from update 2{,}000 its predictions become the force target of every training sequence, the human ones included, but in this run the teacher is the ablated model's own average and has never seen a paired reference. The comparison therefore isolates what the paired gripper force adds rather than the effect of less training data.}

\textbf{Effect of paired supervision.}
Removing paired targets changes measured-gripper peak MAE little ($\nopairMae$ vs.\ $\oursMae$\,N), but raises human-video peak disagreement from $\oursHandRef$ to $\nopairHandRef$\,N and mean signed bias from $\oursHandBias$ to $\nopairHandBias$\,N. Profile disagreement also increases ($\oursHandRefProf$ to $\nopairHandRefProf$\,N). On measured gripper recordings, profile RMSE rises from $\profileRmse\pm\profileRmseSd$ to $\nopairProf\pm\nopairProfSd$\,N; on \EvalP, peak MAE remains similar ($\nopairHardMae\pm\nopairHardMaeSd$\,N without pairing, $\pairHardMae\pm\pairHardMaeSd$\,N with it).

\textbf{Comparison with other predictors.}
\reve{\method{} improves on adapted ForceSight with paired supervision in every reference \profscaled{peak} metric, with a peak RMSE of $\oursHandRefRmse$ against $\fsLCropHandRefRmse$\,N and a mean bias of $\oursHandBias$ against $\fsLCropHandBias$\,N.} \reve{GPT-5.5 estimates the peak only: it reaches $\gptHandRef$\,N with text and $\gptImgHandRef$\,N from frames alone on the reference peaks, with $\maeGpt$ and $\maeGptImg$\,N on measured gripper recordings against $\oursMae$\,N for \method{}.} \profscaled{Its reference profile RMSE ($\oursHandRefProf\pm\oursHandRefProfSd$\,N) is above that of adapted ForceSight ($\fsLCropHandRefProf\pm\fsLCropHandRefProfSd$\,N).} \revr{Table~\ref{tab:embodiment-gpt} compares \method{} and GPT-5.5 on the same \nHandRef{} episodes from the gripper video and from the human video, for force and for sensor-frame torque.}
\input{tables/rebuttal_embodiment_gpt}

\setcounter{topnumber}{2}\renewcommand{\floatpagefraction}{0.5}
\subsection{Prior types}
\label{app:prior-designs}
\textbf{Four prior families.}
We compare a \emph{global prior}, which supplies one shared peak and normalized profile; a \emph{word list} with a fixed vocabulary, which groups measurements by four object names and two actions \revr{(the GT category of the sequence, or one that a language model assigns from this list based on the prompt)}; \emph{flat clusters}, which assign each \revc{prompt} to its nearest $k$-means cluster ($K{=}10$); and \emph{learned prototypes}, whose force levels, profiles, and semantic keys are optimized with soft routing from video and text ($K{=}10$; Sec.~\ref{sec:prior}). The reference designs share the motion representation, instance adaptation, and joint flow; Table~\ref{tab:organizations} compares them.

\textbf{Construction and controls.}
Force statistics use training \revc{sequences} only. The clustering reference uses all \revc{prompts}, including those of evaluation \revc{sequences}, to construct its groups. Category and cluster tokens are dropped for part of training so that these references also run from the \revc{prompt} alone.

\begin{table}[htbp]
\centering\footnotesize\setlength{\tabcolsep}{4pt}
\caption{\textbf{Prior types.} Manual click and manual prompt; peak MAE and profile RMSE of these rows also appear in Table~\ref{tab:prior-types}. \revr{\emph{Category input}: what the prior receives at test time besides the video, click and prompt. \emph{Withheld}: peak MAE of the same checkpoints with that category withheld at test time.}}
\label{tab:organizations}
\resizebox{\linewidth}{!}{%
\begin{tabular}{llccccc}
\toprule
& & & \multicolumn{3}{c}{Force error (N) $\downarrow$} & \revr{Withheld} \\
\cmidrule(lr){4-6}
Prior type & \revr{Category input at test time} & ATE (cm) $\downarrow$ & Peak MAE & Peak RMSE & Profile RMSE & \revr{Peak MAE} \\
\midrule
Global prior + adaptation & \revr{none} & $\mechGlobalAte\pm\mechGlobalAteSd$ & $\mechGlobalMae\pm\mechGlobalMaeSd$ & $\mechGlobalRmse\pm\mechGlobalRmseSd$ & $\mechGlobalProf\pm\mechGlobalProfSd$ & \revr{--} \\
Word list (categorical), LLM-assigned labels & \revr{fixed list; category read from the prompt by the language model} & $\orgLlmAte\pm\orgLlmAteSd$ & $\orgLlmMae\pm\orgLlmMaeSd$ & $\orgLlmRmse\pm\orgLlmRmseSd$ & $\orgLlmProf\pm\orgLlmProfSd$ & \revr{$\catWithheldLlm\pm\catWithheldLlmSd$} \\
Word list (categorical), annotated labels & \revr{fixed list; annotated category of the sequence} & $\orgAnnotAte\pm\orgAnnotAteSd$ & $\orgAnnotMae\pm\orgAnnotMaeSd$ & $\orgAnnotRmse\pm\orgAnnotRmseSd$ & $\orgAnnotProf\pm\orgAnnotProfSd$ & \revr{$\catWithheldAnnot\pm\catWithheldAnnotSd$} \\
Flat clusters (k-means, $K{=}10$) & \revr{nearest cluster of the prompt (clusters fit on all prompts)} & $\orgKmeansAte\pm\orgKmeansAteSd$ & $\orgKmeansMae\pm\orgKmeansMaeSd$ & $\orgKmeansRmse\pm\orgKmeansRmseSd$ & $\orgKmeansProf\pm\orgKmeansProfSd$ & \revr{$\catWithheldKmeans\pm\catWithheldKmeansSd$} \\
\midrule
Learned prototypes (\method) & \revr{none} & $\oursAte\pm\oursAteSd$ & $\oursMae\pm\oursMaeSd$ & $\oursRmse\pm\oursRmseSd$ & $\profileRmse\pm\profileRmseSd$ & \revr{--} \\
\bottomrule
\end{tabular}}
\end{table}

\subsection{Additional torque results}
\label{app:torque}
Torque models differ from their force counterparts only in the supervised channel: sensor-frame torque magnitude from the same wrench sensor. They use the same sequences and training recipe, with prototypes learned separately for torque. Targets are scaled so that the Huber threshold has the same relative position as for force and are reported in N\,m. Table~\ref{tab:torque-full} compares the retained prior families.

\begin{table}[htbp]
\centering\scriptsize\setlength{\tabcolsep}{5pt}
\caption{\textbf{Sensor-frame torque.} Comparison of GPT-5.5 with models trained using alternative prior types.}
\label{tab:torque-full}
\begin{tabular}{lccc}
\toprule
& \multicolumn{3}{c}{Torque error (N\,m) $\downarrow$} \\
\cmidrule(lr){2-4}
Model & Peak MAE & Peak RMSE & Profile RMSE \\
\midrule
GPT-5.5 (8 frames) & \tqGptImg & \tqGptImgRmse & -- \\
GPT-5.5 (frames + text) & \tqGpt & \tqGptRmse & -- \\
\midrule
Global prior & $\tqGlobal\pm\tqGlobalSd$ & $\tqGlobalRmse\pm\tqGlobalRmseSd$ & $\tqGlobalProf\pm\tqGlobalProfSd$ \\
Word list, annotated labels & $\tqAnnot\pm\tqAnnotSd$ & $\tqAnnotRmse\pm\tqAnnotRmseSd$ & $\tqAnnotProf\pm\tqAnnotProfSd$ \\
Word list, LLM-assigned labels & $\tqLlm\pm\tqLlmSd$ & $\tqLlmRmse\pm\tqLlmRmseSd$ & $\tqLlmProf\pm\tqLlmProfSd$ \\
Flat clusters (k-means) & $\tqKmeans\pm\tqKmeansSd$ & $\tqKmeansRmse\pm\tqKmeansRmseSd$ & $\tqKmeansProf\pm\tqKmeansProfSd$ \\
\midrule
\multicolumn{4}{l}{\method{}, learned prototypes} \\
\quad $K{=}10$ & $\tqOurs\pm\tqOursSd$ & $\tqOursRmse\pm\tqOursRmseSd$ & $\tqOursProf\pm\tqOursProfSd$ \\
\quad $K{=}20$ & $\tqKtwenty\pm\tqKtwentySd$ & $\tqKtwentyRmse\pm\tqKtwentyRmseSd$ & $\tqKtwentyProf\pm\tqKtwentyProfSd$ \\
\bottomrule
\end{tabular}
\end{table}

The learned formulation transfers directly from force to torque: both prototype configurations improve peak estimation over GPT-5.5 and achieve the lowest profile errors among priors that do not use annotated labels. Increasing the number of prototypes further reduces peak error, suggesting that torque benefits from a larger prototype set. \reve{Under re-framing, the prototype model stays close to its canonical result ($\tqOursHard\pm\tqOursHardSd$ against $\tqOurs\pm\tqOursSd$\,N\,m).} Text improves the GPT-5.5 estimate but does not close the gap to the trained models.

\subsection{Supplied and generated guidance}
\label{app:inputs}
\textbf{Evaluation protocol.}
Table~\ref{tab:selfclick-variants} compares, for the same checkpoints, the annotated contact with the vision--language model's proposal used directly (crop re-centered; the self-click rows of the main text) and with the proposal refined by the localization head, and the recorded \revc{prompt} with the self-written one; the tracker baselines receive the same proposal in Table~\ref{tab:trajectory}. The generation of the proposal and of the self-written \revc{prompt} is described in Appendix~\ref{app:impl:inference}. \revr{Manual-click rows sample the DA3 depth at the annotated contact pixel, and self-click rows at the contact pixel predicted by Qwen.} Table~\ref{tab:selfclick-variants} adds peak RMSE and the localization-head refinement to the four combinations of Table~\ref{tab:ablations} (\emph{Guidance}).

\begin{table}[htbp]
\centering\scriptsize\setlength{\tabcolsep}{5pt}
\caption{\textbf{Guidance variants.} Self-click uses Qwen3-VL's contact proposal directly (crop re-centred) and self-prompt its self-written prompt; contact error is the median pixel distance from the annotated contact, omitted when the contact is supplied; refinement applies the localization head to the proposal. \revr{The depth anchor is obtained from DA3 at the annotated contact pixel for manual-click runs and at the Qwen-predicted contact pixel for self-click runs.}}
\label{tab:selfclick-variants}
\begin{tabular}{llcccc}
\toprule
& & & & \multicolumn{2}{c}{Force error (N) $\downarrow$} \\
\cmidrule(lr){5-6}
Contact point & Prompt & Contact error (px) $\downarrow$ & ATE (cm) $\downarrow$ & Peak MAE & Peak RMSE \\
\midrule
manual click & \revc{manual prompt} & -- & $\oursAte\pm\oursAteSd$ & $\oursMae\pm\oursMaeSd$ & $\oursRmse\pm\oursRmseSd$ \\
self-click & \revc{manual prompt} & \selfClickRawLoc & $\revr{\selfClickRawAte\pm\selfClickRawAteSd}$ & $\revr{\selfClickRawMae\pm\selfClickRawMaeSd}$ & $\revr{\selfClickRawRmse\pm\selfClickRawRmseSd}$ \\
manual click & self-prompt & -- & $\selfPromptAte\pm\selfPromptAteSd$ & $\selfPromptMae\pm\selfPromptMaeSd$ & $\selfPromptRmse\pm\selfPromptRmseSd$ \\
self-click & self-prompt & \revr{\selfPromptSelfClickLoc} & $\revr{\selfPromptSelfClickAte\pm\selfPromptSelfClickAteSd}$ & $\revr{\selfPromptSelfClickMae\pm\selfPromptSelfClickMaeSd}$ & $\revr{\selfPromptSelfClickRmse\pm\selfPromptSelfClickRmseSd}$ \\
\midrule
self-click, refined & \revc{manual prompt} & \locPxSelfClick & $\revr{\selfClickAte\pm\selfClickAteSd}$ & $\revr{\selfClickMae\pm\selfClickMaeSd}$ & $\revr{\selfClickRmse\pm\selfClickRmseSd}$ \\
\bottomrule
\end{tabular}
\end{table}

\textbf{Contact proposals.}
With the recorded prompt, the self-click proposal lies \selfClickRawLoc\,px from the annotation on median. \reve{Localization-head refinement reduces this distance to \locPxSelfClick\,px while ATE \revr{is \selfClickAte\,cm (\selfClickRawAte\,cm without refinement)} and peak MAE \revr{is \selfClickMae\,N (\selfClickRawMae\,N without refinement)}, so the predictions are insensitive to this localization difference.} Table~\ref{tab:trajectory} compares trajectory baselines given the same proposed contact.

\textbf{Prompt sensitivity.}
\reve{With self-prompt, peak MAE stays within 1.04\,N (manual click) and 1.16\,N (self-click) of the manual prompt, and peak RMSE within 1.66 and 2.22\,N. The generated prompt identifies opening versus closing correctly in 92 of the \nVal{} sequences, including 46 of the \nValForce{} instrumented ones. With manual click, substituting predictions obtained with the recorded prompt for the two misread instrumented sequences} changes peak MAE and RMSE by at most 0.12\,N across seeds. \reve{This diagnostic points to the prompt's wording beyond its action label.} Appendix~\ref{app:rw:inputs} separately examines \revc{prompt} changes and misread actions on \revb{real-world} videos, where force is unmeasured.

\textbf{\revr{Wrong action prompts.}}
\revr{Table~\ref{tab:flipped} tests the prompt dependence directly by flipping the action word of every recorded prompt.}
\input{tables/rebuttal_flipped}

\textbf{\revr{Fully automatic inference.}}
\revr{Table~\ref{tab:automatic} separates the inputs of the manual-click and self-click settings. Self-click uses no annotation: the contact pixel comes from Qwen, the depth is sampled from DA3 at that pixel, and the crop is the fixed 0.55-frame window. Giving the manual click the same fixed crop changes little. The self-click contact lies \selfClickRawLoc\,px from the annotated contact on median; this raises ADE, mostly on gripper sequences, while the start-aligned ATE stays close to the manual click. Sampling the depth at the annotated contact pixel instead leaves all errors within seed variation.}
\input{tables/automatic_inference}
\FloatBarrier

\subsection{\texorpdfstring{\revr{Within-category evaluation and influence of execution speed}}{Within-category evaluation and influence of execution speed}}
\label{app:speed}
\renewcommand{\topfraction}{0.9}\renewcommand{\bottomfraction}{0.5}\renewcommand{\textfraction}{0.07}\renewcommand{\floatpagefraction}{0.9}
\setcounter{topnumber}{3}\setcounter{totalnumber}{4}
\makeatletter\setlength{\@fptop}{0pt}\makeatother
\revr{This section asks how the execution speed changes the measured force and the prediction. The force is studied on all measured executions of Hoi!, and the errors on the gripper recordings of Eval, overall and within each object--action group. In these recordings, the force was measured during the very execution the model sees. The hand recordings are left out, because their force comes from a different execution (Appendix~\ref{app:transfer}).}

\begin{figure}[htbp]
\centering
\includegraphics[width=\linewidth]{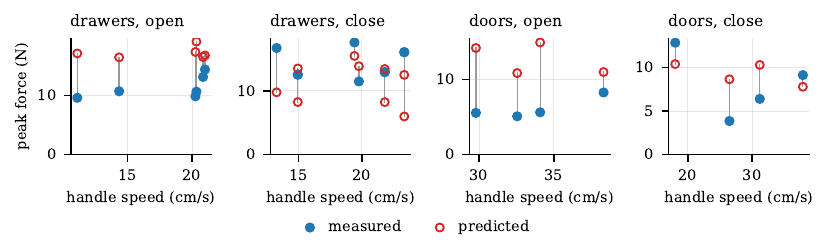}
\caption{\revr{\textbf{Repeated interactions on the same sideboard.} Measured (filled) and predicted (hollow) peak against handle speed for every recording on the living-room sideboard of Eval; a grey line joins the two values of a recording.}}
\label{fig:speed-same-object}
\end{figure}

\revr{\textbf{Speed and resistance.} The \emph{handle speed} is the length of the handle path between 5\% and 95\% of the travel, divided by the time it takes. The travel is the angle about the fitted axis for doors and the displacement along it for drawers, read from the 16 ground-truth waypoints of the handle. The \emph{running resistance} is the median force between 25\% and 75\% of the travel, and the \emph{peak} is the largest measured force; the force is aligned to the waypoints at the first contact. One scene, the living room, was recorded in several trials, so one sideboard was opened and closed repeatedly (Fig.~\ref{fig:speed-same-object}); elsewhere the speed varies across objects.}

\begin{figure}[htbp]
\centering
\includegraphics[width=\linewidth]{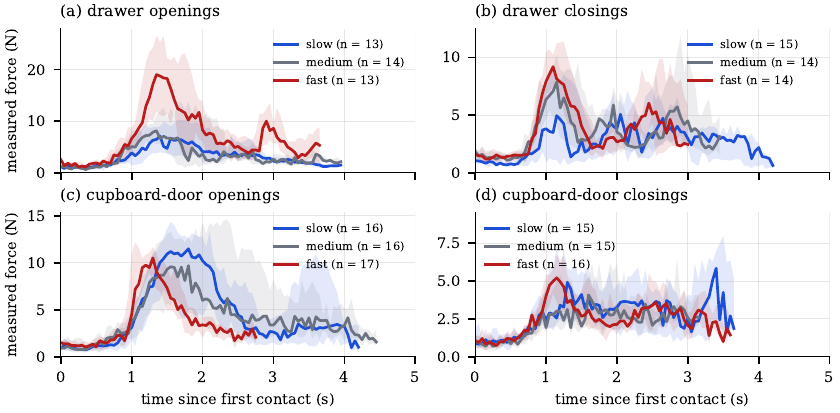}
\caption{\revr{\textbf{Measured force against time.} Drawer and cupboard-door openings and closings of Hoi! (training and Eval), each execution counted once. Lines and bands: median and interquartile range per speed third. A curve ends when fewer than a third of its executions are still in contact.}}
\label{fig:speed-compact}
\end{figure}
\begin{figure}[p]
\centering
\includegraphics[width=0.88\linewidth]{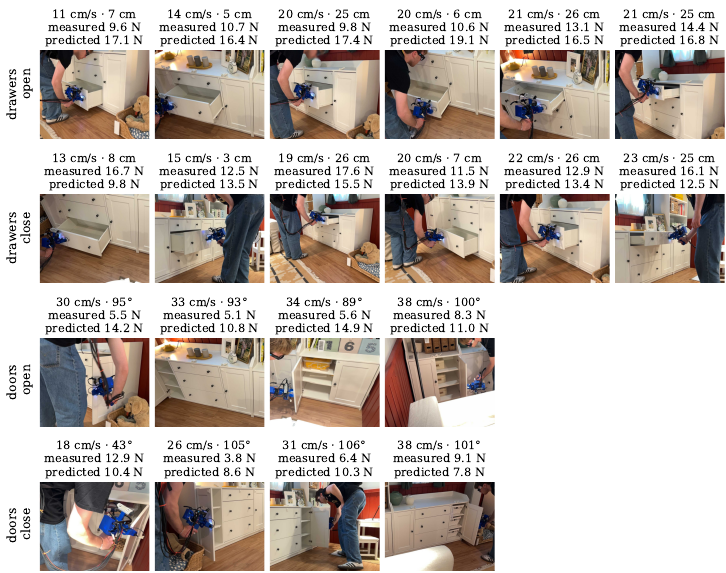}
\caption{\revr{\textbf{The executions of Fig.~\ref{fig:speed-same-object}.} One frame per execution at the open position, one row per object--action group in order of handle speed, with the handle speed, the travel and the measured and predicted peak.}}
\label{fig:speed-same-object-examples}
\vspace{5mm}
\includegraphics[width=0.88\linewidth]{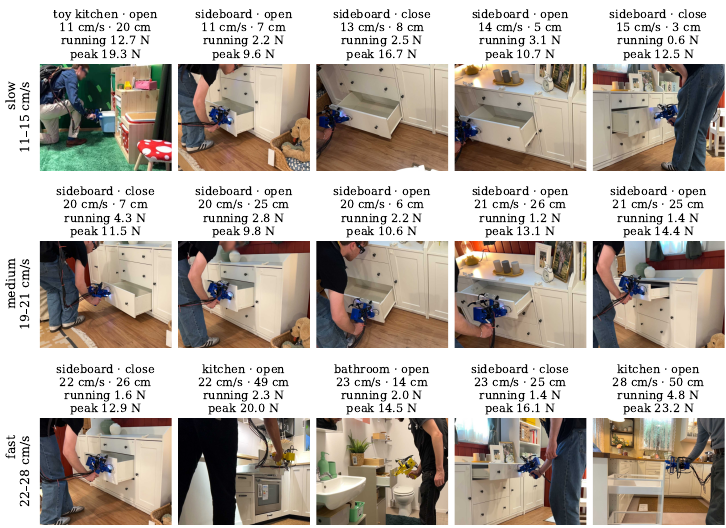}
\caption{\revr{\textbf{The measured drawer executions of Eval by speed third.} One frame per execution at the open position, with the furniture, the action, the handle speed, the travel, the running resistance and the measured peak. Rows: the slow, medium and fast third by handle speed, openings and closings together. The slowest execution of the medium third is not shown.}}
\label{fig:speed-drawers}
\end{figure}
\revr{\textbf{Speed in the measured force.} Within an object--action group of drawers, cupboard doors or oven doors, the running resistance varies by a factor of \spdFxRunIqrRatioMin{} to \spdFxRunIqrRatioMax{} across its interquartile range, so the force to predict varies within a group. Faster drawer openings break free at a higher peak, also with the travel held fixed (partial correlation $\spdFxDrawerOpenPartialPeakSpeedSize$, $p\spdFxDrawerOpenPartialPeakSpeedSizeP$). Among cupboard doors of the same size, faster closings end at a higher peak ($\spdFxCupboardClosePartialPeakSpeedSize$, $p\spdFxCupboardClosePartialPeakSpeedSizeP$). On drawer closings, cupboard-door openings and oven doors, the peak does not rise with speed once the size is held fixed. Over time (Fig.~\ref{fig:speed-compact}), the speed thirds differ more in the height of the breakaway than in its timing: fast drawer openings last about as long as slow ones, because the fast pulls are the long ones, and only the fast cupboard-door openings are clearly shorter.}

\input{tables/rebuttal_speed_ranges_gripper}
\revr{\textbf{Three speed ranges.} The error grows from the slow to the fast range (Table~\ref{tab:speed-all-gripper}), but the ranges hold different objects: the slow range is mostly drawers, while the fast range holds most of the cupboard doors and all the oven doors, the class with the largest errors.}

\input{tables/rebuttal_speed_action_thirds_drawer}
\input{tables/rebuttal_speed_action_thirds_cupboard}
\revr{\textbf{Within each object--action group.} Within an action, the error does not rise steadily with the speed (Tables~\ref{tab:speed-drawer} and~\ref{tab:speed-cupboard}). The oven doors, the refrigerator doors and the other objects have too few executions for thirds; both oven-door openings are predicted far below the measurement. The error does not follow the duration of the execution either ($\rho=\spdFigEvalRhoErrDur$, $p\spdFigEvalRhoErrDurP$).}

\revr{\textbf{Repeated interactions on the same object.} The living room of Eval has no recording in the training split, and its sideboard was opened and closed several times at different speeds, each time with the force measured at the handle (Figs.~\ref{fig:speed-same-object} and~\ref{fig:speed-same-object-examples}). On its drawer openings, the measured peak rises with the speed ($\rho=\spdSoDrawerOpenRhoRef$, $p\spdSoDrawerOpenRhoRefP$), and the faster openings also tend to be the longer pulls; on the other three groups, the trend is not significant. The predicted peak does not rise with the speed in any group. It lies above the measurement on the openings and below it on the drawer closings.}

\revr{\textbf{Which drawers are slow and which are fast.} The slow third holds the toy-kitchen drawer, which has the highest running resistance, and short pulls of the sideboard drawers; the fast third holds full pulls of the sideboard, the bathroom drawer and the two long kitchen drawers (Fig.~\ref{fig:speed-drawers}). Over all measured drawer executions of Hoi!, the handle speed rises with the travel ($\rho=\spdDxAllBothRhoTravel$, $p\spdDxAllBothRhoTravelP$) and falls weakly with the running resistance ($\rho=\spdDxAllBothRhoRun$, $p\spdDxAllBothRhoRunP$): people pull long drawers faster, and a drawer that resists more a little slower.}

\begin{figure}[htbp]\profscaledtable\latefigure
\centering
\includegraphics[width=0.88\linewidth]{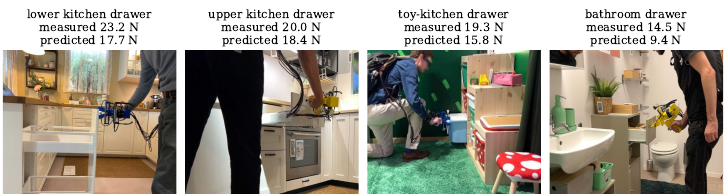}
\caption{\profscaled{\textbf{The single drawers of Table~\ref{tab:drawer-examples}.} One frame per drawer at the open position, with the measured and the predicted peak.}}
\label{fig:drawer-examples}
\end{figure}

\profscaled{\textbf{Drawer openings by object.} Among the drawer openings of Eval, \method{} predicts the two long kitchen drawers well above the bathroom drawer, as measured (Table~\ref{tab:drawer-examples}, Fig.~\ref{fig:drawer-examples}). It does not separate the six drawers of the living-room sideboard, which all look the same.}

\begin{table}[htbp]\profscaledtable\latetable
\centering\scriptsize\setlength{\tabcolsep}{5pt}
\caption{\profscaled{\textbf{Drawer openings of Eval by object.} Measured peak, prediction of \method{} (mean over the three seeds) and peak MAE (mean over the seeds; for the sideboard also over its six drawers).}}
\label{tab:drawer-examples}
\input{tables/drawer_examples}
\end{table}

\revr{\textbf{Limits.} The set is small: each speed third of an action holds only a few recordings, so its numbers are indicative, not significant. The speeds are read from 16 waypoints 0.2 to 0.7\,s apart. The repeats on the sideboard are at the speeds the person happened to use, not at set speeds.}

\FloatBarrier
\renewcommand{\topfraction}{0.95}\renewcommand{\bottomfraction}{0.3}\renewcommand{\textfraction}{0.05}\renewcommand{\floatpagefraction}{0.8}
\setcounter{topnumber}{2}\setcounter{totalnumber}{3}
\makeatletter\setlength{\@fptop}{0pt plus 1fil}\makeatother
\clearpage
\subsection{Additional qualitative results}
\label{app:qualitative}
\begin{figure}[!ht]
\centering
\includegraphics[width=0.88\linewidth]{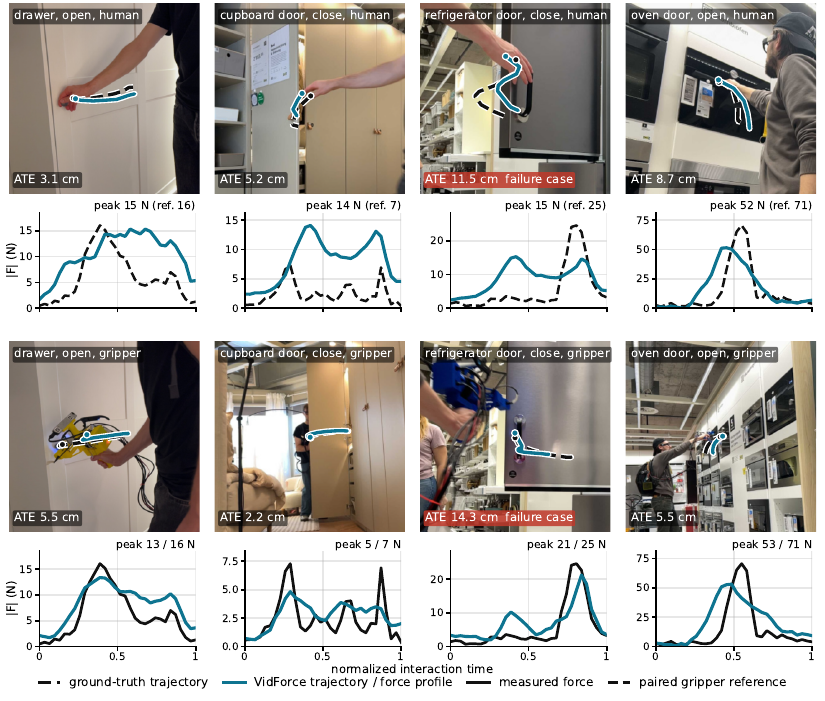}
\caption{\textbf{Qualitative evaluation examples.} Paired human (top) and gripper (bottom) episodes for four interactions. Trajectory overlays report ATE; force plots compare predictions with measured force or paired gripper references. Peak labels show predicted/reference values.}
\label{fig:eval-gallery}
\end{figure}
\begin{figure}[!ht]
\centering
\includegraphics[width=0.88\linewidth]{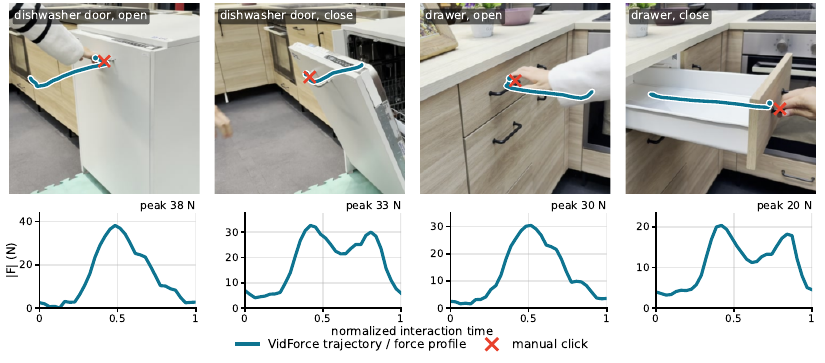}
\caption{\textbf{Real-world videos.} Predicted trajectories and force profiles.}
\label{fig:wild-gallery}
\end{figure}

%% file: tables/extended_trajectory.tex
\begin{table}[htbp]
\centering\scriptsize\setlength{\tabcolsep}{3.5pt}
\caption{\textbf{Trajectory error of every baseline.} Median over sequences on Eval and \EvalP{} (cm); ATE aligns the start by translation, ADE averages the raw waypoint error and so includes the contact-depth error. All: \nVal{} sequences (\nHard{} views on \EvalP); Human: the \nValHand{} human sequences (\nHardHand{} views). CoTracker3 is lifted with per-frame DA3 depth, raw or median-smoothed over five frames; EgoInfinity is read in its own camera or re-projected with our intrinsics at the DA3 contact depth, human sequences only; Articulate-Anything is our adaptation with Gemini~2.5 Flash. Baselines are single runs.}
\label{tab:extended-trajectory}
\begin{tabular}{lcccccccc}
\toprule
 & \multicolumn{4}{c}{Eval} & \multicolumn{4}{c}{\EvalP} \\
\cmidrule(lr){2-5}\cmidrule(lr){6-9}
 & \multicolumn{2}{c}{ATE (cm) $\downarrow$} & \multicolumn{2}{c}{ADE (cm) $\downarrow$} & \multicolumn{2}{c}{ATE (cm) $\downarrow$} & \multicolumn{2}{c}{ADE (cm) $\downarrow$} \\
\cmidrule(lr){2-3}\cmidrule(lr){4-5}\cmidrule(lr){6-7}\cmidrule(lr){8-9}
Method & All & Human & All & Human & All & Human & All & Human \\
\midrule
\multicolumn{9}{l}{\emph{Tracking}} \\
\quad D4RT initialization & 11.8 & 11.1 & 23.7 & 23.0 & 12.8 & 13.9 & 25.4 & 25.4 \\
\quad CoTracker3 + DA3 depth, per-frame depth & 13.7 & 14.2 & 20.6 & 21.9 & 13.5 & 14.2 & 21.0 & 21.2 \\
\quad CoTracker3 + DA3 depth, depth smoothed over 5 frames & 13.3 & 14.2 & 20.3 & 21.1 & 12.6 & 13.4 & 21.2 & 21.4 \\
\multicolumn{9}{l}{\emph{Reconstruction}} \\
\quad Articulate-Anything & 17.5 & 17.3 & 30.4 & 28.9 & 18.9 & 18.9 & 31.6 & 35.1 \\
\quad EgoInfinity, its own camera model & -- & 15.1 & -- & 24.8 & -- & 15.8 & -- & 49.1 \\
\quad EgoInfinity, our camera intrinsics + DA3 depth & -- & 14.2 & -- & 28.1 & -- & 15.9 & -- & 30.3 \\
\midrule
\method{} & \oursAte & \oursHandAte & 11.4 & 14.8 & \oursHardAte & \oursHardHandAte & 13.2 & 16.2 \\
 & $\pm$\oursAteSd & $\pm$\oursHandAteSd & $\pm$1.1 & $\pm$2.0 & $\pm$\oursHardAteSd & $\pm$\oursHardHandAteSd & $\pm$0.2 & $\pm$1.1 \\
\bottomrule
\end{tabular}
\end{table}

%% file: tables/tab_categories.tex
\begin{table}[t]
\centering
\caption{\textbf{Breakdown by partition and object.} We report ATE and force errors with openings and closings combined; the category mean predicts the mean training peak of the object--action category. $n$ denotes all sequences and $n_F$ the instrumented subset used for the measured peak and force errors.}
\label{tab:categories}
\scriptsize\setlength{\tabcolsep}{2pt}
\begin{tabular}{lcccccccc}
\toprule
 & & Measured & \multicolumn{2}{c}{ATE (cm) $\downarrow$} & \multicolumn{2}{c}{Peak MAE (N) $\downarrow$} & \multicolumn{2}{c}{\method{} (N) $\downarrow$} \\
\cmidrule(lr){4-5}\cmidrule(lr){6-7}\cmidrule(lr){8-9}
Subset & $n$ ($n_F$) & peak (N) & D4RT init. & \method & Cat.\ mean & \method & Peak RMSE & Profile RMSE \\
\midrule
\multicolumn{9}{l}{\emph{Partition}} \\
\quad Unseen scene & 47 (23) & 11.0 & 11.1 & $6.7\pm0.3$ & 4.14 & $4.77\pm0.54$ & $5.67\pm0.62$ & \profscaled{$4.80\pm0.49$} \\
\quad Held-out interactions & 52 (25) & 22.2 & 11.9 & $6.6\pm0.4$ & 7.88 & $4.84\pm0.24$ & $6.13\pm0.25$ & \profscaled{$6.11\pm0.13$} \\
\multicolumn{9}{l}{\emph{Object}} \\
\quad Drawer & 40 (20) & 14.7 & 7.8 & $5.3\pm0.6$ & 3.30 & $4.54\pm0.93$ & $5.41\pm0.85$ & \profscaled{$4.67\pm0.54$} \\
\quad Cupboard door & 33 (18) & 8.1 & 13.7 & $6.8\pm0.2$ & 4.85 & $4.26\pm0.23$ & $5.08\pm0.34$ & \profscaled{$4.06\pm0.32$} \\
\quad Oven door & 9 (3) & 69.4 & 11.8 & $7.6\pm1.6$ & 25.28 & $11.30\pm2.32$ & $11.84\pm2.00$ & \profscaled{$12.46\pm0.80$} \\
\quad Refrigerator door & 9 (4) & 32.6 & 23.0 & $9.6\pm0.6$ & 6.69 & $6.05\pm1.83$ & $6.59\pm1.97$ & \profscaled{$7.31\pm0.85$} \\
\quad Other & 8 (3) & 10.2 & 13.9 & $9.5\pm0.8$ & 12.07 & $1.77\pm0.33$ & $2.09\pm0.17$ & \profscaled{$3.92\pm0.16$} \\
\bottomrule
\end{tabular}
\end{table}

%% file: tables/scene_results.tex
\begin{table}[!ht]
\centering\scriptsize\setlength{\tabcolsep}{5pt}
\caption{\textbf{Results by evaluation partition.} Trajectory and force errors for the unseen scene and held-out interactions in seen scenes, together with the combined Eval result.}
\label{tab:scene}
\begin{tabular}{lccccc}
\toprule
Partition & Counts & ATE (cm) $\downarrow$ & \revp{ADE (cm) $\downarrow$} & Peak MAE (N) $\downarrow$ & Peak RMSE (N) $\downarrow$ \\
\midrule
Unseen scene & 47 / 23 & $6.7\pm0.3$ & \revp{$\unseenAde\pm\unseenAdeSd$} & $4.77\pm0.54$ & $5.67\pm0.62$ \\
Held-out interactions, seen scenes & 52 / 25 & $6.6\pm0.4$ & \revp{$\seenAde\pm\seenAdeSd$} & $4.84\pm0.24$ & $6.13\pm0.25$ \\
Combined Eval & 99 / 48 & $6.6\pm0.4$ & \revp{$\oursAde\pm\oursAdeSd$} & $4.81\pm0.38$ & $5.92\pm0.42$ \\
\bottomrule
\end{tabular}
\end{table}

%% file: tables/additional_scene_results.tex
\begin{tabular}{lccccccc}
\toprule
 & & \multicolumn{2}{c}{Trajectory (cm) $\downarrow$} & \multicolumn{2}{c}{Peak force (N) $\downarrow$} & \multicolumn{2}{c}{Force profile (N) $\downarrow$} \\
\cmidrule(lr){3-4}\cmidrule(lr){5-6}\cmidrule(lr){7-8}
Scene & Counts & ATE & ADE & MAE & RMSE & MAE & RMSE \\
\midrule
\texttt{livingroom\_1} (Eval) & 47 / 23 & $6.7\pm0.3$ & $11.0\pm0.6$ & $4.77\pm0.54$ & $5.67\pm0.62$ & \profscaled{$3.76\pm0.42$} & \profscaled{$4.80\pm0.49$} \\
\texttt{livingroom\_2} (new) & 35 / 11 & $11.8\pm1.0$ & $17.7\pm1.8$ & $6.42\pm0.30$ & $7.43\pm0.24$ & \profscaled{$4.05\pm0.18$} & \profscaled{$5.29\pm0.14$} \\
\texttt{kitchen\_6} (new) & 21 / 11 & $4.7\pm0.8$ & $12.7\pm1.4$ & $4.50\pm0.46$ & $7.02\pm0.85$ & \profscaled{$5.17\pm0.48$} & \profscaled{$7.13\pm0.42$} \\
Unseen scenes, combined & 103 / 45 & $7.1\pm0.3$ & $14.1\pm1.7$ & $5.11\pm0.44$ & $6.48\pm0.53$ & \profscaled{$4.17\pm0.35$} & \profscaled{$5.57\pm0.36$} \\
\bottomrule
\end{tabular}

%% file: tables/rebuttal_within_category.tex
\begin{table}[htbp]\revtable
\centering\scriptsize\setlength{\tabcolsep}{1pt}
\caption{\revr{\textbf{Where the prior matters.} Peak MAE (N) on the instrumented Eval sequences, by object class. The word lists and the lookups use GT category information at test time.}}
\label{tab:within-category}
\begin{tabular*}{\linewidth}{@{}p{127pt}@{\extracolsep{\fill}}cccccc@{}}
\toprule
 & Drawer & Cupboard & Oven & Fridge & Other & All \\
Prior & ($n=20$) & ($n=18$) & ($n=3$) & ($n=4$) & ($n=3$) & ($n=48$) \\
\midrule
Global prior + adaptation & $3.73\pm0.56$ & $4.95\pm0.55$ & $22.28\pm0.11$ & $7.84\pm0.45$ & $2.57\pm0.32$ & $5.61\pm0.28$ \\
Word-list prior (LLM-assigned GT category) & $4.29\pm0.56$ & $4.78\pm0.08$ & $4.36\pm1.29$ & $7.00\pm1.38$ & $6.89\pm0.84$ & $4.86\pm0.25$ \\
Word-list prior (GT category) & $4.13\pm0.18$ & $4.83\pm0.23$ & $10.21\pm2.32$ & $6.69\pm1.66$ & $3.42\pm0.34$ & $4.94\pm0.24$ \\
Category-mean lookup\newline(LLM-assigned GT category) & 3.30 & 4.85 & 25.28 & 6.69 & 9.63 & 5.93 \\
Category-mean lookup (GT object--action) & 3.30 & 4.85 & 25.28 & 6.69 & 12.07 & 6.09 \\
\midrule
\method{} (full model) & $4.54\pm0.93$ & $4.26\pm0.23$ & $11.30\pm2.32$ & $6.05\pm1.83$ & $1.77\pm0.33$ & $4.81\pm0.38$ \\
\bottomrule
\end{tabular*}
\end{table}

%% file: tables/rebuttal_force_range.tex
\begin{table}[htbp]\revtable
\centering\scriptsize\setlength{\tabcolsep}{4pt}
\caption{\revr{\textbf{Low and high forces by prior.} Peak MAE split by measured peak. \emph{Predicted peak}: mean on the oven doors and highest over all sequences. The word lists and the lookup use GT category information at test time.}}
\label{tab:force-range}
\begin{tabular}{lcccc}
\toprule
& \multicolumn{2}{c}{Peak MAE (N) $\downarrow$} & \multicolumn{2}{c}{Predicted peak (N)} \\
\cmidrule(lr){2-3}\cmidrule(lr){4-5}
Prior & $<20$\,N ($n=39$) & $\geq 20$\,N ($n=9$) & Oven doors (GT 69.4) & Highest \\
\midrule
Global prior + adaptation & $\frngLightGlobal\pm\frngLightGlobalSd$ & $\frngHeavyGlobal\pm\frngHeavyGlobalSd$ & $\frngOvenGlobal\pm\frngOvenGlobalSd$ & $\frngMaxGlobal\pm\frngMaxGlobalSd$ \\
Flat clusters (k-means, $K{=}10$) & $\frngLightKmeans\pm\frngLightKmeansSd$ & $\frngHeavyKmeans\pm\frngHeavyKmeansSd$ & $\frngOvenKmeans\pm\frngOvenKmeansSd$ & $\frngMaxKmeans\pm\frngMaxKmeansSd$ \\
Word-list prior (LLM-assigned GT category) & $\frngLightLlm\pm\frngLightLlmSd$ & $\frngHeavyLlm\pm\frngHeavyLlmSd$ & $\frngOvenLlm\pm\frngOvenLlmSd$ & $\frngMaxLlm\pm\frngMaxLlmSd$ \\
Word-list prior (GT category) & $\frngLightAnnot\pm\frngLightAnnotSd$ & $\frngHeavyAnnot\pm\frngHeavyAnnotSd$ & $\frngOvenAnnot\pm\frngOvenAnnotSd$ & $\frngMaxAnnot\pm\frngMaxAnnotSd$ \\
Category-mean lookup (GT object--action) & \frngLightCatMean & \frngHeavyCatMean & \frngOvenCatMean & \frngMaxCatMean \\
\midrule
\method{} (full model) & $\frngLightOurs\pm\frngLightOursSd$ & $\frngHeavyOurs\pm\frngHeavyOursSd$ & $\frngOvenOurs\pm\frngOvenOursSd$ & $\frngMaxOurs\pm\frngMaxOursSd$ \\
\bottomrule
\end{tabular}
\end{table}

%% file: tables/rebuttal_limited_supervision_by_k.tex
\begin{table}[htbp]\revtable
\centering\scriptsize\setlength{\tabcolsep}{5pt}
\caption{\revr{\textbf{Limited force supervision by number of prototypes.} Peak MAE (N) on Eval with force labels on a share of the training episodes. Every cell is a run of the limited-supervision study, including the 100\,\% column, so the numbers of prototypes are compared within one set of runs. $K=1$ is the global prior.}}
\label{tab:sensory-by-k}
\begin{tabular}{lcccc}
\toprule
 & \multicolumn{4}{c}{Training episodes with force labels} \\
\cmidrule(lr){2-5}
$K$ & 12\,\% & 25\,\% & 50\,\% & 100\,\% \\
\midrule
1 (global prior) & $8.48\pm0.25$ & $7.20\pm0.75$ & $6.44\pm0.25$ & $5.88\pm0.30$ \\
2 & $7.65\pm0.48$ & $7.27\pm0.13$ & $6.08\pm0.15$ & $4.98\pm0.61$ \\
5 & $7.66\pm0.38$ & $6.34\pm0.28$ & $4.67\pm0.13$ & $5.05\pm0.47$ \\
10 & $7.53\pm0.37$ & $6.27\pm0.46$ & $5.23\pm0.28$ & $5.13\pm0.26$ \\
\bottomrule
\end{tabular}
\end{table}

%% file: tables/rebuttal_k_stage_force_range_by_labels.tex
\begin{table}[htbp]\revtable
\centering\scriptsize\setlength{\tabcolsep}{2.8pt}
\caption{\revr{\textbf{Peak error by share of training force labels, number of prototypes, stage and force range.} The limited-supervision study. Readouts and stages as in Table~\ref{tab:k-stage-range}.}}
\label{tab:k-stage-range-labels}
\begin{tabular}{lllccccc}
\toprule
Training & & & \multicolumn{4}{c}{Peak MAE (N) $\downarrow$} & Predicted peak (N) \\
\cmidrule(lr){4-7}\cmidrule(lr){8-8}
force labels & $K$ & Stage & All ($n=48$) & $<20$\,N ($n=39$) & $\geq20$\,N ($n=9$) & Tail ($n=3$) & Oven doors (GT 69.4) \\
\midrule
all & Category mean & -- & $6.09$ & $4.60$ & $12.53$ & $12.07$ & $44.11$ \\
\midrule
12\,\% & 1 (global prior) & prior & $8.69\pm0.00$ & $3.88\pm0.00$ & $29.50\pm0.00$ & $2.71\pm0.00$ & $12.93\pm0.00$ \\
 &  & + adaptation & $8.47\pm0.24$ & $4.51\pm0.46$ & $25.64\pm0.69$ & $2.79\pm0.72$ & $18.82\pm1.22$ \\
 &  & + flow & $8.48\pm0.25$ & $4.51\pm0.46$ & $25.65\pm0.66$ & $2.81\pm0.72$ & $18.89\pm1.15$ \\
\addlinespace[2pt]
 & 2 & prior & $8.09\pm0.32$ & $3.99\pm0.19$ & $25.84\pm0.94$ & $2.94\pm0.36$ & $17.80\pm0.29$ \\
 &  & + adaptation & $7.64\pm0.47$ & $4.11\pm0.27$ & $22.94\pm1.41$ & $2.48\pm0.15$ & $23.10\pm1.05$ \\
 &  & + flow & $7.65\pm0.48$ & $4.12\pm0.26$ & $22.94\pm1.44$ & $2.44\pm0.18$ & $23.19\pm1.30$ \\
\addlinespace[2pt]
 & 5 & prior & $7.98\pm0.27$ & $3.82\pm0.11$ & $25.99\pm1.02$ & $2.77\pm1.06$ & $18.59\pm1.39$ \\
 &  & + adaptation & $7.66\pm0.37$ & $4.32\pm0.28$ & $22.13\pm1.21$ & $3.72\pm1.67$ & $25.20\pm1.38$ \\
 &  & + flow & $7.66\pm0.38$ & $4.30\pm0.29$ & $22.19\pm1.26$ & $3.67\pm1.65$ & $25.04\pm1.56$ \\
\addlinespace[2pt]
 & 10 & prior & $7.97\pm0.24$ & $3.74\pm0.07$ & $26.29\pm1.02$ & $2.22\pm0.57$ & $19.03\pm0.96$ \\
 &  & + adaptation & $7.55\pm0.38$ & $4.18\pm0.31$ & $22.13\pm0.71$ & $3.70\pm1.43$ & $25.58\pm1.06$ \\
 &  & + flow & $7.53\pm0.37$ & $4.16\pm0.30$ & $22.13\pm0.68$ & $3.61\pm1.33$ & $25.56\pm0.98$ \\
\midrule
25\,\% & 1 (global prior) & prior & $8.91\pm0.00$ & $4.41\pm0.00$ & $28.39\pm0.00$ & $3.82\pm0.00$ & $14.04\pm0.00$ \\
 &  & + adaptation & $8.04\pm0.70$ & $5.34\pm0.99$ & $19.72\pm0.64$ & $3.88\pm1.50$ & $28.07\pm0.00$ \\
 &  & + flow & $7.20\pm0.75$ & $5.38\pm1.03$ & $15.11\pm0.50$ & $3.86\pm1.54$ & $39.54\pm1.72$ \\
\addlinespace[2pt]
 & 2 & prior & $8.38\pm0.36$ & $5.64\pm0.41$ & $20.22\pm0.13$ & $9.88\pm0.67$ & $25.48\pm0.08$ \\
 &  & + adaptation & $7.29\pm0.14$ & $5.85\pm0.31$ & $13.52\pm0.81$ & $9.62\pm0.75$ & $41.38\pm1.92$ \\
 &  & + flow & $7.27\pm0.13$ & $5.86\pm0.29$ & $13.38\pm0.77$ & $9.64\pm0.71$ & $41.55\pm1.89$ \\
\addlinespace[2pt]
 & 5 & prior & $7.51\pm0.21$ & $4.37\pm0.26$ & $21.15\pm0.44$ & $5.46\pm0.72$ & $26.95\pm0.42$ \\
 &  & + adaptation & $6.33\pm0.27$ & $4.91\pm0.42$ & $12.48\pm0.37$ & $6.31\pm0.89$ & $46.20\pm1.53$ \\
 &  & + flow & $6.34\pm0.28$ & $4.91\pm0.38$ & $12.55\pm0.20$ & $6.27\pm0.93$ & $46.02\pm0.98$ \\
\addlinespace[2pt]
 & 10 & prior & $7.51\pm0.14$ & $4.22\pm0.17$ & $21.77\pm0.60$ & $4.49\pm0.20$ & $28.87\pm0.59$ \\
 &  & + adaptation & $6.26\pm0.46$ & $4.98\pm0.43$ & $11.77\pm3.87$ & $6.35\pm0.50$ & $50.46\pm7.85$ \\
 &  & + flow & $6.27\pm0.46$ & $4.97\pm0.41$ & $11.89\pm3.79$ & $6.34\pm0.48$ & $50.13\pm7.72$ \\
\midrule
50\,\% & 1 (global prior) & prior & $8.69\pm0.00$ & $3.88\pm0.00$ & $29.52\pm0.00$ & $2.69\pm0.00$ & $12.91\pm0.00$ \\
 &  & + adaptation & $7.50\pm0.22$ & $4.26\pm0.14$ & $21.52\pm0.57$ & $4.20\pm0.26$ & $25.82\pm0.00$ \\
 &  & + flow & $6.44\pm0.25$ & $4.33\pm0.18$ & $15.59\pm0.61$ & $4.21\pm0.10$ & $41.98\pm0.60$ \\
\addlinespace[2pt]
 & 2 & prior & $7.57\pm0.46$ & $4.52\pm0.53$ & $20.75\pm0.14$ & $9.11\pm0.24$ & $24.48\pm0.56$ \\
 &  & + adaptation & $6.14\pm0.07$ & $4.49\pm0.28$ & $13.26\pm1.00$ & $8.73\pm0.18$ & $45.75\pm3.44$ \\
 &  & + flow & $6.08\pm0.15$ & $4.52\pm0.30$ & $12.83\pm1.03$ & $8.81\pm0.25$ & $46.83\pm3.47$ \\
\addlinespace[2pt]
 & 5 & prior & $6.93\pm0.07$ & $3.63\pm0.03$ & $21.23\pm0.28$ & $3.56\pm0.37$ & $28.14\pm0.32$ \\
 &  & + adaptation & $4.89\pm0.13$ & $3.63\pm0.05$ & $10.36\pm0.49$ & $4.39\pm0.65$ & $55.50\pm0.38$ \\
 &  & + flow & $4.67\pm0.13$ & $3.63\pm0.04$ & $9.19\pm0.51$ & $4.35\pm0.71$ & $58.86\pm0.85$ \\
\addlinespace[2pt]
 & 10 & prior & $7.05\pm0.13$ & $3.65\pm0.15$ & $21.80\pm0.75$ & $3.34\pm0.21$ & $28.77\pm0.15$ \\
 &  & + adaptation & $5.37\pm0.33$ & $3.97\pm0.05$ & $11.42\pm1.85$ & $5.02\pm0.70$ & $54.01\pm2.19$ \\
 &  & + flow & $5.23\pm0.28$ & $3.96\pm0.05$ & $10.74\pm1.67$ & $4.99\pm0.67$ & $56.07\pm1.52$ \\
\midrule
100\,\% & 1 (global prior) & prior & $9.00\pm0.00$ & $4.62\pm0.00$ & $27.98\pm0.00$ & $4.23\pm0.00$ & $14.45\pm0.00$ \\
 &  & + adaptation & $7.15\pm0.23$ & $4.33\pm0.17$ & $19.38\pm0.75$ & $3.11\pm0.37$ & $28.91\pm0.00$ \\
 &  & + flow & $5.88\pm0.30$ & $4.31\pm0.17$ & $12.68\pm1.09$ & $3.18\pm0.32$ & $47.21\pm0.12$ \\
\addlinespace[2pt]
 & 2 & prior & $6.91\pm0.31$ & $4.19\pm0.41$ & $18.72\pm0.23$ & $4.31\pm0.35$ & $26.16\pm0.28$ \\
 &  & + adaptation & $5.23\pm0.39$ & $4.40\pm0.48$ & $8.84\pm0.27$ & $3.39\pm0.43$ & $52.10\pm0.33$ \\
 &  & + flow & $4.98\pm0.61$ & $4.42\pm0.47$ & $7.40\pm1.41$ & $3.39\pm0.34$ & $56.36\pm3.46$ \\
\addlinespace[2pt]
 & 5 & prior & $6.78\pm0.18$ & $3.90\pm0.41$ & $19.23\pm0.82$ & $2.00\pm0.42$ & $29.07\pm0.77$ \\
 &  & + adaptation & $5.14\pm0.48$ & $4.38\pm0.76$ & $8.41\pm0.86$ & $1.76\pm0.26$ & $56.16\pm0.71$ \\
 &  & + flow & $5.05\pm0.47$ & $4.39\pm0.76$ & $7.93\pm0.80$ & $1.73\pm0.27$ & $57.44\pm1.14$ \\
\addlinespace[2pt]
 & 10 & prior & $6.91\pm0.08$ & $4.01\pm0.14$ & $19.51\pm0.94$ & $2.02\pm0.61$ & $30.50\pm0.53$ \\
 &  & + adaptation & $5.14\pm0.24$ & $4.39\pm0.30$ & $8.37\pm1.35$ & $1.59\pm0.54$ & $56.29\pm0.95$ \\
 &  & + flow & $5.13\pm0.26$ & $4.39\pm0.31$ & $8.32\pm1.53$ & $1.57\pm0.57$ & $56.45\pm1.47$ \\
\bottomrule
\end{tabular}
\end{table}

%% file: tables/tab_ksweep.tex
\begin{table}[htbp]
\centering\scriptsize\setlength{\tabcolsep}{5pt}
\caption{\textbf{Number of prototypes.} Fixed recipe on the evaluation set. \emph{Used}: prototypes that carry the largest weight for at least one sequence\revr{, also given as a share of $K$ in percent}; \emph{effective}: exponential of the entropy of that usage over the \nValForce{} instrumented sequences.}
\label{tab:ksweep}
\begin{tabular}{lcccccc}
\toprule
& & & & & \multicolumn{2}{c}{Force error (N) $\downarrow$} \\
\cmidrule(lr){6-7}
$K$ & Used (of $K$) & \revr{Used (\%)} & Effective & ATE (cm) $\downarrow$ & Peak MAE & Peak RMSE \\
\midrule
5 & $\kFiveUsed\pm\kFiveUsedSd$ & \revr{$\kFiveUsedPct\pm\kFiveUsedPctSd$} & $\kFiveEff\pm\kFiveEffSd$ & $\kFiveAte\pm\kFiveAteSd$ & $\kFiveMae\pm\kFiveMaeSd$ & $\kFiveRmse\pm\kFiveRmseSd$ \\
10 (ref.) & $\kTenUsed\pm\kTenUsedSd$ & \revr{$\kTenUsedPct\pm\kTenUsedPctSd$} & $\kTenEff\pm\kTenEffSd$ & $\kTenAte\pm\kTenAteSd$ & $\kTenMae\pm\kTenMaeSd$ & $\kTenRmse\pm\kTenRmseSd$ \\
15 & $\kFifteenUsed\pm\kFifteenUsedSd$ & \revr{$\kFifteenUsedPct\pm\kFifteenUsedPctSd$} & $\kFifteenEff\pm\kFifteenEffSd$ & $\kFifteenAte\pm\kFifteenAteSd$ & $\kFifteenMae\pm\kFifteenMaeSd$ & $\kFifteenRmse\pm\kFifteenRmseSd$ \\
20 & $\kTwentyUsed\pm\kTwentyUsedSd$ & \revr{$\kTwentyUsedPct\pm\kTwentyUsedPctSd$} & $\kTwentyEff\pm\kTwentyEffSd$ & $\kTwentyAte\pm\kTwentyAteSd$ & $\kTwentyMae\pm\kTwentyMaeSd$ & $\kTwentyRmse\pm\kTwentyRmseSd$ \\
30 & $\kThirtyUsed\pm\kThirtyUsedSd$ & \revr{$\kThirtyUsedPct\pm\kThirtyUsedPctSd$} & $\kThirtyEff\pm\kThirtyEffSd$ & $\kThirtyAte\pm\kThirtyAteSd$ & $\kThirtyMae\pm\kThirtyMaeSd$ & $\kThirtyRmse\pm\kThirtyRmseSd$ \\
\bottomrule
\end{tabular}
\end{table}

%% file: tables/tab_ksweep_torque.tex
\begin{table}[htbp]\revtable
\centering\scriptsize\setlength{\tabcolsep}{5pt}
\caption{\revr{\textbf{Number of prototypes, sensor-frame torque.} The recipe of Table~\ref{tab:ksweep} with the torque target, on the evaluation set.}}
\label{tab:ksweep-torque}
\begin{tabular}{lcccc}
\toprule
& & \multicolumn{3}{c}{Torque error (N\,m) $\downarrow$} \\
\cmidrule(lr){3-5}
$K$ & ATE (cm) $\downarrow$ & Peak MAE & Peak RMSE & Profile RMSE \\
\midrule
10 (ref.) & $\tqAte\pm\tqAteSd$ & $\tqOurs\pm\tqOursSd$ & $\tqOursRmse\pm\tqOursRmseSd$ & $\tqOursProf\pm\tqOursProfSd$ \\
20 & $\tqKtwentyAte\pm\tqKtwentyAteSd$ & $\tqKtwenty\pm\tqKtwentySd$ & $\tqKtwentyRmse\pm\tqKtwentyRmseSd$ & $\tqKtwentyProf\pm\tqKtwentyProfSd$ \\
\bottomrule
\end{tabular}
\end{table}

%% file: tables/rebuttal_k_stage_force_range.tex
\begin{table}[htbp]\revtable
\centering\scriptsize\setlength{\tabcolsep}{4pt}
\caption{\revr{\textbf{Peak error by number of prototypes, stage and force range.} \emph{Prior}: before the instance adaptation, i.e.\ the routed prototype mixture. \emph{+ adaptation}: after the instance adaptation, without the flow. \emph{+ flow}: the final prediction.}}
\label{tab:k-stage-range}
\begin{tabular}{llccccc}
\toprule
 & & \multicolumn{4}{c}{Peak MAE (N) $\downarrow$} & Predicted peak (N) \\
\cmidrule(lr){3-6}\cmidrule(lr){7-7}
$K$ & Stage & All ($n=48$) & $<20$\,N ($n=39$) & $\geq20$\,N ($n=9$) & Tail ($n=3$) & Oven doors (GT 69.4) \\
\midrule
Category mean & -- & $6.09$ & $4.60$ & $12.53$ & $12.07$ & $44.11$ \\
\midrule
1 (global prior) & prior & $8.99\pm0.00$ & $4.59\pm0.00$ & $28.02\pm0.00$ & $4.19\pm0.00$ & $14.41\pm0.00$ \\
 & + adaptation & $6.95\pm0.29$ & $4.18\pm0.34$ & $18.97\pm0.21$ & $2.53\pm0.38$ & $28.81\pm0.00$ \\
 & + flow & $5.61\pm0.28$ & $4.18\pm0.33$ & $11.82\pm0.07$ & $2.57\pm0.32$ & $47.11\pm0.11$ \\
\midrule
2 & prior & $6.91\pm0.31$ & $4.19\pm0.41$ & $18.72\pm0.23$ & $4.31\pm0.35$ & $26.16\pm0.28$ \\
 & + adaptation & $5.23\pm0.39$ & $4.40\pm0.48$ & $8.84\pm0.27$ & $3.39\pm0.43$ & $52.10\pm0.33$ \\
 & + flow & $4.98\pm0.61$ & $4.42\pm0.47$ & $7.40\pm1.41$ & $3.39\pm0.34$ & $56.36\pm3.46$ \\
\midrule
5 & prior & $6.81\pm0.23$ & $3.88\pm0.11$ & $19.53\pm1.45$ & $1.90\pm0.82$ & $28.76\pm1.37$ \\
 & + adaptation & $4.90\pm0.43$ & $4.12\pm0.17$ & $8.30\pm2.88$ & $1.58\pm0.81$ & $55.51\pm3.52$ \\
 & + flow & $4.83\pm0.47$ & $4.12\pm0.17$ & $7.90\pm3.09$ & $1.55\pm0.79$ & $56.70\pm4.45$ \\
\midrule
10 (\method) & prior & $6.88\pm0.29$ & $3.99\pm0.34$ & $19.38\pm0.09$ & $2.59\pm0.43$ & $30.64\pm0.50$ \\
 & + adaptation & $4.84\pm0.31$ & $4.23\pm0.46$ & $7.45\pm0.34$ & $1.79\pm0.26$ & $57.55\pm1.27$ \\
 & + flow & $4.81\pm0.38$ & $4.24\pm0.45$ & $7.26\pm0.08$ & $1.77\pm0.33$ & $58.09\pm2.32$ \\
\bottomrule
\end{tabular}
\end{table}

%% file: tables/rebuttal_embodiment_gpt.tex
\begin{table}[htbp]\revtable
\centering\footnotesize\setlength{\tabcolsep}{3.5pt}
\caption{\revr{\textbf{\method{} and GPT-5.5 by embodiment.} Both videos are scored against the peak force and torque measured in the gripper execution. $\Delta$: \method{} minus the better GPT-5.5 row (negative: lower error for \method{}).}}
\label{tab:embodiment-gpt}
\resizebox{\linewidth}{!}{%
\begin{tabular}{lcccccccccc}
\toprule
 & \multicolumn{2}{c}{Predicts} & \multicolumn{4}{c}{Peak force error (N) $\downarrow$} & \multicolumn{4}{c}{Peak torque error (N\,m) $\downarrow$} \\
\cmidrule(lr){2-3}\cmidrule(lr){4-7}\cmidrule(lr){8-11}
 & & & \multicolumn{2}{c}{Gripper video} & \multicolumn{2}{c}{Human video} & \multicolumn{2}{c}{Gripper video} & \multicolumn{2}{c}{Human video} \\
\cmidrule(lr){4-5}\cmidrule(lr){6-7}\cmidrule(lr){8-9}\cmidrule(lr){10-11}
Model & trajectory & force profile & MAE & RMSE & MAE & RMSE & MAE & RMSE & MAE & RMSE \\
\midrule
GPT-5.5 (8 frames) & no & no & 7.98 & 13.11 & 7.95 & 13.72 & 0.675 & 0.829 & 0.630 & 0.788 \\
GPT-5.5 (8 frames + text) & no & no & 8.09 & 12.59 & 8.32 & 13.09 & 0.571 & 0.785 & 0.627 & 0.827 \\
\method{} & yes & yes & $5.15\pm0.25$ & $6.40\pm0.36$ & $10.64\pm0.50$ & $14.09\pm0.72$ & $0.318\pm0.006$ & $0.400\pm0.013$ & $0.356\pm0.023$ & $0.453\pm0.027$ \\
\midrule
$\Delta$ \method{} vs.\ GPT-5.5 & & & $-2.8$ & $-6.2$ & $+2.7$ & $+1.0$ & $-0.25$ & $-0.38$ & $-0.27$ & $-0.33$ \\
\bottomrule
\end{tabular}}
\end{table}

%% file: tables/rebuttal_flipped.tex
\begin{table}[htbp]\revtable
\centering\scriptsize\setlength{\tabcolsep}{4pt}
\caption{\revr{\textbf{Wrong action prompts.} Eval with the manual click; the flipped prompts are encoded with the text pipeline of Appendix~\ref{app:impl:inference}. A wrong action word pulls the predicted peak towards the force level of that action and raises the peak error, but the error stays below that of every baseline in Table~\ref{tab:force}.}}
\label{tab:flipped}
\begin{tabular}{lcccc}
\toprule
Prompt & ATE (cm) $\downarrow$ & ADE (cm) $\downarrow$ & Peak MAE (N) $\downarrow$ & Peak RMSE (N) $\downarrow$ \\
\midrule
recorded prompt, cached embeddings (Table~\ref{tab:selfclick-variants}) & $6.6\pm0.4$ & $11.4\pm1.1$ & $4.81\pm0.38$ & $5.92\pm0.42$ \\
recorded prompt, embeddings re-derived with the text pipeline & $6.6\pm0.4$ & $11.4\pm1.0$ & $4.79\pm0.37$ & $5.90\pm0.41$ \\
action word flipped in every prompt (open $\leftrightarrow$ close) & $7.0\pm0.2$ & $12.0\pm1.2$ & $5.50\pm0.06$ & $7.84\pm0.26$ \\
\bottomrule
\end{tabular}
\end{table}

%% file: tables/automatic_inference.tex
\begin{table}[htbp]\revtable
\centering\scriptsize\setlength{\tabcolsep}{2.5pt}
\caption{\revr{\textbf{Fully automatic inference.} Eval. ATE and ADE in cm (medians); ATE is aligned at the start, ADE is not. Manual click uses the annotated contact and self-click the contact predicted by Qwen. The depth is sampled from DA3 at the pixel named in the second column.}}
\label{tab:automatic}
\resizebox{\linewidth}{!}{%
\begin{tabular}{llcccccc}
\toprule
Contact and crop & Depth pixel & ATE & ADE & ATE human / gripper & ADE human / gripper & Peak MAE (N) & Peak RMSE (N) \\
\midrule
manual click, crop from the annotated extent & annotated & $6.6\pm0.4$ & $11.4\pm1.1$ & $7.9\pm0.8$ / $5.6\pm0.2$ & $14.8\pm2.0$ / $9.0\pm0.3$ & $4.81\pm0.38$ & $5.92\pm0.42$ \\
manual click, fixed crop (0.55 of the frame) & annotated & $7.0\pm0.5$ & $12.8\pm0.1$ & $8.4\pm0.7$ / $5.7\pm0.6$ & $15.4\pm1.0$ / $9.2\pm1.0$ & $4.54\pm0.48$ & $5.57\pm0.64$ \\
self-click, fixed crop & Qwen & $7.1\pm0.5$ & $17.0\pm0.5$ & $8.8\pm0.9$ / $5.7\pm0.0$ & $17.3\pm0.5$ / $16.1\pm0.3$ & $4.48\pm0.48$ & $5.55\pm0.59$ \\
self-click, fixed crop & annotated & $7.0\pm0.3$ & $16.9\pm0.6$ & $8.8\pm0.4$ / $5.8\pm0.3$ & $18.1\pm0.6$ / $16.1\pm1.5$ & $4.53\pm0.42$ & $5.62\pm0.57$ \\
\bottomrule
\end{tabular}}
\end{table}

%% file: tables/rebuttal_speed_ranges_gripper.tex
\begin{table}[htbp]\revtable
\centering\scriptsize\setlength{\tabcolsep}{5pt}
\caption{\revr{\textbf{The measured Eval recordings in three speed ranges.} Gripper recordings of Eval, cut by handle speed into three ranges of about equal size; both phone views of an execution stay in one range. Classes: dr drawer, cu cupboard door, ov oven door, fr refrigerator door, ot other. Measured: median peak. Bias: predicted minus measured.}}
\label{tab:speed-all-gripper}
\begin{tabular}{l l r l c cc c}
\toprule
range & speed (cm/s) & $n$ & classes & measured (N) & MAE (N) & RMSE (N) & bias (N) \\
\midrule
slow & 11--20 & 16 & dr 9, cu 4, ot 3 & 11.3 & 3.71 & 4.48 & $+0.8$ \\
medium & 20--28 & 15 & dr 11, cu 4 & 12.9 & 4.52 & 5.42 & $+0.4$ \\
fast & 28--42 & 17 & cu 10, fr 4, ov 3 & 16.1 & 6.09 & 7.33 & $-1.6$ \\
\midrule
all & 11--42 & 48 &  & 12.7 & 4.81 & 5.92 & $-0.2$ \\
\bottomrule
\end{tabular}
\end{table}

%% file: tables/rebuttal_speed_action_thirds_drawer.tex
\begin{table}[htbp]\revtable
\centering\scriptsize\setlength{\tabcolsep}{5pt}
\caption{\revr{\textbf{The drawers of Eval by action and speed.} For each action, all recordings, then the slow, medium and fast third by handle speed; $n$: phone views. Columns as in Table~\ref{tab:speed-all-gripper}.}}
\label{tab:speed-drawer}
\begin{tabular}{l l r c c cc c}
\toprule
action & third & $n$ & speed (cm/s) & measured (N) & MAE (N) & RMSE (N) & bias (N) \\
\midrule
open & all & 11 & 11--28 & 14.4 & 4.99 & 5.61 & $+1.4$ \\
 & slow & 4 & 11--14 & 15.0 & 5.21 & 5.67 & $+1.6$ \\
 & medium & 3 & 20--21 & 10.6 & 6.46 & 6.95 & $+6.5$ \\
 & fast & 4 & 21--28 & 17.3 & 3.67 & 4.18 & $-2.4$ \\
\addlinespace[3pt]
close & all & 9 & 13--23 & 12.9 & 3.98 & 4.99 & $-3.1$ \\
 & slow & 3 & 13--15 & 12.5 & 4.11 & 4.85 & $-3.4$ \\
 & medium & 2 & 19--20 & 14.5 & 2.26 & 2.32 & $+0.1$ \\
 & fast & 4 & 22--23 & 14.5 & 4.74 & 5.92 & $-4.5$ \\
\bottomrule
\end{tabular}
\end{table}

%% file: tables/rebuttal_speed_action_thirds_cupboard.tex
\begin{table}[htbp]\revtable
\centering\scriptsize\setlength{\tabcolsep}{5pt}
\caption{\revr{\textbf{The cupboard doors of Eval by action and speed.} For each action, all recordings, then the slow, medium and fast third by handle speed; $n$: phone views. Columns as in Table~\ref{tab:speed-all-gripper}.}}
\label{tab:speed-cupboard}
\begin{tabular}{l l r c c cc c}
\toprule
action & third & $n$ & speed (cm/s) & measured (N) & MAE (N) & RMSE (N) & bias (N) \\
\midrule
open & all & 10 & 18--38 & 7.9 & 4.99 & 5.94 & $+4.3$ \\
 & slow & 3 & 18--25 & 7.5 & 4.15 & 5.12 & $+4.1$ \\
 & medium & 3 & 30--33 & 5.5 & 7.30 & 7.53 & $+7.3$ \\
 & fast & 4 & 33--38 & 12.2 & 3.88 & 5.07 & $+2.3$ \\
\addlinespace[3pt]
close & all & 8 & 11--38 & 6.2 & 3.34 & 3.73 & $+1.7$ \\
 & slow & 2 & 11--18 & 9.5 & 2.14 & 2.26 & $-0.3$ \\
 & medium & 3 & 26--26 & 4.8 & 4.81 & 4.90 & $+4.8$ \\
 & fast & 3 & 31--38 & 7.3 & 2.68 & 3.00 & $-0.1$ \\
\bottomrule
\end{tabular}
\end{table}

%% file: tables/drawer_examples.tex
\begin{tabular}{lccc}
\toprule
Drawer opening (Eval) & Measured peak (N) & \method{} (N) & Peak MAE (N) \\
\midrule
Lower kitchen drawer & 23.2 & 17.7 & 5.49 \\
Upper kitchen drawer & 20.0 & 18.4 & 1.78 \\
Toy-kitchen drawer & 19.3 & 15.8 & 3.79 \\
Bathroom drawer & 14.5 & 9.4 & 5.05 \\
Living-room sideboard, 6 drawers & 9.6--14.4 & 16.4--19.1 & 5.84 \\
\bottomrule
\end{tabular}

%% file: sections/appendix_profile_scaling.tex
\subsection{\texorpdfstring{\profscaled{Effect of scaling on force profile}}{Effect of scaling on force profile}}
\label{app:profile-scaling}
\profscaled{The model predicts the peak force and the normalized profile separately. On Eval, the maximum of the predicted profile is \profMaxRatio{} of the predicted peak on median. We scale the predicted profile so that its maximum equals the predicted peak, as part of the standard formulation (Sec.~\ref{sec:flow}, Appendix~\ref{app:impl:flow}). The figures, the robot (Appendix~\ref{app:rw:robot}), the simulation (Appendix~\ref{app:rw:sim}), and all profile errors use the scaled profile. Table~\ref{tab:profile-scaling} compares the profile RMSE before and after scaling, and the peak MAE of the maximum of the unscaled profile with that of the predicted peak.}

\begin{table}[htbp]\profscaledtable\latetable
\centering\scriptsize\setlength{\tabcolsep}{5pt}
\caption{\profscaled{\textbf{Force profile scaled to the predicted peak.} Profile RMSE of the profile as predicted and after scaling its maximum to the predicted peak, and peak MAE of the maximum of the predicted profile and of the predicted peak. Instrumented sequences of Eval and of the unseen scenes of Table~\ref{tab:additional-scenes}.}}
\label{tab:profile-scaling}
\input{tables/profile_scaling}
\end{table}

%% file: tables/profile_scaling.tex
\begin{tabular}{lccccc}
\toprule
 & Instrumented & \multicolumn{2}{c}{Profile RMSE (N) $\downarrow$} & \multicolumn{2}{c}{Peak MAE (N) $\downarrow$} \\
\cmidrule(lr){3-4}\cmidrule(lr){5-6}
 & sequences & predicted & scaled to the peak & maximum of the profile & predicted peak \\
\midrule
Combined Eval & 48 & $4.95\pm0.20$ & $5.52\pm0.25$ & $5.99\pm0.21$ & $4.81\pm0.38$ \\
\texttt{livingroom\_1} (Eval) & 23 & $3.10\pm0.22$ & $4.80\pm0.49$ & $4.13\pm0.46$ & $4.77\pm0.54$ \\
\texttt{livingroom\_2} (new) & 11 & $4.54\pm0.13$ & $5.29\pm0.14$ & $5.13\pm0.15$ & $6.42\pm0.30$ \\
\texttt{kitchen\_6} (new) & 11 & $5.55\pm0.10$ & $7.13\pm0.42$ & $6.31\pm0.43$ & $4.50\pm0.46$ \\
Unseen scenes, combined & 45 & $4.18\pm0.14$ & $5.57\pm0.36$ & $4.91\pm0.17$ & $5.11\pm0.44$ \\
\bottomrule
\end{tabular}

%% file: sections/appendix_realworld.tex
This appendix details the real-world video protocol, supplementary analyses, and robot deployment of Sec.~\ref{sec:realworld}.

\subsection{Real-world videos}
\label{app:rw:videos}
Seven tripod-mounted phone videos show one person opening and closing six kitchen objects: a refrigerator door, freezer, cabinet door, top drawer, oven door, and dishwasher door. We split them into 46 episodes, from the hand reaching the handle until motion stops\rev{, following the motion-span criterion of Appendix~\ref{app:data}}, and sample 32 frames per episode. The \revc{manual inputs} are the operator's \revc{prompt} (e.g., ``open the oven door'') and a manually annotated contact pixel on the 640-pixel frame. Self-click uses the vision--language model's contact proposal; self-prompt asks the model over eight frames what the person does and to which part of which object, without naming the action or appliance. We evaluate all four input combinations on the same episodes, assess repeatability with the manual inputs, and identify action misreads by comparing generated and operator-provided actions.

\revr{\paragraph{Luggage-scale readings.} Table~\ref{tab:luggage} compares the predicted opening peaks with the opening force read from a digital luggage scale. \revp{The scale showed no reading for the cabinet door and the drawer: it needs a few seconds for its reading to stabilize, and these two light objects were opened relatively fast.}}

\input{tables/rebuttal_luggage}

\begin{figure}[t]
\centering
\includegraphics{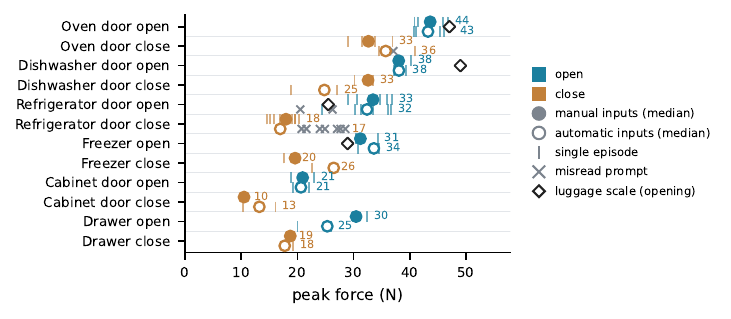}
\caption{\textbf{Real-world peaks by input source.} Filled and open circles show median predictions with manual inputs (operator click and prompt) and automatic inputs (self-click and self-prompt), respectively. Ticks mark individual episodes and crosses misread prompts, which are excluded from the medians; blue denotes opening and orange closing. Diamonds mark the opening force measured with a luggage scale, as in Fig.~\ref{fig:wild}b.}
\label{fig:wild-inputs}
\end{figure}

\paragraph{\revr{Manual and automatic inputs.}}\label{app:rw:inputs}
Fig.~\ref{fig:wild-inputs} shows that predicted peaks vary with both the object and the action, with opening generally requiring more force than closing. Although prototype routing often selects a common force level\reve{, 31\,N for 20 of the 21 correctly read openings with automatic inputs}, instance adaptation and the joint flow separate the interactions using the observed video.

\reve{Self-click gives nearly the same predictions as the manual contact, while the self-prompt steers the estimated effort. Most of its misreads are refrigerator closings read as openings, for which the routed prior rises to the opening level; a reworded prompt can also shift the peak, as in the oven-closing example of Fig.~\ref{fig:wild}.} The trajectory direction \reve{follows the demonstrated action in every episode} under every input combination, even when the prompt is wrong, while the tracked initialization alone \reve{barely moves or moves the wrong way in about a third of them}. Repeated recordings also yield similar peak estimates, with oven opening showing the greatest variation.

\renewcommand{\floatpagefraction}{0.9}
\subsection{Robot deployment}
\label{app:rw:robot}
\begin{figure}[t]
\centering
\includegraphics[width=0.8\linewidth]{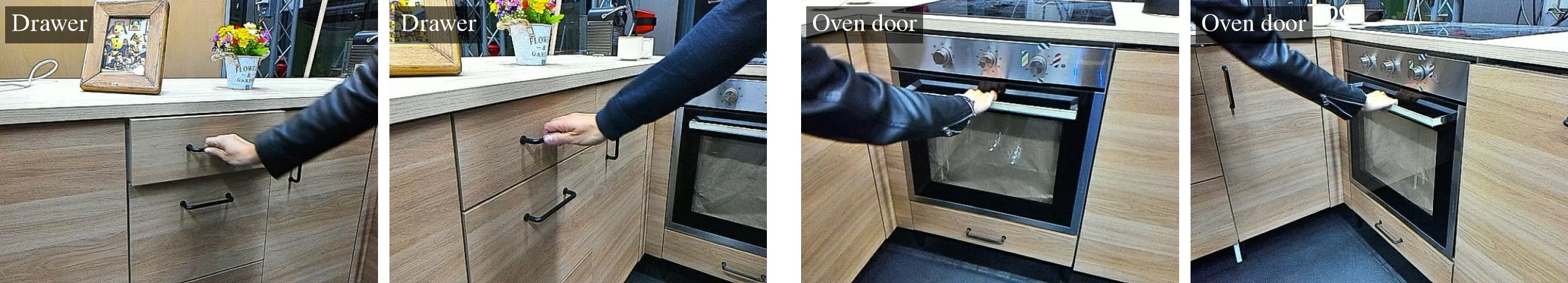}
\caption{\textbf{Spot demonstrations.} A person opening the drawer (left) and the oven door (right), recorded with Spot's arm camera from the two viewpoints per object used in the robot trials.}
\label{fig:spot-demos}
\end{figure}

Spot records a person operating a kitchen unit with its \revb{arm} RGB-D camera from two viewpoints (Fig.~\ref{fig:spot-demos}). The model proposes the contact and generates the prompt, so all robot trials use self-click and self-prompt. \method{} predicts a contact trajectory and force profile from the demonstration, which are exported as a manipulation plan: the grasp, the trajectory, and the predicted force profile as an additional controller signal. Execution trials cover a prismatic and a revolute joint, the drawer and the oven door.
\revp{From each viewpoint, Spot performs five openings of each object in each of three conditions, without force, with a constant force, and with the predicted force profile, 60 openings in total. All three conditions use the same predicted trajectories.} The drawer is empty for the first viewpoint and holds an object of about 5\,kg for the second. A trial succeeds when the robot opens the panel. Failures fall into force errors, where the arm stalled or reached its force limit; trajectory errors, where the gripper left the demonstrated path; and partial openings, where the panel opened partway before the gripper slipped off \revp{or the run was stopped}. A partial opening counts as a failure (\revg{Table~\ref{tab:spot}}). \revp{\revg{The trajectory errors recur in every condition.} Without force, there is one more, on the loaded drawer, where Spot grasped the slab together with the handle.}

\revr{On the real robot, \revp{all three} conditions run the same Cartesian impedance controller on Spot's arm, streamed at 20\,Hz through the SDK's arm impedance command with the task frame fixed to the gravity-aligned odometry frame: the commanded wrench at the tool is $\mathbf F=\mathbf K(\mathbf x_{\mathrm{ref}}-\mathbf x)+\mathbf B(\dot{\mathbf x}_{\mathrm{ref}}-\dot{\mathbf x})+\mathbf F_{\mathrm{ff}}$, with diagonal stiffness $\mathbf K$ (300\,N/m translational, 60\,Nm/rad rotational) and damping $\mathbf B$ (20\,N$\cdot$s/m, 3\,Nm$\cdot$s/rad). The reference pose $\mathbf x_{\mathrm{ref}}(t)$ is the predicted contact-point trajectory placed in the robot frame with the recording's camera pose, \revg{resampled to the control rate} and slowed by a factor of two, with the gripper orientation set at the grasp and either held fixed or turned with the panel along the path. \revp{The conditions differ only in the feed-forward term. Without force, $\mathbf F_{\mathrm{ff}}=\mathbf 0$, so every newton of interaction force must come from the spring, i.e.\ from the hand lagging behind the reference. With the force profile, $\mathbf F_{\mathrm{ff}}(t)$ is the predicted force profile, rescaled to the \revp{planned} peak and applied along the local path tangent, ramped in over 0.3\,s. With a constant force, $\mathbf F_{\mathrm{ff}}$ has the same direction and ramp and a fixed magnitude: 8\,N on the drawer and 20\,N on the oven door. \revg{In all three conditions, the stiffness along the path is 60\,N/m}; lateral and vertical stiffness stay at full value to keep the hand on the path. \revg{All our experiments use exactly matched configurations.}} \revp{All conditions} share the same stop rules, which act only on the measured end-effector force after subtracting a 0.4\,s free-air baseline: \revg{abort at the set safe limit defined below} \revp{(40\,N on the loaded drawer)}, when the hand stalls for 2\,s after 20\% of the path, or at the end of the plan. No measured force is fed back into the command, so the comparison isolates the effect of the predicted force profile itself.}

\revp{\paragraph{Calibrating a constant force.} With the same trajectories and controller settings, we raise a constant force step by step until the panel opens (Table~\ref{tab:spot-sweep}; Fig.~\ref{fig:spot-oven}). \revg{We set a safe limit of 1.3 times the maximum planned force to stop the gripper, to prevent bad forces and motions from damaging the robot.} \revg{The drawer opens at 8\,N. The oven door stays shut up to 20\,N; at 40\,N the gripper gets stuck in the oven (photo in Table~\ref{tab:spot-sweep}), at 50\,N the door opens and closes again, and at 60\,N it opens until the set safe limit stops the run.} A constant force would therefore need an object-specific calibration, from 8\,N for the drawer to 60\,N for the oven door. Table~\ref{tab:spot-oven-numbers} gives the numbers of the runs in Fig.~\ref{fig:spot-oven}.}

\revp{\paragraph{Evidence from real robots.}
\citet{Wang2025ArticuBot} report that force feedback matters for sharp turning motions such as opening a revolute door. Without it, small errors of a predicted trajectory, such as turning too sharply, cause excessive force and stop the arm. \citet{Ji2026DreamingSound} show that a constant force has a failure of its own at first contact. When the desired force steps from 0\,N to a constant of 10 or 15\,N, the measured force can overshoot to the 20\,N safety limit and trigger a protective stop. Like \method{}, they execute a temporal force profile instead; it rises smoothly after contact, and they do not observe this failure. One cause of such an overshoot is that the trajectory term of the controller interferes with the feed-forward: while the hand lags behind its reference, for example at the breakaway, the stiffness adds force on top of the feed-forward. \revg{A natural approach is to relatively lower the stiffness} in the direction of motion while the feed-forward is on. Their profile study uses whiteboard wiping and carrot peeling, where the tool stays on a surface. \revg{Doors and drawers add another issue, the build-up of kinetic energy: under a constant force high enough to open them, the panel hits the end stop with a high impact energy, and a temporal profile keeps it low (Appendix~\ref{app:rw:sim}).} A constant-force controller can be calibrated, but it has to be designed by hand for each task. \citet{Ji2026DreamingSound} also note that, on sustained-contact tasks, a constant force can perform comparably when it is tuned for that specific task.}

\makeatletter\setlength{\@fptop}{0pt}\setlength{\@fpsep}{\floatsep}\makeatother
\begin{table}[htbp]\revtable
\centering
\caption{\revg{\textbf{Calibrating a constant force.} The constant force is raised step by step until the panel opens. Left: the gripper stuck in the oven at 40\,N. The trials of Table~\ref{tab:spot} use 8\,N on the drawer and 20\,N on the oven door.}}
\label{tab:spot-sweep}
\begin{minipage}[c]{0.28\textwidth}
\includegraphics[width=\linewidth]{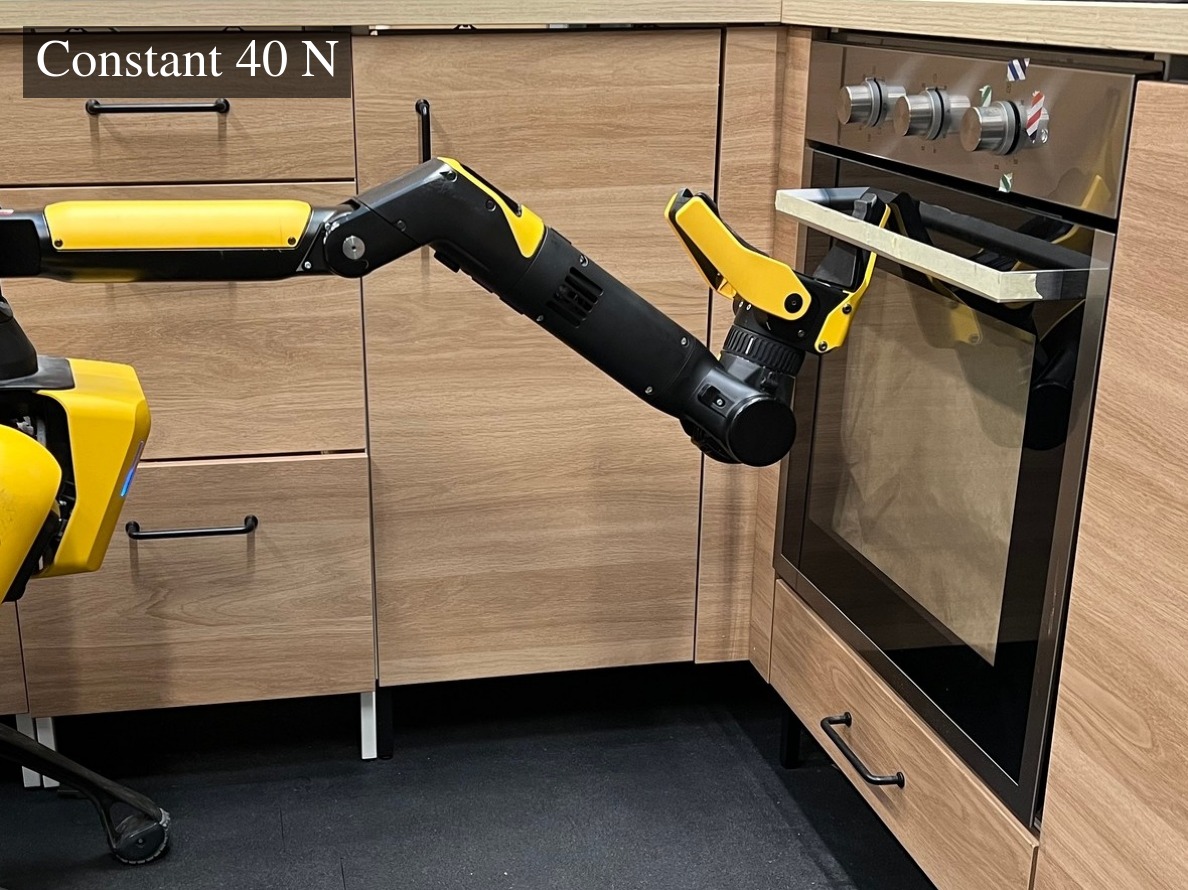}
\end{minipage}\hfill
\begin{minipage}[c]{0.68\textwidth}
\centering\scriptsize
\begin{tabular*}{\linewidth}{@{\extracolsep{\fill}}lccl@{}}
\toprule
\revp{Object} & \revp{Force (N)} & \revp{Opening (cm)} & \revp{Outcome} \\
\midrule
\revp{Drawer} & \revp{1} & \revp{2.1} & \revp{does not open} \\
 & \revp{4} & \revp{5} & \revp{does not open} \\
 & \revp{8} & \revp{22.2} & \revp{opens} \\
\midrule
\revp{Oven door} & \revp{5} & \revp{2.3} & \revp{does not open} \\
 & \revp{10} & \revp{2.3} & \revp{does not open} \\
 & \revp{20} & \revp{2.7} & \revp{does not open} \\
 & \revp{40} & \revp{3.2} & \revp{gripper stuck} \\
 & \revp{50} & \revp{24} & \revp{opens, then closes} \\
 & \revp{60} & \revg{19.9} & \revp{opens} \\   %
\bottomrule
\end{tabular*}
\end{minipage}
\end{table}

\begin{table}[htbp]\revtable
\centering
\caption{\revg{\textbf{Oven door on Spot.} One of the 10 trajectories is left out: it had a localization error and did not result in a grasp. Success: the door opens, stays open, and the run is not stopped. Means over runs (force profile: its successful runs). \emph{At the end}: largest value in the last 0.5\,s. Kinetic energy: assumed 9\,kg door from the simulation. $\times$: the motion was not completed.}}
\label{tab:spot-oven-numbers}
\scriptsize\setlength{\tabcolsep}{5pt}
\input{tables/spot_oven_numbers}
\end{table}

\subsection{\texorpdfstring{\revr{Simulation: temporal force profile versus constant feed-forward}}{Simulation: temporal force profile versus constant feed-forward}}
\label{app:rw:sim}
\begin{figure}[htbp]
\centering
\includegraphics[width=0.8\linewidth]{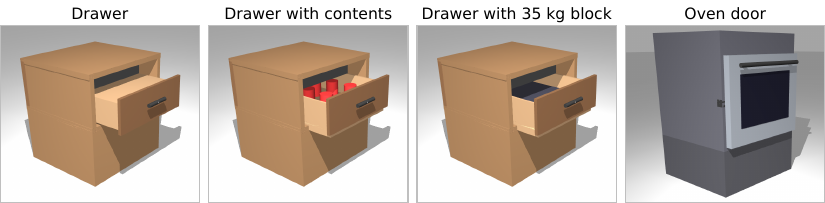}
\caption{\revr{\textbf{The hand-built twins.} The drawer with its three loads, shown halfway open, and the oven door, shown closed.}}
\label{fig:sim-twins}
\end{figure}

\revr{A constant force is the natural control for the predicted force profile: would a constant at the predicted peak, or the category's mean force, do what the profile does on the robot? The trials of \revg{Table~\ref{tab:spot}} compare the profile with \revp{a constant force and with} no feed-forward; here we answer the question in simulation, in two studies in which every feed-forward runs under the same controller gains, force cap and metrics: hand-built twins with the literature's mechanism forces, and twins of the Eval episodes driven by the model's own predictions.

\paragraph{Why a constant cannot stand in for the profile.}
The force that a door or drawer requires is not stationary, and it differs from one object to the next \citep{Jain2010doors,Jain2013forces}. A latch, seal or spring can make the force at the start of the travel far larger than the force behind it: a refrigerator door needs a high force while it is closed and very little after that \citep{Jain2010doors}, and so does a drawer on detent slides \citep{Schweers2011DrawerSlide}. A constant pull equal to the breakaway force keeps pulling after the panel has broken free; the surplus work, $\int (F-F_{\mathrm{req}})\,\mathrm{d}s$, becomes kinetic energy of the panel, which is dissipated at the end stop or on whatever is in the gap \citep{Haddadin2009safe}, and ISO/TS~15066 \citep{ISO15066} limits the energy a moving part may transfer to a hand to 0.49\,J and the clamping force to 140\,N (these limits can be taken as a proxy for a robotic gripper or hand for the analysis). A constant equal to the mean of the profile does not overcome the breakaway force. Only a force that is high at the breakaway and low afterwards does both (a time-varying command, as in learned variable-impedance control; \citealp{Buchli2011vic}), and that is what the model predicts.

\paragraph{Simulator and twins.}
Both studies run in MuJoCo~3.4 \citep{Todorov2012MuJoCo} at a time step of 0.2\,ms. We did not scan or download any assets: each twin is a small model written by hand from boxes and cylinders, with a cabinet, a panel on a single slide or hinge joint, its handle, and a solid block as the end stop (Fig.~\ref{fig:sim-twins}). The joint carries a running friction, a small viscous damping, and a position-dependent resistance that stands for the breakaway force of a latch or seal. On the hand-built twins we set these by hand. The drawer is modeled on a drawer measured on detent slides, which took 34 to 39\,N to break out of the detent and then 0.5 to 2.0\,N to move along its 28\,cm of travel \citep{Schweers2011DrawerSlide}: ours has the same travel, a mass of 4\,kg, a breakaway resistance of 36\,N that falls linearly to zero over the first 15\,mm, and a running friction of 1.5\,N. The 9\,kg oven door is counterbalanced and has an 18\,N latch that falls linearly to zero over its first $10^\circ$. Loose contents are free bodies on the drawer floor. On the Eval twins we read them from the force measured in that episode: its 25th percentile while the panel moves is the running friction, and what lies above it is kept as a function of the panel's position. Each Eval twin takes its travel from the measured trajectory and a typical mass for its object class (3\,kg for a cupboard door to 15\,kg for a refrigerator door).}

\input{tables/rebuttal_sim_controlled}
\begin{figure}[htbp]
\centering
\includegraphics[width=\linewidth]{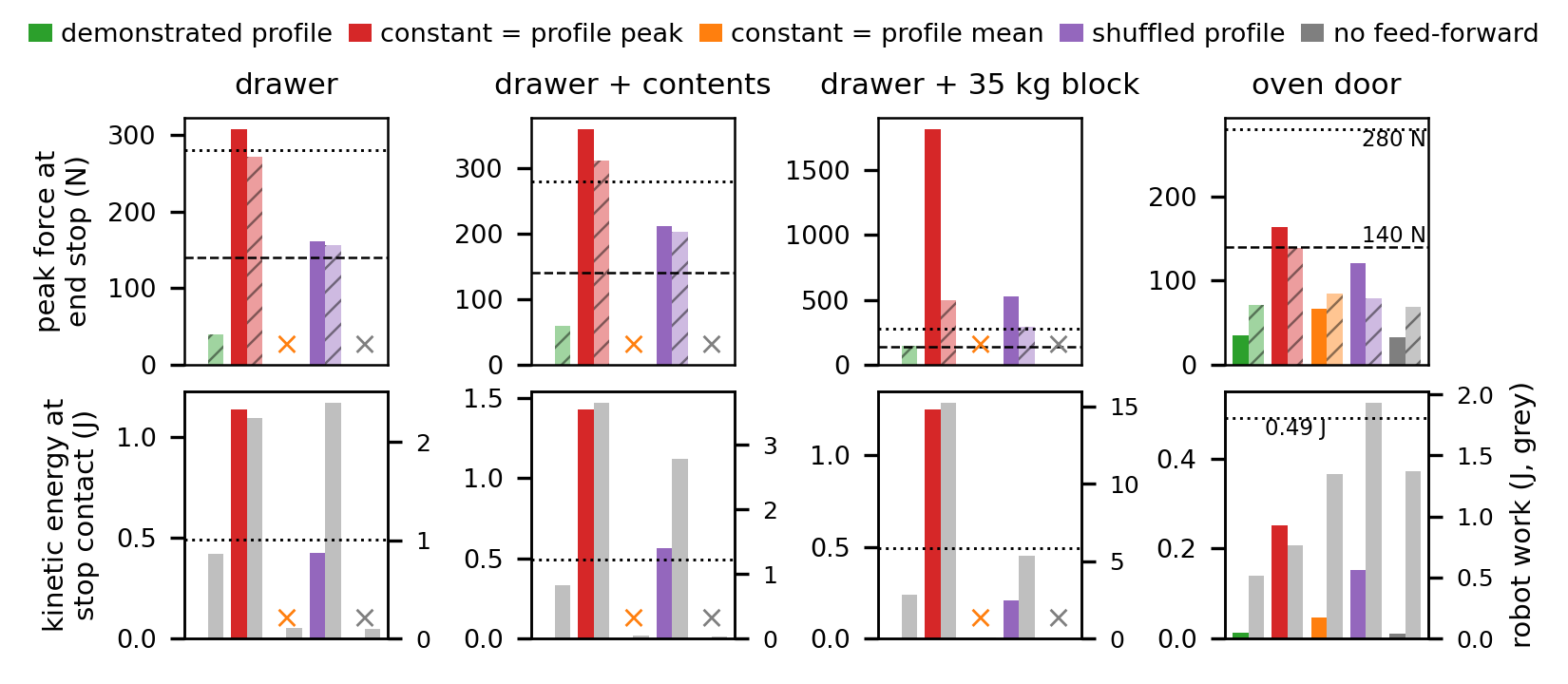}
\caption{\revr{\textbf{Hand-built twins at $K=100$\,N/m.} Top: peak force at the end stop (hatched: on the ISO hand in the gap). Bottom: kinetic energy at the first stop contact, and the work done by the robot (grey, right axis). Dashed and dotted lines: ISO/TS~15066 hand limits; $\times$: the panel does not reach the end of its travel.}}
\label{fig:sim-controlled}
\end{figure}

\input{tables/rebuttal_sim_tier0}
\revr{\paragraph{Controller and metrics.}
The simulated robot runs the controller of the real trials (Appendix~\ref{app:rw:robot}): a Cartesian impedance law \citep{Hogan1985impedance} along the panel's path, $F=m\,F_{\mathrm{ff}}(t)+K\,(s_{\mathrm{ref}}-s)-D\,\dot s$, that follows the demonstration's progress $s_{\mathrm{ref}}$ with the compliant stiffness $K=100$\,N/m (30 and 300\,N/m where stated) and $D$ at 0.7 of critical damping. All feed-forwards share these gains, the force cap (100\,N on the Eval twins, 40 to 100\,N on the hand-built twins) and the margin $m=1.10$, the smallest at which the \emph{measured} profile completes 90\% of the Eval twins (a feed-forward equal to the resistance leaves no net force at the breakaway). On the Eval twins, the conditions differ only in $F_{\mathrm{ff}}$: the predicted profile, rescaled to the predicted peak and played over the execution window as on the robot; a constant at the predicted peak; a constant at the category mean (the mean predicted force of the other Eval episodes of the same class and action) or at the category peak (the mean training peak of the object and action); the predicted profile with its samples shuffled; the measured profile as an oracle; and no feed-forward. Shuffling keeps the force values, and with them the peak and the mean, and changes only their timing (five random orders, averaged). We record whether the panel reaches the end of its travel, its kinetic energy when it first hits the end stop, the peak force on an ISO/TS~15066 hand model in the gap before the stop, and whether a closing slams into the frame (above 0.49\,J).

\paragraph{Force profiles, success and impact energy.}
In the simulation, the robot is reduced to the force it applies: the force of the control law acts at the handle, along its path, with no arm or gripper. On the hand-built twins, the profile comes from a demonstration in the simulator: a stiff position controller pulls the panel open along a minimum-jerk motion (2\,s for the drawers, 3\,s for the oven door), and the force it needs is reduced to 32 samples, like the model's output. On the Eval twins, it is the model's prediction. The feed-forward interpolates the samples over the execution window, and the simulation runs 1.5\,s beyond it. An execution succeeds if the panel reaches 95\% of its travel in this time. The impact energy is the panel's kinetic energy just before it first touches the end stop, $E=\tfrac{1}{2}\,m_{\mathrm{eff}}\,v^2$, with $v$ the handle's speed along its path and $m_{\mathrm{eff}}$ the panel's mass as felt at the handle (for a door, its moment of inertia about the hinge over the squared handle distance). A panel that never touches the stop counts as 0\,J, and loose contents are not included.

\paragraph{Hand-built twins.}
Table~\ref{tab:sim-controlled} and Fig.~\ref{fig:sim-controlled} isolate the mechanism on the four hand-built twins (Fig.~\ref{fig:sim-twins}): the drawer, empty, with loose cans and boxes, or with a free 35\,kg block, and the oven door. On each, the demonstrated profile is compared with a constant at its peak or its mean, its shuffled samples and no feed-forward. At $K=100$\,N/m, a constant at the peak keeps pulling after the breakaway: the three drawers exceed both ISO hand limits, the contents slide into the drawer walls, and the block strikes the drawer front. A stiffer controller ($K=300$\,N/m) keeps the two lighter drawers within the limits, but not the one with the 35\,kg block. A constant at the mean, like no feed-forward, does not overcome the drawers' breakaway at $K=100$\,N/m. The demonstrated profile opens all four twins with almost no impact, while the same forces in a shuffled order open them but hit the stop: the timing of the force matters, not only its values.}

\revr{\paragraph{The predictions on twins of the Eval episodes.}
On the twin of each instrumented Eval episode, the simulated robot executes the model's prediction for that episode (Table~\ref{tab:sim-tier0}). Played against time, the predicted profile completes most episodes, with no impact above the 0.49\,J limit and no slammed closing; it stalls on the others, the oven twins among them. Indexed by the panel's progress instead, it completes nearly every episode (\simTzProfProgReach{}), the oven twins included, at the same low impact energy. No constant does both. A constant at the predicted or the category peak completes every episode, but at twice the impact energy, with impacts above the limit and slammed closings. A constant at the category mean is as gentle as the profile, because it leaves most of the work to the position loop, but it never opens the oven twins, even when indexed by progress, and at $K=30$\,N/m it completes fewer episodes than the profile. The shuffled profile, with the same forces in a random order, hits harder than the profile, and without feed-forward the compliant controller opens fewer than half of the episodes.

\paragraph{Limitations.}
The twins are single-joint models with category masses and the literature's mechanism forces; the simulated robot has no joint-level dynamics, grasp compliance or sensing noise.}

\FloatBarrier

%% file: tables/rebuttal_luggage.tex
\begin{table}[htbp]\revtable
\centering\scriptsize\setlength{\tabcolsep}{5pt}
\caption{\revr{\textbf{Predicted opening peaks against the luggage-scale readings.} Openings: episodes with a correctly read action. Predicted: median peak with automatic (self-click, self-prompt) or manual inputs. $\Delta$: predicted minus reading.}}
\label{tab:luggage}
\begin{tabular}{lccccc}
\toprule
Object & Scale (N) & Openings & Predicted, automatic (N) & Predicted, manual (N) & $\Delta$ automatic / manual (N) \\
\midrule
oven door & 47.1 & 4 & 43.3 & 43.6 & $-3.8$ / $-3.4$ \\
dishwasher door & 49.0 & 3 & 38.1 & 38.0 & $-11.0$ / $-11.0$ \\
refrigerator door & 25.5 & 7 & 33.5 & 33.5 & $+8.0$ / $+8.0$ \\
freezer drawer & 28.9 & 2 & 32.6 & 31.2 & $+3.7$ / $+2.3$ \\
cabinet door & no reading & 2 & 20.6 & 20.9 & -- \\
drawer & no reading & 3 & 25.3 & 30.5 & -- \\
\bottomrule
\end{tabular}
\end{table}

%% file: tables/spot_oven_numbers.tex
\begin{tabular}{@{}lccccc@{}}
\toprule
 & \multirow{2}{*}{\begin{tabular}{@{}c@{}}Force\\profile\end{tabular}} & \multicolumn{4}{c}{Constant force} \\
\cmidrule(lr){3-6}
 &  & 20\,N & 40\,N & 50\,N & 60\,N \\
\midrule
Success & 8/9 & 0/9 & 0/1 & 0/2 & 0/1 \\
Outcome & opens & \begin{tabular}[t]{@{}c@{}}never\\opens\end{tabular} & \begin{tabular}[t]{@{}c@{}}stuck\\at 3\,cm\end{tabular} & \begin{tabular}[t]{@{}c@{}}opens,\\then closes\end{tabular} & \begin{tabular}[t]{@{}c@{}}opens, stopped\\at the limit\end{tabular} \\
Set safe limit (N) & 74.9 & 75.2 & 71.5 & 71.5 & 71.5 \\
Largest measured force at the gripper (N) & 66.8 & 61.3 & 73.1 & 71.3 & 74.2 \\
Measured force at the gripper after 0.5\,s (N) & 16.0 & 27.0 & 40.3 & 55.1 & 60.8 \\
Feed-forward at the end of the trajectory (N) & 5.9 & 20.0 & 40.0 & 50.0 & $\times$ \\
Gripper speed at the end (cm/s) & 18 & 2 & 2 & 52 & 91 \\
Kinetic energy at the end (J) & 0.07 & $\times$ & $\times$ & 0.53 & 1.53 \\
\bottomrule
\end{tabular}

%% file: tables/rebuttal_sim_controlled.tex
\begin{table}[htbp]\revtable
\centering\scriptsize\setlength{\tabcolsep}{5pt}
\caption{\revr{\textbf{Hand-built twins: what a constant pull does to the panel, a hand and the contents.} Each cell gives the value at $K=100$ / $K=300$\,N/m. \emph{Reached end}: the panel reaches at least 95\% of its travel (shuffled: draws out of five). \emph{Impact energy}: kinetic energy of the panel when it first hits the end stop. \emph{Stop force}: peak force on the end stop. \emph{Hand force}: in a separate run, peak force on an ISO/TS~15066 hand model (0.6\,kg on a 75\,N/mm spring) in the gap before the stop, a proxy for a gripper in that gap. \emph{Contents}: peak force of the loose contents on the drawer walls, and their slide at $K=100$\,N/m.}}
\label{tab:sim-controlled}
\newcommand*{\twohead}[2]{\begin{tabular}[b]{@{}c@{}}#1\\#2\end{tabular}}\newcommand*{\twolab}[2]{\begin{tabular}{@{}l@{}}#1\\#2\end{tabular}}
\begin{tabular}{llccccc}
\toprule
Twin & Feed-forward & \twohead{Reached}{end} & \twohead{Impact}{energy (J)} & \twohead{Stop}{force (N)} & \twohead{Hand}{force (N)} & \twohead{Contents:}{force (N); slide} \\
\midrule
\multirow{5}{*}{\twolab{drawer}{(4 kg, 36 N breakaway)}} & demonstrated profile & yes / yes & 0.000 / 0.000 & 0 / 0 & 39 / 50 & -- \\
 & constant = profile peak & yes / yes & 1.14 / 0.138 & 307 / 107 & 272 / 91 & -- \\
 & constant = profile mean & no / yes & -- / 0.016 & -- / 35 & -- / 66 & -- \\
 & shuffled profile & 5/5 / 5/5 & 0.426 / 0.202 & 161 / 109 & 157 / 96 & -- \\
 & no feed-forward & no / yes & -- / 0.000 & -- / 0 & -- / 51 & -- \\
\midrule
\multirow{5}{*}{\twolab{drawer + 3.4 kg}{free contents}} & demonstrated profile & yes / yes & 0.000 / 0.000 & 0 / 0 & 59 / 71 & 0 / 0; 1.0 cm \\
 & constant = profile peak & yes / yes & 1.43 / 0.132 & 359 / 115 & 312 / 95 & 85 / 0; 8.1 cm \\
 & constant = profile mean & no / yes & -- / 0.055 & -- / 79 & -- / 124 & 0 / 0; 0.9 cm \\
 & shuffled profile & 5/5 / 5/5 & 0.565 / 0.153 & 212 / 116 & 202 / 132 & 31 / 10; 3.6 cm \\
 & no feed-forward & no / yes & -- / 0.008 & -- / 32 & -- / 108 & 0 / 0; 0.9 cm \\
\midrule
\multirow{5}{*}{\twolab{drawer + 35 kg}{free block}} & demonstrated profile & yes / yes & 0.000 / 0.000 & 0 / 0 & 143 / 150 & 0 / 0; 0.0 cm \\
 & constant = profile peak & yes / yes & 1.25 / 0.496 & 1817 / 805 & 501 / 373 & 1544 / 0; 6.2 cm \\
 & constant = profile mean & no / yes & -- / 0.465 & -- / 701 & -- / 358 & 0 / 0; 0.0 cm \\
 & shuffled profile & 5/5 / 5/5 & 0.207 / 0.520 & 526 / 828 & 289 / 381 & 0 / 0; 1.3 cm \\
 & no feed-forward & no / yes & -- / 0.277 & -- / 692 & -- / 314 & 0 / 0; 0.0 cm \\
\midrule
\multirow{5}{*}{\twolab{oven door}{(9 kg, 18 N latch)}} & demonstrated profile & yes / yes & 0.012 / 0.003 & 35 / 19 & 71 / 69 & -- \\
 & constant = profile peak & yes / yes & 0.253 / 0.100 & 163 / 102 & 139 / 107 & -- \\
 & constant = profile mean & yes / yes & 0.047 / 0.015 & 66 / 39 & 85 / 76 & -- \\
 & shuffled profile & 5/5 / 5/5 & 0.153 / 0.087 & 121 / 88 & 79 / 99 & -- \\
 & no feed-forward & yes / yes & 0.011 / 0.003 & 33 / 19 & 68 / 69 & -- \\
\bottomrule
\end{tabular}
\end{table}

%% file: tables/rebuttal_sim_tier0.tex
\begin{table}[htbp]\revtable
\centering\scriptsize\setlength{\tabcolsep}{5pt}
\caption{\revr{\textbf{Twins of the 48 instrumented Eval episodes: the predicted profile against constant and control feed-forwards.} Same controller for every row, $K=100$\,N/m. Medians over episodes; IQR: interquartile range of the impact energy. Reached end, impact energy and hand force as in Table~\ref{tab:sim-controlled}; $>0.49$\,J and $>140$\,N: episodes above the ISO/TS~15066 hand limits. Slammed: closings that hit the frame above 0.49\,J. The oracle replays the measured profile.}}
\label{tab:sim-tier0}
\newcommand*{\twohead}[2]{\begin{tabular}[b]{@{}c@{}}#1\\#2\end{tabular}}
\begin{tabular}{lccccccc}
\toprule
Feed-forward & \twohead{Reached}{end} & \twohead{Impact}{energy (J)} & IQR (J) & $>0.49$\,J & \twohead{Hand}{force (N)} & $>140$\,N & \twohead{Slammed}{closings} \\
\midrule
predicted profile & 83\% & 0.066 & 0.012--0.117 & 0\% & 68 & 15\% & 0 of 19 \\
constant = predicted peak & 100\% & 0.135 & 0.086--0.212 & 12\% & 87 & 25\% & 2 of 19 \\
constant = category mean & 85\% & 0.045 & 0.018--0.115 & 0\% & 80 & 8\% & 0 of 19 \\
constant = category peak & 100\% & 0.134 & 0.076--0.192 & 10\% & 78 & 25\% & 2 of 19 \\
shuffled profile & 92\% & 0.095 & 0.041--0.158 & 8\% & 83 & 15\% & 0 of 19 \\
measured profile (oracle) & 96\% & 0.028 & 0.003--0.344 & 10\% & 76 & 31\% & 2 of 19 \\
no feed-forward & 38\% & 0.000 & 0.000--0.000 & 0\% & 34 & 4\% & 0 of 19 \\
\bottomrule
\end{tabular}
\end{table}